\documentclass[journal]{IEEEtran}
\usepackage{amsmath,amsfonts,bbm}
\usepackage{algorithm}
\usepackage{array}
\usepackage[caption=false,font=footnotesize,labelfont=sf,textfont=sf]{subfig}
\usepackage{textcomp}
\usepackage{stfloats}
\usepackage{url}
\usepackage{verbatim}
\usepackage{graphicx}
\usepackage{cite}
\usepackage{booktabs}
\usepackage{multirow}
\usepackage{color,soul}
\usepackage{algpseudocode} 

\begin{document}

\title{A Deep Positive-Negative Prototype Approach to Integrated Prototypical Discriminative Learning}

\author{\IEEEauthorblockN{Ramin Zarei-Sabzevar\IEEEauthorrefmark{1},
Ahad Harati\IEEEauthorrefmark{1}}

\IEEEauthorblockA{\IEEEauthorrefmark{1}
Department of Computer Engineering, Ferdowsi University of Mashhad (FUM), Mashhad, Iran}
\thanks{This work has been submitted to the IEEE for possible publication. Copyright may be transferred without notice, after which this version may no longer be accessible. (email: rzarei@mail.um.ac.ir, a.harati@um.ac.ir)}}

\markboth{Journal of \LaTeX\ Class Files,~Vol.~14, No.~8, August~2024}%
{Shell \MakeLowercase{\textit{et al.}}: A Sample Article Using IEEEtran.cls for IEEE Journals}


\maketitle

\begin{abstract}
This paper proposes a novel Deep Positive-Negative Prototype (DPNP) model that combines prototype-based learning (PbL) with discriminative methods to improve class compactness and separability in deep neural networks. While PbL traditionally emphasizes interpretability by classifying samples based on their similarity to representative prototypes, it struggles with creating optimal decision boundaries in complex scenarios. Conversely, discriminative methods effectively separate classes but often lack intuitive interpretability. Toward exploiting advantages of these two approaches, the suggested DPNP model bridges between them by unifying class prototypes with weight vectors, thereby establishing a structured latent space that enables accurate classification using interpretable prototypes alongside a properly learned feature representation.

Based on this central idea of unified prototype-weight representation, Deep Positive Prototype (DPP) is formed in the latent space as a representative for each class using off-the-shelf deep networks as feature extractors. Then, rival neighboring class DPPs are treated as implicit negative prototypes with repulsive force in DPNP, which push away DPPs from each other. This helps to enhance inter-class separation without the need for any extra parameters. Hence, through a novel loss function that integrates cross-entropy, prototype alignment, and separation terms, DPNP achieves well-organized feature space geometry, maximizing intra-class compactness and inter-class margins. We show that DPNP can organize prototypes in nearly regular positions within feature space, such that it is possible to achieve competitive classification accuracy even in much lower-dimensional feature spaces. This fact may increase performance multiple folds. Experimental results on several datasets demonstrate that DPNP outperforms state-of-the-art models, while using smaller networks.
\end{abstract}

\begin{IEEEkeywords}
prototype-based learning, positive and negative prototypes, discriminative learning, low-dimensional latent space, regular geometry feature space.
\end{IEEEkeywords}

\section{Introduction}
\IEEEPARstart{T}{he} rapid evolution of artificial intelligence has led to significant advances in the field of representation learning, particularly with the integration of deep learning methods. Representation learning aims to automatically extract meaningful features from raw data, enabling efficient and accurate decision-making in various machine learning tasks, such as classification, clustering, and retrieval. However, the challenge of developing representations that are both discriminative and interpretable remains a key research area.

Prototype-based learning (PbL) and discriminative approaches have emerged as two prominent methodologies to address these challenges. PbL is rooted in representing classes with a set of prototypes, allowing classification based on similarity to these prototypes. This approach has a natural interpretability due to its reliance on distances to representative points. However, it often struggles with defining optimal decision boundaries in complex scenarios. On the other hand, discriminative methods aim to create clear decision boundaries that separate classes effectively, using a range of sophisticated loss functions to enhance class separability.

Recent efforts have sought to integrate these two approaches to harness the interpretability of prototype-based models while benefiting from the powerful discrimination capabilities of modern deep learning techniques. State-of-the-art methods are quite successful in offering hybrid models that use prototypes in combination with discriminative terms in the loss function \cite{Shi2023,Yang2022}. Specially when the number of classes is very large, such as in face recognition applications, these methods are frequently used \cite{Deng2022}. However, such models mostly use different sets of parameters for classification and discrimination. This redundancy separates between discrimination and classification aspects of the model, leading to more complicated learning algorithms and potentially inferior representations. In this paper, we propose a novel approach that shares one set of parameters for both classification and discrimination. The resulting model, called Deep Positive-Negative Prototype (DPNP), directly utilizes the learned prototypes for classification, while reusing them in the discriminative terms of its loss function. Therefore, DPNP combines prototype representation benefits with the regularity and generalization offered by discriminative learning, aiming to enhance accuracy while using interpretable prototypes. On the other hand, using both positive and negative prototypes, our method can learn models that enhance class compactness and inter-class separability. Consequently, DPNP provides a more nuanced understanding of class boundaries, delivering better representations in more complex scenarios and limited feature spaces.

This paper is structured as follows: Section \ref{sec:related_work} reviews the related work on prototype-based learning and discriminative learning approaches. Section \ref{sec:preliminaries} provides preliminary background on the concepts used throughout the paper. Section \ref{sec:proposed_method} introduces a unified representation for the roles of class centers and weight vectors, which we call the Deep Positive Prototype (DPP) model. This is then further extended to the DPNP model by integration of negative prototypes. Section \ref{sec:experiments} discusses the experimental setup and results, demonstrating the effectiveness of our approach on well-known image classification datasets. Finally, Section \ref{sec:conclusion} concludes the paper and highlights potential directions for future research.

The contributions of this paper are summarized as follows:
\begin{enumerate}
\item{{\bf{Unified Prototype and Weight Representation:}} We propose a unified approach that shares and aligns class weights and prototypes, providing a consistent representation for both classification and discrimination.}
\item{{\bf{Negative Prototype Integration:}} We introduce prototypes of neighboring rival classes as the negative prototypes for the current class. Therefore, in contrast to our previous work \cite{Zarei}, these negative prototypes are not stored separately, eliminating the need for extra memory or dedicated learning procedure. The repulsion from negative prototypes coincides with the inter-class separation term and completes the discriminative learning aspect.
}
\item{{\bf{Reduction of Feature Space Dimensionality:}} Benefiting discriminative terms in loss function based on the introduced positive and negative prototypes, and the proposed weight/center unification, our model is able to establish a structured geometry and regularity in the latent space which let us obtain competitive results in much lower dimensions using smaller networks. This in turn improves generalization which is studied on several benchmark datasets.}
\end{enumerate}

The proposed approach aims to bridge the gap between PbL and discriminative learning, offering a compelling solution to the challenges of modern representation learning in deep neural networks \cite{Bengio2013}.

\section{Related Work}
\label{sec:related_work}
PbL is a classification approach where classes are represented by prototypes, input samples are then classified based on their distance to these prototypes. Traditional methods like k-Nearest Neighbors (k-NN) and Learning Vector Quantization (LVQ) established the foundation for PbL in the 1990s. While k-NN classifies samples based on their proximity to the nearest neighbors (data points), LVQ refines prototypes to better capture class distributions, thereby reducing storage space and improving computational efficiency \cite{Kohonen1990,Kohonen1998}.

Recent advancements have integrated PbL with deep learning. For instance, Prototypical Networks \cite{Snell2017} use the mean of feature vectors as prototypes, which may be acceptable in some few-shot learning scenarios, but can lead to outdated and ineffective prototypes due to infrequent updates. 
Also, some methods \cite{Chen2022, Mettes2019} define fixed prototypes based on latent space geometries, aiming to maximize the cosine similarity of each feature vector with their corresponding prototypes. Furthermore, Oyedotun and Khashman \cite{Oyedotun2018} incorporated prototypes into neural networks to enhance emotional recognition, and Li et al. \cite{Li2018} introduced a network structure that includes a unique prototype layer, enabling the network to provide explanations for its predictions. The interpretability of such models is a significant advantage, particularly in fields requiring transparent decision-making. However, they often exhibit lower accuracy compared to state-of-the-art methods. 

Discriminative learning approaches on the other hand, focus on finding representations that maximize class separability in the feature space. Popular techniques include Linear Discriminant Analysis (LDA), contrastive learning approaches, and various neural network models with loss functions inspired by LDA terms. LDA and its variants seek to maximize the ratio of between-class variance to within-class variance, creating a linear combination of features that properly separates different classes \cite{Chang2016}. However, LDA is limited to linear transformations and is not suitable for complex data distributions. Accordingly, there is a relatively long history of research that attempts to combine LDA with neural networks \cite{Mao1993}. In this context, a deep belief network using generalized discriminant analysis was introduced to enhance feature extraction capabilities \cite{Stuhlsatz2012}. Also, Dorfer et al. proposed Deep Linear Discriminant Analysis, which combines LDA with deep neural networks (DNNs) to learn non-linear feature transformations while maintaining the discriminative power of LDA \cite{Dorfer2016}. Other approaches, such as Regularized Deep Linear Discriminant Analysis, aim to improve the stability and performance of these models by incorporating regularization terms to handle within-class scatter matrices \cite{Lu2022}. However, these methods require either the computation of eigenvalues or the inversion of scatter matrices, making them computationally expensive facing high-dimensional data scenarios. Additionally, when implemented as DNNs, they complicate the backpropagation algorithm.

Another prominent discriminative approach is the use of contrastive learning, which involves training models to distinguish between similar and dissimilar pairs of data points. This approach is mainly developed for unsupervised learning \cite{Chen2020,He2020,Li2021}, though methods like triplet loss \cite{Schroff2015} N-pair loss \cite{Sohn2016}, and supervised contrastive learning \cite{Khosla2020} have been proposed for supervised settings. The major issue with these models is computational complexity since input data is processed in terms of pairwise relations between learning samples. Therefore, instance-based contrastive learning is not suited for large-scale deep learning tasks.

The well-known Cross-Entropy (CE) loss, widely used for classification tasks in DNNs, acts as a discriminative term by optimizing the output of the softmax function \cite{Pang2019, He2016, Gu2018}. Although CE loss pushes different classes away from each other, it does not try to explicitly gather samples of the same class together by decreasing intra-class distances, nor does it create any safe margins between classes to enhance generalization \cite{Wang2017, Liu2017, Wen2019}. To compensate for these shortcomings, center-based and margin-based approaches are introduced and studied particularly in the face recognition applications.
Liu et al. \cite{Liu2016} proposed the large-margin softmax loss, which adds an angular margin into CE loss using cosine similarity. This idea has been extended in subsequent works such as SphereFace \cite{SphereFace2017}, CosFace \cite{Wang2018}, and ArcFace \cite{Deng2022}, which have achieved state-of-the-art performance in face recognition. The need for proper adjustment of the added margin hyperparameters and the lack of attention to intra-class variance are disadvantages of such methods.

Center-based methods are the closest PbL counterparts in DNNs. 
The Center Loss (CL) function \cite{Wen2016} is a key example, designed to minimize the distance between deep features and their corresponding class centers, thereby promoting intra-class compactness while leaving the responsibility of class separation to the CE term. Variants like Range Loss \cite{Zhang2017} and Island Loss \cite{Cai2018} leverage this concept while addressing challenges such as handling data imbalances and enhancing distinct class boundaries. 
As a more explainable alternative, Convolutional Prototype Network (CPN) \cite{Yang2022} uses a loss based on Euclidean distance to align features with their class centers for open set recognition. However, the reported accuracies drop by a few percent as learning becomes trickier with Euclidean distance \cite{Zarei, Yang2022}.
Constrained Center Loss (CCL) \cite{Shi2023} is another step forward by adding a center absorbing term alongside the CE loss. CCL explicitly learns class centers as averages of class samples in the feature space at the end of every other epoch, while keeping these centers and classifier weights constrained on hyperspheres by length-normalizing them.
Overall, most of the aforementioned center-based methods use separate parametrizations in one way or another for representing class centers and classifier weights which sometimes are even treated with different learning algorithms. 

In this paper, we start by first unifying this problematic separate parametrization into a single set of shared parameters with the two roles of inner product-based classification and prototype learning. Then, we propose additional novel terms for the loss function to formulate a discriminative learning system based on the concept of positive and negative prototypes. In this way, our proposed model is not only able to compensate for the common accuracy drop observed in PbL but also can achieve comparable performance in low-dimensional latent spaces.

\section{Preliminaries}
\label{sec:preliminaries}
To facilitate understanding of the proposed model, we first present the essential mathematical foundations and introduce the notations used throughout the paper. Assuming $x_i\in\mathbb{R}^D$ represents the $i^{\text{th}}$ input data sample, where $D$ is the dimensionality of the input space. We denote the feature representation of $x_i$ in the latent space as $h(x_i;\theta)\in\mathbb{R}^d$, where $\theta$ is the parameters  of the deep neural network and $d$ is the dimensionality of the latent space. For a classification problem with $M$ classes, the network has $M$ neurons in the output layer. Let $z_{ij}= w_j^T h(x_i; \theta)$ be the output of the $j^{\text{th}}$ neuron in the output layer for the $i^{\text{th}}$ sample, where $w_j\in\mathbb{R}^d$ is the weight vector associated with the $j^{\text{th}}$ class. Applying the softmax function to the scaled outputs $z_{ij}/\alpha$, the predicted probability distribution ${\hat{y}}_i$ over the classes is obtained as:
\begin{equation}
\label{eq1}
\hat{y}_{ij} = \frac{e^{z_{ij}/\alpha}}{\sum_{k=1}^{M} e^{z_{ik}/\alpha}}, \quad \text{for } j = 1, \dots, M
\end{equation}
The network is trained by minimizing CE loss over the entire training set consisting of $N$ samples. The cross-entropy loss function is defined as:
\begin{equation}
\label{eq3}
L_{\text{CE}} = -\frac{1}{N} \sum_{i=1}^N \sum_{j=1}^M \mathbbm{1}\{y_i=j \} \log \hat{y}_{ij}
\end{equation}
where $\mathbbm{1}{\{ . \}}$ is the indicator function, while $y_i$ and ${\hat{y}}_{ij}$ denote the true class label and the predicted probability of class $j$ for sample $x_i$, respectively. 
The CE loss function forms a probabilistic framework that penalizes incorrect classifications to adjust network weights toward improving classification accuracy in subsequent iterations.

As discussed in the related work section, to enhance the clustering of features within each class and improve classification accuracy, some methods combine CE loss with additional terms that enforce feature compactness and inter-class separation. A common approach is to use $c_{y_i}\in\mathbb{R}^d$ as the center point of class $y_i$, and the distance between the feature vectors and their corresponding class centers is minimized using the following extra term:
\begin{equation}
\label{eq5}
L_{\text{center}} = \frac{1}{2N} \sum_{i=1}^{N} \left\| h(x_i; \theta) - c_{y_i} \right\|^2_2
\end{equation}
This term reduces intra-class variations by pulling the feature vectors $h\left(x_i;\theta\right)$ towards their corresponding class centers in the latent space. To achieve a balance between classification accuracy and feature compactness, CL is then defined as:
\begin{equation}
\label{eq6}
L_{\text{CL}} = L_{\text{CE}} + \lambda_{\text{center}} L_{\text{center}}
\end{equation}
where $\lambda_{\text{center}}$ is the hyperparameter that controls the aforementioned trade-off and its value is typically selected through cross-validation.

\section{Proposed method}
\label{sec:proposed_method}
Although rarely discussed in detail, some researchers implicitly assume that the class centers are captured by the classifier weights. However, this assumption has been challenged in multiple studies \cite{Kansizoglou2022, Zarei}. Specifically, due to the stochastic nature of weight initialization and the optimization process during training, weight vectors $w_j$ frequently do not align with the geometric centers of their corresponding classes in the feature space, particularly under conventional training procedures. This mismatch between classifier weights and geometric class centers is frequently ignored \cite{Kansizoglou2022}, but it can lead to potential stability issues during DNNs training, resulting in suboptimal feature representations. 
The aforementioned mismatch becomes more evident when additional variables are introduced to represent class centers, such as in the discriminative terms of the loss function \cite{Shi2023}.

As discussed in our previous study \cite{Zarei}, the weights of neurons in single-layer winner-take-all (WTA) networks generally should be interpreted as a combination of positive and negative prototypes within each class, rather than as simple geometric centers. This interpretation naturally explains the misalignment between weight vectors and the actual class centers, especially in the input space for single-layer WTA networks. This finding led us to add a repulsive term to the loss function, resembling a negative prototype, which will be discussed shortly. 

In the following, we explain our method, which innovatively unifies classifier weights with class centers in the feature space, uses these centers to induce a center tendency behavior in the feature extractor, and finally completes the discrimination between classes by considering other class centers as negative prototypes. By delineating class boundaries more accurately, this approach provides a more regular population of the feature space and leads to better generalization. We will show that the dimensionality of the feature space may dramatically be reduced while still achieving competitive results, as a consequence of the geometric regularity induced into the feature space.

\subsection{Modeling Classifier Weights and Class Centers as Shared Parameters (Prototypes)}
In previous models, the classifier weight vectors $w_j\in\mathbb{R}^d$ in the final layer of the network and the class centers $c_j\in\mathbb{R}^d$ in the feature space are typically considered separately and kept in distinct variables which usually do not coincide. The weight vector $w_j$ is used for the decision-making process during classification, typically based on the inner product similarity, while the center $c_j$ is used to initiate a central tendency on the feature space representations of the corresponding class samples, usually based on Euclidean distance. However, apart from the challenges in determining how to learn the centers, this distinct parametrization for $w_j$ and $c_j$ can lead to inconsistencies during training; since the decision boundaries defined by the weight vectors $w_j$ may not optimally align with the clusters of data points around the centers $c_j$ in the feature space \cite{Kansizoglou2022}. This phenomenon causes problems when classifier weights are meant to be used as class prototypes \cite{Zarei}.

To address this issue, we propose to share the parameters between class centers and weight vectors; see Figure \ref{figure_1}. This allows the classifier weights to play a double role and also serve as class centers, thereby streamlining the learning process and improving consistency. This parameter sharing not only smooths the training process by eliminating the need for a separate learning algorithm \cite{Shi2023}, but also provides the necessary terms for a discriminative loss function, which is introduced in the following subsections.

\subsection{Deep Positive Prototype Model}
\begin{figure}[!t]
\centering
\includegraphics[width=3.4in]{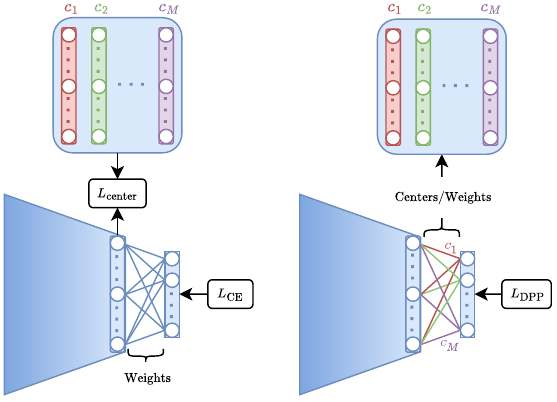}
\caption{Adding discriminative terms to the loss function of deep neural networks:
On the left, a commonly used approach is depicted where class centers and classifier weights are separately stored and learned. On the right, our proposed model is shown that unifies storage and learning of the weights and the centers. Here, each color represents a specific class center or the classifier weight vector associated with the corresponding neuron.}
\label{figure_1}
\end{figure}

Given the insights discussed above, we propose a shared parametrization that unifies the class centers and the weight vectors into a single entity. This unification not only simplifies the model and avoids instabilities, but also improves the performance, increases the interpretability, and supports discriminative terms in the loss function; see Figure \ref{figure_1}.

Formally, each class center $c_j$, lying on a hypersphere with radius $\alpha$, is designed to serve both as the decision boundary (i.e., the weight vector of the neuron $j$ in the classifier) and the representative prototype of the class which can be considered as its center point. The obtained model is then referred to as the {\bf{Deep Positive Prototype (DPP)}} model.
The learning objective in the DPP model thus consists of the usual CE term along with a central tendency term to adjust the weight vectors, ensuring that they represent the class centers while optimally positioned in the feature space, thereby enhancing the model's ability to classify new instances based on their similarity to these prototypes. Formally:
\begin{equation}
\label{eq8}
\begin{aligned}
L_{\text{DPP}} = & -\frac{1}{N} \sum_{i=1}^{N} \log \frac{e^{c_{y_i}^T h(x_i; \theta)/\alpha}}{\sum_{j=1}^{M} e^{c_j^T h(x_i; \theta)/\alpha}} \\
& + \frac{\lambda_{\text{pos}}}{2N} \sum_{i=1}^{N} \left\| h(x_i; \theta) - c_{y_i} \right\|^2_2
\end{aligned}
\end{equation}
where $c_{y_i}$ serves as both the weight vector in CE loss and the center (positive prototype) for class $y_i$, and $\lambda_{\text{pos}}$ has the same role as $\lambda_{\text{center}}$ in \eqref{eq6}. This unified representation aims to ensure that feature vectors $h\left(x_i;\theta\right)$ are tightly clustered around their corresponding DPPs. In fact, DPPs are simultaneously points representing class centers in the feature space and also weight vectors that form class boundaries via the inner product similarity in classifying neurons. This dual role ensures that decision boundaries are directly adapted to the aggregations of class members, which helps to achieve compact and well-separated feature clusters in the latent space. Furthermore, to ensure numerical stability during training and receiving proper gradient magnitudes from CE, we renormalize the prototypes at the start of each epoch, to keep them on the surface of a hypersphere with radius $\alpha$ as a hyperparameter. 

\subsection{Integration of Negative Prototypes: Deep Positive-Negative Prototype Model}

In our previous work \cite{Zarei}, we noted the presence of both positive and negative prototypes during the training process of a single-layer WTA networks. Positive prototypes are derived from data within the same class, while negative prototypes are separately learned from data of the other classes proportional to the mount of their activation. However, an interesting observation was made during the final stages of learning \cite{Zarei}, when the training was almost complete, the negative prototypes closely resembled positive prototypes in the MNIST dataset \cite{Mnist} or bared some similarity to the opposite gender in the FERET dataset \cite{Feret}. It seems that the reason behind these observations is that the training data from other classes, which occasionally activate wrong neurons, are quite similar to them. Such cases of wrong activation generate proportionally strong feedback in the learning of negative prototypes. Therefore, for interpretability purposes in single-layer networks, in addition to customization of the learning algorithm, it is essential to maintain two independent sets of weights—positive and negative prototypes—for each class, despite the additional memory overhead. This necessity arises from the fact that the input space cannot be modified in single-layer networks, and the positive and negative prototypes are the only adaptable network parameters which are correspondingly updated.

However, in networks with more than one layer, where the feature space can be modified, it is no longer necessary to maintain two independent sets of prototypes. Instead, the prototypes of other neighboring classes can naturally serve as negative prototypes for the target class. This aligns with our previous observation that negative prototypes resembled positive prototypes of nearby rival classes. This approach leverages the flexibility of the feature space in deep networks to achieve more efficient representation learning. Meanwhile, in a discriminative learning process, this reuse of rival class DPPs as negative prototypes makes it much easier to add additional terms to the loss function for intra-class compactness and inter-class separation. 

Hence, the nearest DPP from a different class is considered as a negative prototype serving as a critical reference to push the feature space representation of the input away.
Formally, let $c_{y_i}$ denote the positive prototype for the input sample $i$. The corresponding negative prototype $c_{i}^\text{neg}$ is determined by identifying the nearest class center of a different class to the sample $i$, computed as:
\begin{equation}
\label{eq10}
c_i^{\text{neg}} \leftarrow c_{j^*}, \quad \text{where} \quad j^* = \underset{j \ne y_i}{\operatorname*{arg\,min}} \left\| h(x_i; \theta) - c_j \right\|
\end{equation}
This choice of the nearest class center ensures that the negative prototype of each training sample helps to organize feature space representation of input and distinguish nearby instances from the neighboring class properly. The effect of this negative prototype is incorporated into the model through sample negative prototype loss, $L_\text{neg}^{\text{sample}}$:
\begin{equation}
\label{eq11}
L_{\text{neg}}^{\text{sample}} = -\frac{1}{2N} \sum_{i=1}^{N} \left\| h(x_i; \theta) - c_{i}^{\text{neg}} \right\|^{1/2}_{1/2}
\end{equation}
which penalizes the proximity of feature vectors to negative prototypes, encouraging feature vectors to be pushed away from the incorrect class centers. The proper norm for this penalization should have an exponent smaller than one, since in contrast to pulling toward positive prototypes as centers which has to create a bigger pull on the further samples; here we aim to push more when the negative prototype is nearer. Therefore, instead of the usual square norm, $L_{1/2}$ is used.

Additionally, to distinguish between similar yet distinct classes, another repulsive term is introduced to move each class center away from its nearest DPP as the negative prototype of the current class. This ensures greater separation between class centers, preventing them from clustering too closely, which could blur the decision boundaries between classes. The nearest center $c_j^{neg}$ for class $j$ is obtained as:
\begin{equation}
\label{eq12}
c_j^{\text{neg}} \leftarrow c_{k^*}, \quad \text{where} \quad k^* = \underset{j \ne k}{\operatorname*{arg\,min}} \left\| c_j - c_k \right\|
\end{equation}
To enforce separation between DPPs, we introduce the class negative prototype loss $L_{\text{neg}}^{\text{class}}$:
\begin{equation}
\label{eq13}
L_{\text{neg}}^{\text{class}} = -\frac{1}{2M} \sum_{j=1}^{M} \left\| c_j - c_j^{\text{neg}} \right\|^{1/2}_{1/2}
\end{equation}
which again using $L_{1/2}$ reduces the impact of far DPPs while strongly penalizing the smaller distances. 

Combining the introduced loss terms together (\ref{eq8},\ref{eq11},\ref{eq13}) into a single loss function, we obtain the {\bf{Deep Positive-Negative Prototype (DPNP)}} model which is encapsulated in the architecture of conventional neural networks without any need for extra parameters. 
In this model, both positive and negative prototypes play a critical role in refining classification boundaries. The positive prototypes pull the feature representations toward the center of the correct classes, promoting intra-class compactness. Meanwhile, the negative prototypes push them away from each other, leading to better separation and more regularity in the feature space. The total loss function of the DPNP model is given as:
\begin{equation}
\label{eq14}
\begin{aligned}
L_{\text{DPNP}} = & \, L_{\text{DPP}} + \lambda_{\text{neg}}^{\text{sample}} L_{\text{neg}}^{\text{sample}} + \lambda_{\text{neg}}^{\text{class}} L_{\text{neg}}^{\text{class}} \\
= & -\frac{1}{N} \sum_{i=1}^{N} \log \frac{e^{c_{y_i}^T h(x_i; \theta)}}{\sum_{j=1}^{M} e^{c_j^T h(x_i; \theta)}}  \\
& + \frac{\lambda_{\text{pos}}}{2N} \sum_{i=1}^{N} \left\| h(x_i; \theta) - c_{y_i} \right\|^2_2 \\
& - \frac{\lambda_{\text{neg}}^{\text{sample}}}{2N} \sum_{i=1}^{N} \left\| h(x_i; \theta) - c_{i}^{\text{neg}} \right\|^{1/2}_{1/2} \\
& - \frac{\lambda_{\text{neg}}^{\text{class}}}{2M} \sum_{j=1}^{M} \left\| c_j - c_j^{\text{neg}} \right\|^{1/2}_{1/2}
\end{aligned}
\end{equation}
where $\lambda_{\text{neg}}^{\text{sample}}$ and $\lambda_{\text{neg}}^{\text{class}}$ are regularization parameters that control the trade-off between ensuring center separation and the primary classification task. Different stages of the learning process for this model are given in Algorithm \ref{alg:DPNP}.

\begin{algorithm}[!t]
\caption{Training Algorithm for DPNP Model}\label{alg:DPNP}
\begin{algorithmic}[1]
\Require Training data $\mathcal{D}=\{(x_i, y_i)\}_{i=1}^N$, number of epochs $E$, batch size $B$, network and classifier learning rates $\eta$ and $\eta_c$, hyperparameters $\lambda_{\text{pos}}$, $\lambda_{\text{neg}}^{\text{sample}}$, $\lambda_{\text{neg}}^{\text{class}}$, radius $\alpha$.

\State Initialize network parameters $\theta$ and class centers $\{c_j\}_{j=1}^M$
\For {$\text{epoch} = 1$ to $E$}
    \For{each class $j = 1, \dots, M$}
        \State Normalize class prototypes: $c_j \leftarrow \alpha \frac{c_j}{\| c_j \|}$
    \EndFor
    \For{each mini-batch $\mathcal{B}=\{(x_i, y_i)\}_{i=1}^B \subset \mathcal{D}$}
        \For{each sample $(x_i, y_i)$ in the mini-batch $\mathcal{B}$}
            \State Compute feature representation $h(x_i; \theta)$ 
            \State Find the negative prototype $c_i^{\text{neg}}$ using \eqref{eq10}
        \EndFor
        \For{each class $j = 1, \dots, M$}
          \State Find class negative prototype: $c_j^{\text{neg}}$ using \eqref{eq12}
        \EndFor
        \State Compute the loss function $L_{\text{DPNP}}$ based on \eqref{eq14}
        \State Update network parameters:
         $\theta \leftarrow \theta - \eta \frac{\partial L_{\text{DPNP}}}{\partial \theta}$
        \For{each class $j = 1, \dots, M$}
        \State Update unified classifier/class prototype:
        \State $c_j \leftarrow c_j - \eta_c \frac{\partial L_{\text{DPNP}}}{\partial c_j}$
        \EndFor
    \EndFor
\EndFor
\end{algorithmic}
\end{algorithm}

In \cite{Zarei}, we used Euclidean distance for the exponential term in the softmax function of CE loss to maintain consistency with the prototype approach. We observe that this choice made learning more sensitive and trickier; hence, we substitute inner product similarity for the exponential term in place of Euclidean distance in our current research to remain more consistent with the mainstream neural network learning and improve stability. Depending on the nature of data at hand, it is evident that each term in the proposed loss function can be substituted with a better distance metric whenever possible.

\section{Experiments}
\label{sec:experiments}
The experiments in this study are designed to assess the performance and efficiency of the proposed models particularly in the context of image classification tasks. We selected three widely recognized datasets—CIFAR-10 \cite{Cifar}, CIFAR-100 \cite{Cifar} and Flower-102 \cite{Flower}—to evaluate the model across different levels of complexity and diversity in visual features. We report classification accuracy as the primary performance metric, supplemented by an analysis of intra-class compactness, inter-class separability and the balance between them.

We evaluate our approach by comparing its performance with baseline models using the standard Cross Entropy (CE) \cite{He2016} and Center Loss (CL) \cite{Wen2016} models, in addition to several state-of-the-art methods such as Convolutional Prototype Network (CPN) \cite{Yang2022}, Constrained Center Loss (CCL) and Simplified Constrained Center Loss (SCCL) \cite{Shi2023}. Reported results demonstrate better accuracy, improvement in compactness and separability in feature space representations, and possibility to work with lower dimensions.

In Section \ref{sub:Datasets}, we introduce the datasets used in our experiments, followed by the architectural details and experimental setup in Section \ref{sub:Model_Arch}. Section \ref{sub:Classification_Accuracy} is dedicated to comparing our method with existing approaches, providing a detailed analysis of how our modifications offer a competitive yet simpler alternative. Section \ref{sub:Separability_Compactness} further examines inter-class separability and intra-class compactness through quantitative analyses and visualizations, illustrating the effectiveness of our approach in different dimensionalities.

\subsection{Datasets}
\label{sub:Datasets}
The CIFAR-10 dataset consists of 60,000 color images of everyday objects with a size of 32x32 in 10 different classes, with 6,000 images per class. The dataset is divided into 50,000 training images and 10,000 test images.

CIFAR-100 is an extension of CIFAR-10, containing the same number of images but divided into 100 classes, each containing 600 images. The train-test split remains the same as CIFAR-10, and it is usually considered a tougher benchmark for classification because of its larger variety of categories.

The Flower-102 dataset consists of 8,189 images of 102 different species of flower as its classes, covering various scales, poses, and lighting conditions. This dataset is particularly challenging due to the subtle differences between classes, requiring the model to effectively capture fine-grained features for accurate classification. 

\subsection{Model Architecture and Experimental Setup}
\label{sub:Model_Arch}
\begin{figure}[!t]
\centering
\includegraphics[width=3.45in]{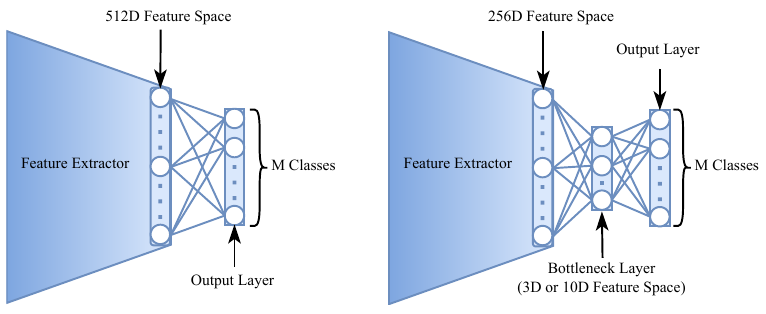}
\caption{Left: The standard ResNet18 architecture with 512D feature space and about 11 million parameters. Right: The reduced ResNet18 architecture with 256 filters in the final convolutional layer, bottlenecked at a (3D or 10D, based on the dataset) feature space and about 5 million parameters. This reduced architecture is used to study the regularity of feature space.}
\label{fig_architecture}
\end{figure}

In this study, we use the popular ResNet18 \cite{He2016} architecture with 512-dimensional output as the feature extractor followed by a fully connected layer with one neuron for each class. This common architecture contains more than 11 million parameters, and the final 512-dimensional feature space is unnecessarily large for simpler datasets, especially with a limited number of classes (e.g., CIFAR-10 or CIFAR-100), potentially leading to over-parametrization and poor generalization. Therefore, to study the effect of injected regularity into the feature space by our unified prototypical discriminative learning approach, we also use a reduced version of ResNet18, and compare the effect of dimensionality drop on the results. Specifically, we reduce the number of filters in the final convolutional layer from 512 to 256, then add a linear layer to map it to a lower-dimensional space, 3D in the case of CIFAR-10 and 10D for the other more complex datasets. The following classifier layer in this reduced architecture is the same as the standard case. This simple modification reduces the number of network parameters to less than half ($\sim$5 million). Figure \ref{fig_architecture} illustrates the standard and reduced ResNet18 architectures.

The training process is performed with standard data augmentation techniques, including random cropping, horizontal flipping, and normalization, utilizing the Stochastic Gradient Descent (SGD) optimizer, with momentum set to 0.9 and a weight decay of 5e-4. The hyperparameter $\alpha$, representing the radius of the hypersphere, is fixed at 40 for all experiments. Additionally in the standard ResNet18 experiments, the hyperparameters $\lambda_{\text{pos}}$, $\lambda_{\text{neg}}^{\text{sample}}$, and $\lambda_{\text{neg}}^{\text{class}}$ are all set to 0.1. However, based on the dimensionality in the case of the reduced ResNet18, they are readjusted to 0.01, 0.01, and 0.02 respectively for CIFAR-10, and to 0.1, 0.1, and 0.2 for the other two datasets. The experiments on two CIFAR datasets are started from random initializations with an initial learning rate of 0.1 for both $\eta$ and $\eta_c$, and a scheduled reduction factor of 10 at 25\%, 50\%, and 75\% of the total training epochs. For the more complex Flower-102 dataset, the standard ResNet18 pretrained on ImageNet is used and the initial network learning rate ($\eta$) is reduced to 0.01 and the first scheduled decrease point is removed. The modified layers of the reduced ResNet18 architecture are also pretrained through a warm-up period of 10 epochs while the unchanged layers are kept frozen. The main training phase consists of 200 epochs with a batch size of 64 which is reduced to 160 epochs when pretrained weights are used.

\subsection{Classification Accuracy}
\label{sub:Classification_Accuracy}
In this section, we provide a comparative analysis of the classification performance of our proposed models against baseline and state-of-the-art methods over the aformentioned two architectures and three datasets. 

Table \ref{tab:acc_standard_resnet18} summarizes the classification accuracies achieved by various methods using the standard ResNet18 architecture (512D feature space). Our proposed DPNP model outperforms the rest in all three datasets. 

\begin{table}[t]
\centering
\caption{Classification accuracy on CIFAR-10, CIFAR-100, and Flower-102 datasets using the standard ResNet18.}
\label{tab:acc_standard_resnet18}
\begin{tabular}{c|ccc}
\toprule
\multicolumn{1}{c|}{} & \multicolumn{3}{c}{\textbf{Dataset}} \\ 
\cmidrule(lr){2-4}
\textbf{Model}       & \textbf{CIFAR-10} & \textbf{CIFAR-100} & \textbf{Flower-102} \\ 
\textbf{}       & \textbf{512D} & \textbf{512D} & \textbf{512D} \\ 
\midrule
CE                   & 93.81\%           & 75.12\%            & 92.16\%            \\
CL                   & 93.87\%           & 75.20\%            & 93.12\%            \\
CPN                  & 92.97\%           & 75.11\%            & 89.85\%            \\
CCL                  & 94.72\%           & 77.61\%            & 94.25\%            \\
SCCL                 & 94.91\%           & 77.88\%            & 94.34\%            \\
DPP (Ours)           & \underline{95.16}\%           & \underline{78.56}\%            & \underline{94.64}\%            \\
\textbf{DPNP (Ours)}           & \textbf{95.40\%}   & \textbf{79.01\%}   & \textbf{95.18\%}    \\ 
\bottomrule
\end{tabular}
\end{table}

The standard ResNet-18 architecture using CE on CIFAR-10 achieves the reasonable accuracy of 93.81\% and incorporating CL yields a slight improvement to 93.87\%. CPN on the hand faces a slight drop in the accuracy which is common for PbL.
In this context, the proposed DPNP achieves the best results with noticeable accuracy growth of nearly 1.6\% which is relatively significant. In the second column, for the more challenging CIFAR-100 dataset, differences in the performance of rival methods are better observed. Again, using discriminative terms shows small improvement over the baseline CE for all methods, but our two models, DPP and DPNP, stand at the top. The situation is similar in the third column for Flower-102, except the accuracy drop of CPN is more noticeable. This suggests the effectiveness of DPP and DPNP models in improving PbL for fine-grained classification tasks. However, since the input images of this dataset are much higher resolution than CIFAR-100, accuracies are generally higher. 

The previous experiment is repeated once more for the reduced ResNet18 architecture, see Figure \ref{fig_architecture}, which forces the feature extractor to compress input representation into a lower-dimensional feature space (3D for CIFAR-10 and 10D for CIFAR-100/Flower-102). The results are summarized in Table \ref{tab:acc_reduced_resnet18}. Comparing corresponding numbers from Table \ref{tab:acc_standard_resnet18} and Table \ref{tab:acc_reduced_resnet18} shows small drops in accuracy for all models and datasets which is expected, as reducing the number of filters weakens the feature extractor. However, DPP and DPNP models surpass the rivals with even larger margins in such cases. In fact, as it will be shown in the following subsections, lower-dimensional feature spaces benefit much more from the regularity that our proposed models create.

\begin{table}[t]
\centering
\caption{Classification accuracy on CIFAR-10, CIFAR-100, and Flower-102 datasets using the reduced ResNet18.}
\label{tab:acc_reduced_resnet18}
\begin{tabular}{c|ccc}
\toprule
\multicolumn{1}{c|}{} & \multicolumn{3}{c}{\textbf{Dataset}} \\ 
\cmidrule(lr){2-4}
\textbf{Model}       & \textbf{CIFAR-10} & \textbf{CIFAR-100} & \textbf{Flower-102} \\ 
\textbf{}       & \textbf{3D} & \textbf{10D} & \textbf{10D} \\ 
\midrule
CE                   & 92.9\%            & 71.54\%            & 87.52\%             \\
CL                   & 92.96\%           & 68.72\%            & 89.23\%            \\
CPN                  & 90.29\%           & 66.65\%            & 82.10\%            \\
CCL                  & 92.17\%           & 71.89\%            & 86.22\%             \\
SCCL                 & 92.52\%           & 66.54\%            & 88.19\%             \\
DPP (Ours)           & \underline{93.78}\%           & \underline{73.39}\%            & \underline{91.51}\%             \\
\textbf{DPNP (Ours)}           & \textbf{94.18\%}   & \textbf{73.84\%}   & \textbf{92.19\%}    \\ 
\bottomrule
\end{tabular}
\end{table}

\subsection{Inter-Class Separability and Intra-Class Compactness}
\label{sub:Separability_Compactness}
\begin{figure*}[!t]
\centering
\subfloat{\includegraphics[width=1in]{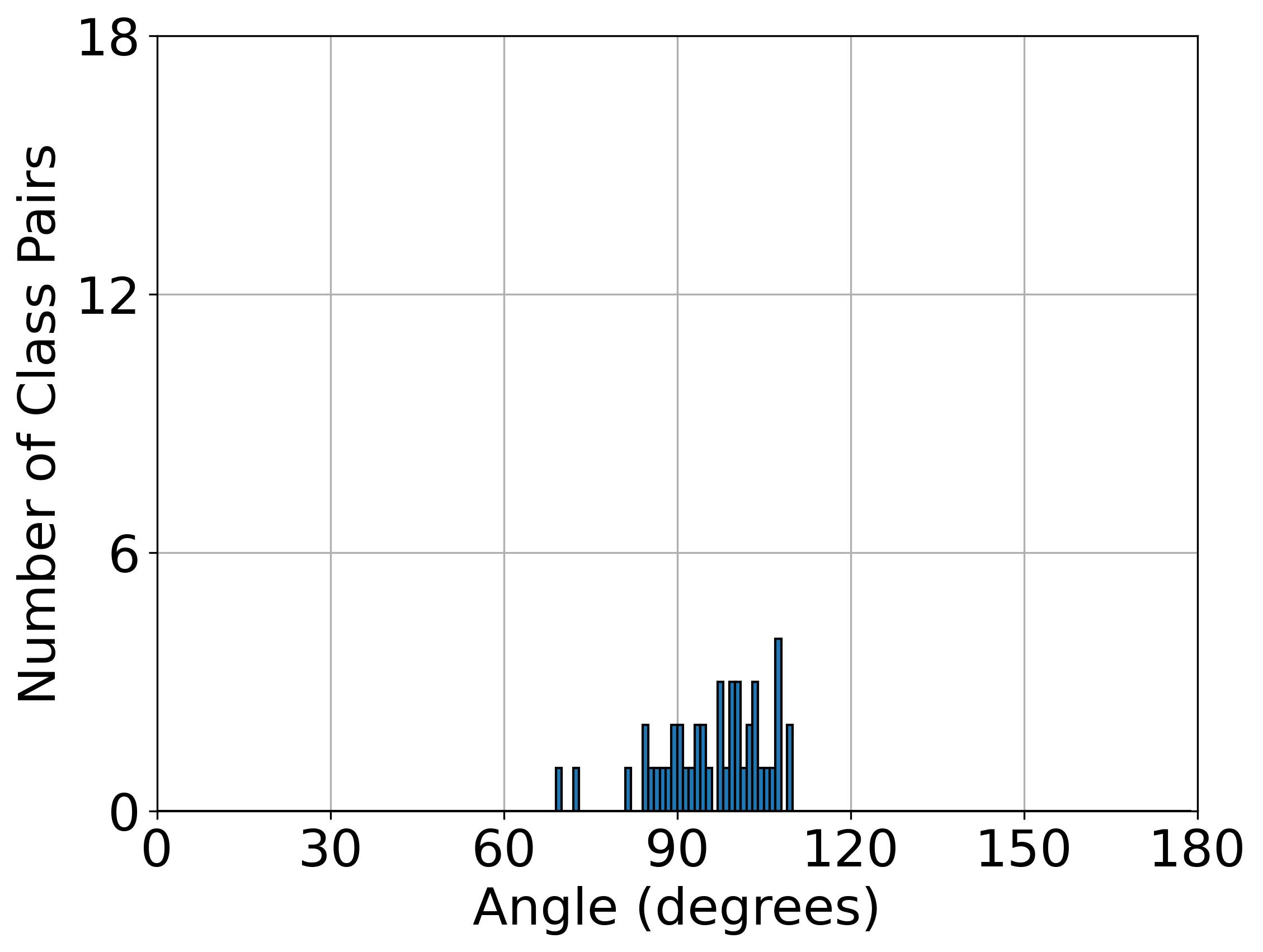}}
\hfil
\subfloat{\includegraphics[width=1in]{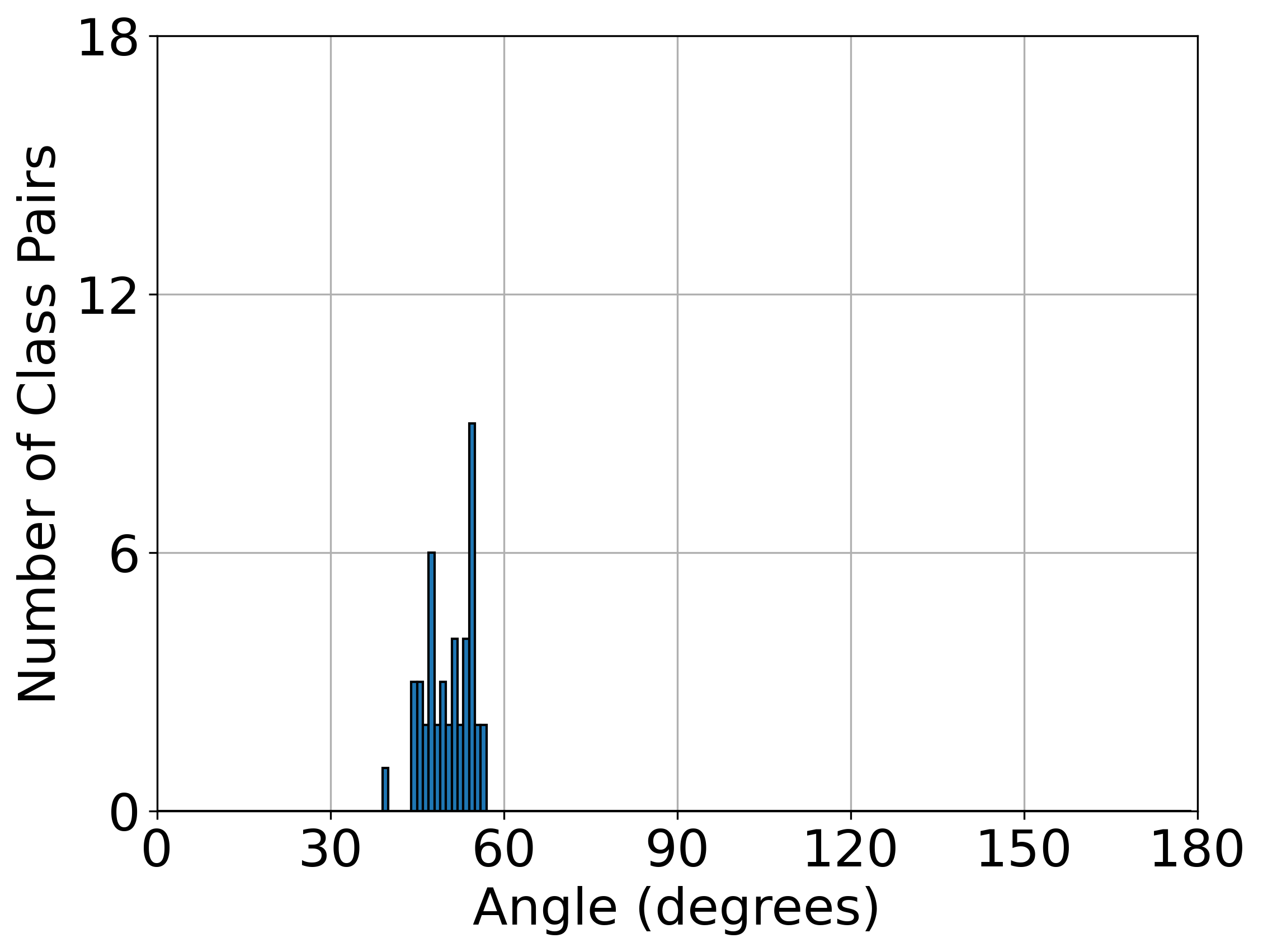}}
\hfil
\subfloat{\includegraphics[width=1in]{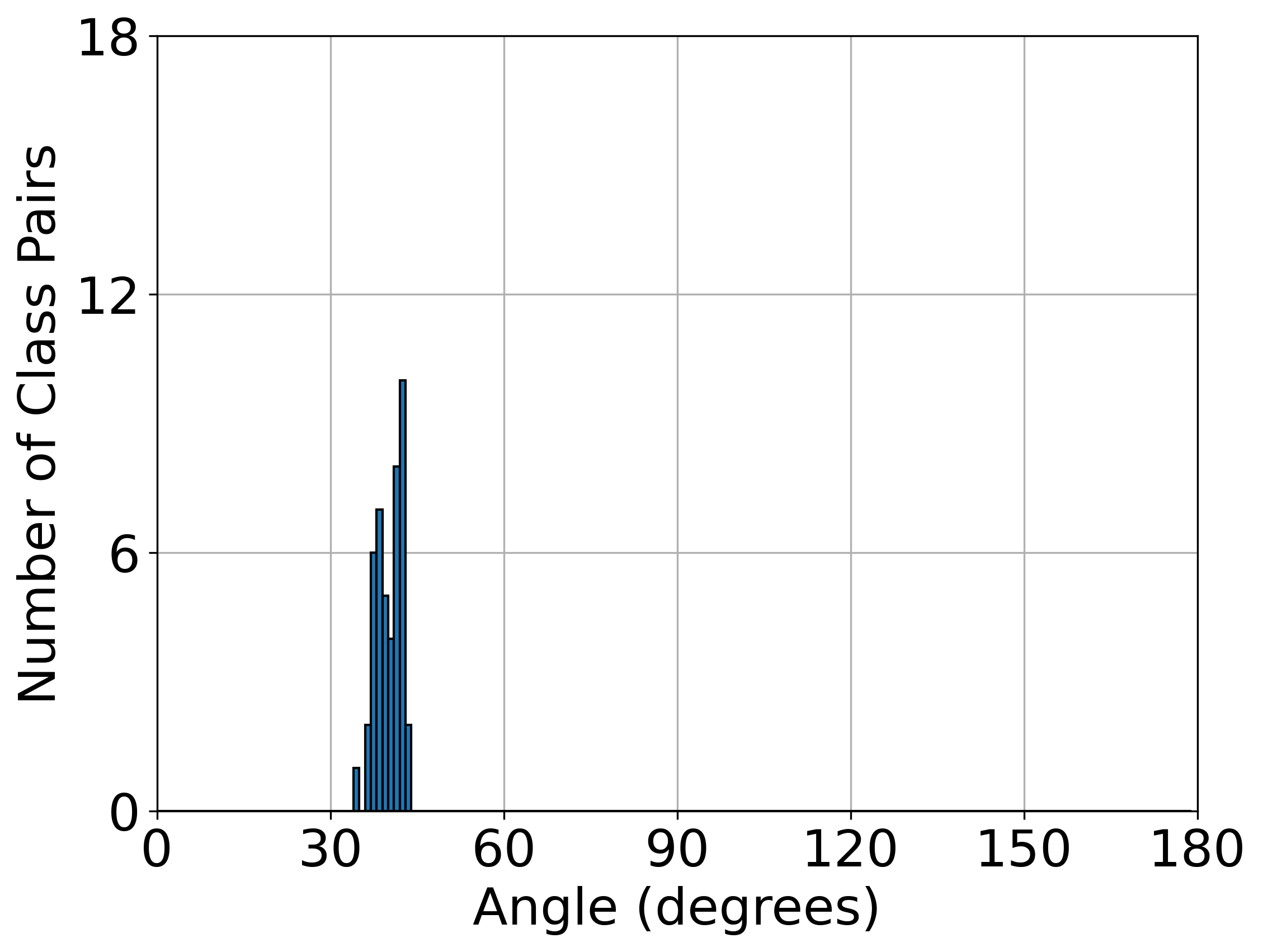}}
\hfil
\subfloat{\includegraphics[width=1in]{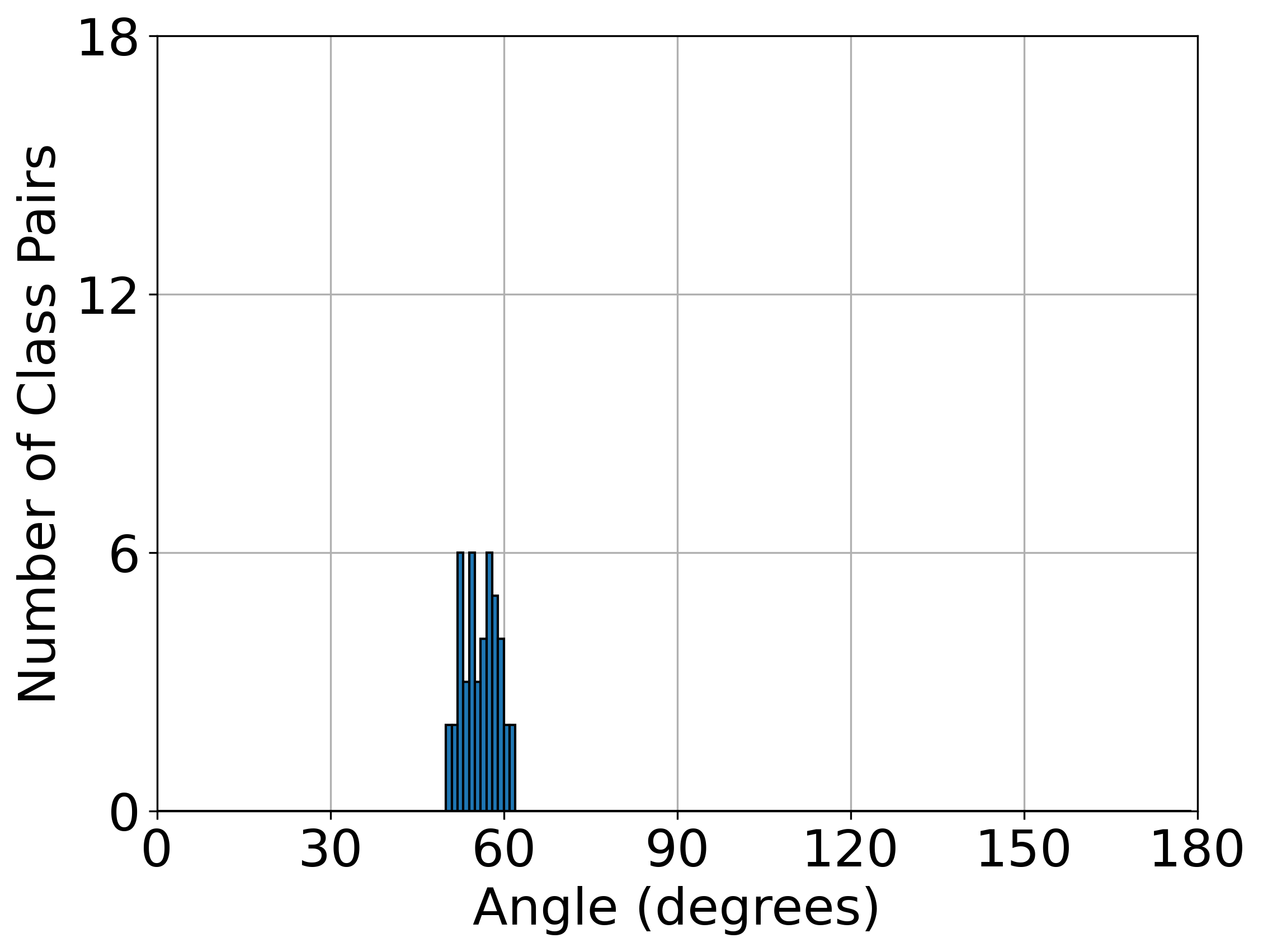}}
\hfil
\subfloat{\includegraphics[width=1in]{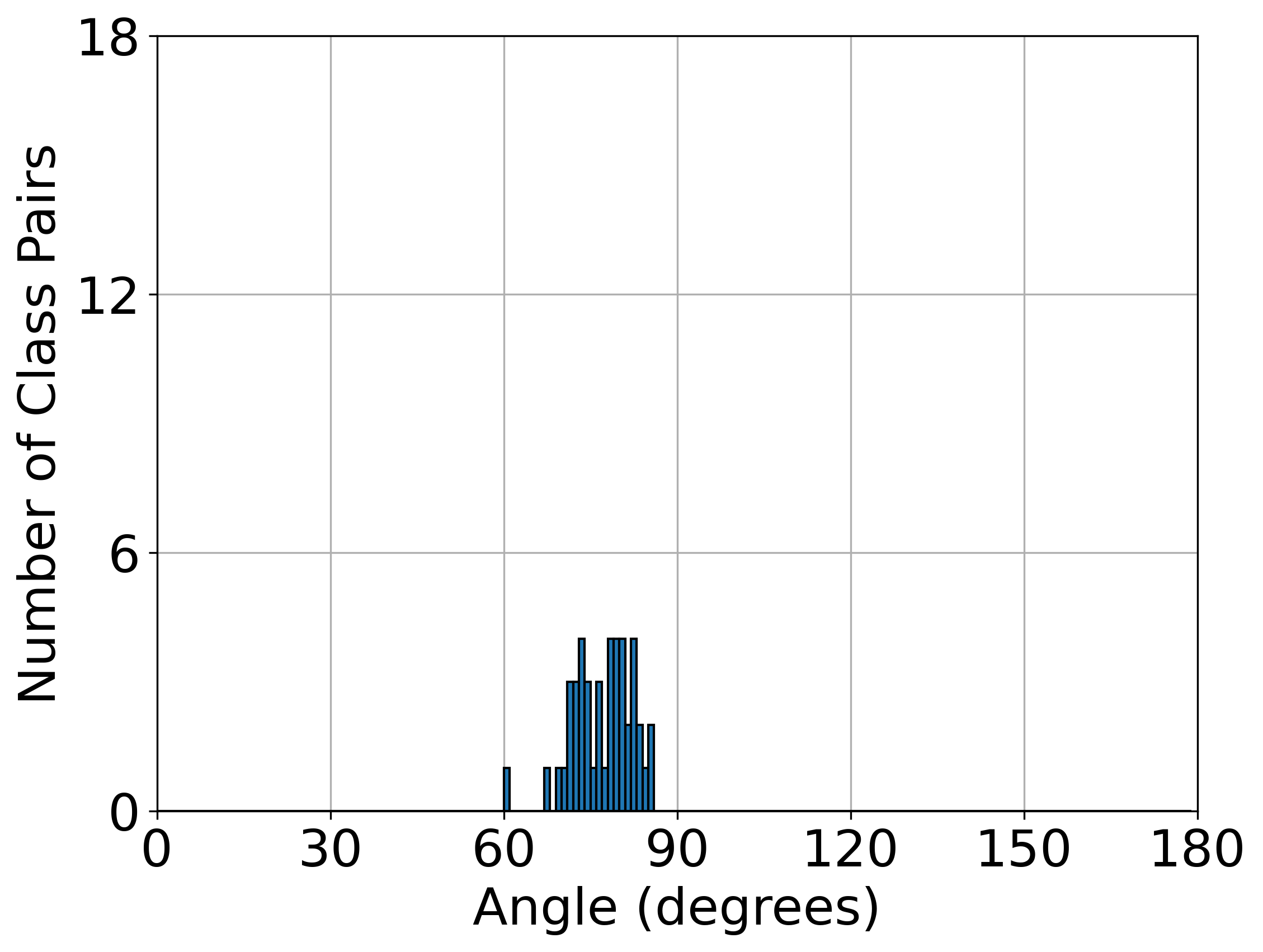}}
\hfil
\subfloat{\includegraphics[width=1in]{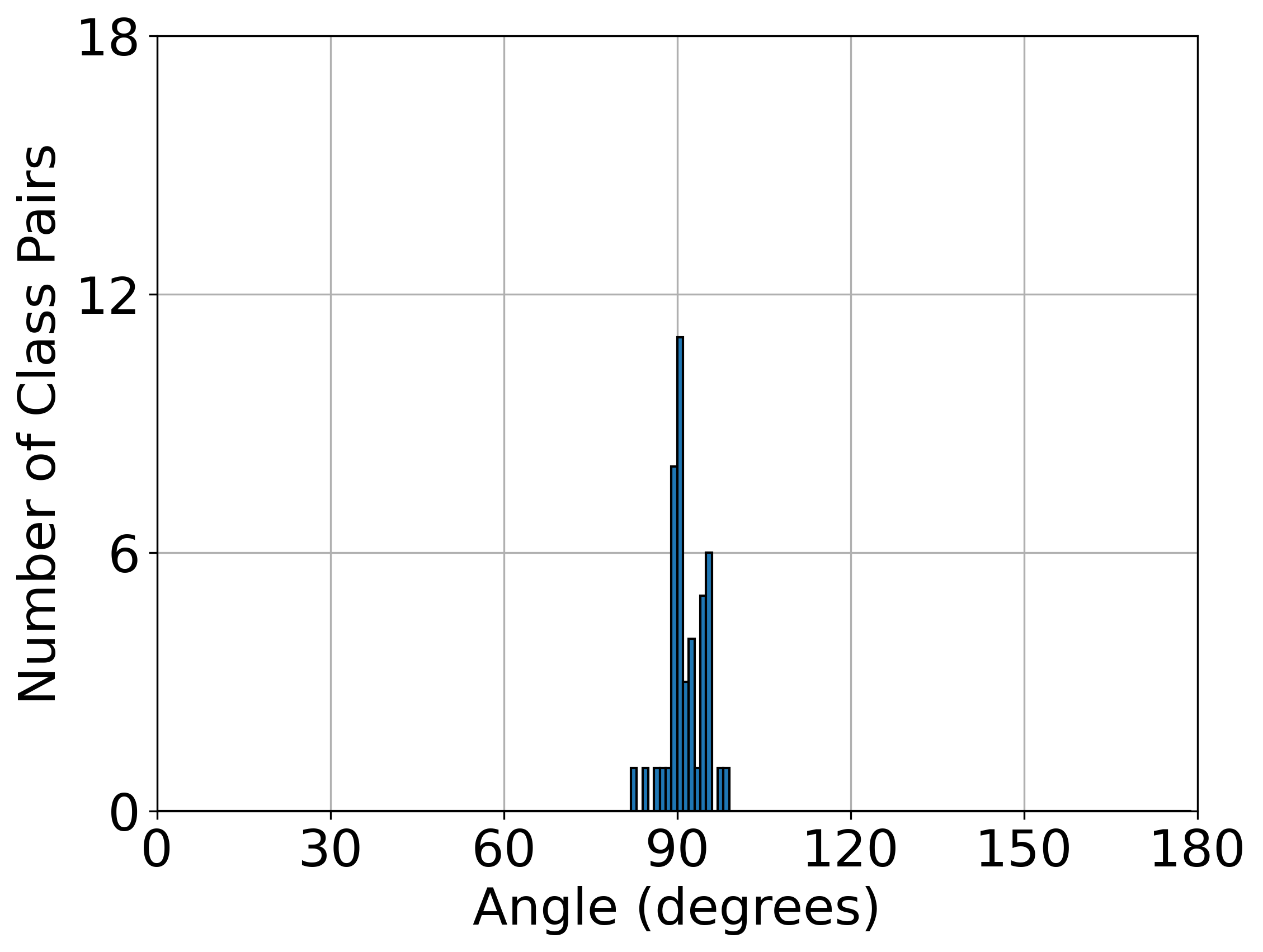}}
\hfil
\subfloat{\includegraphics[width=1in]{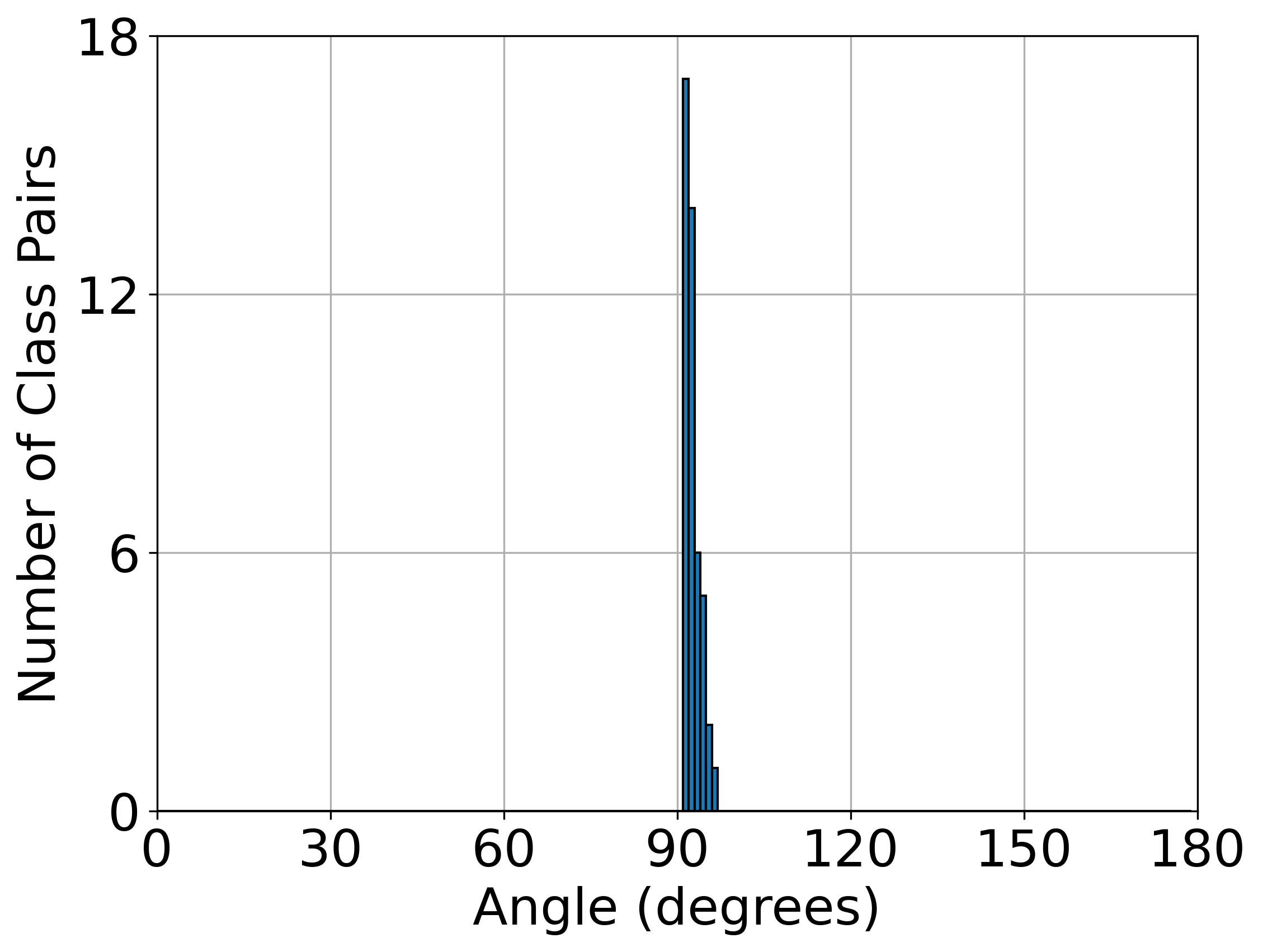}}
\\
\subfloat{\includegraphics[width=1in]{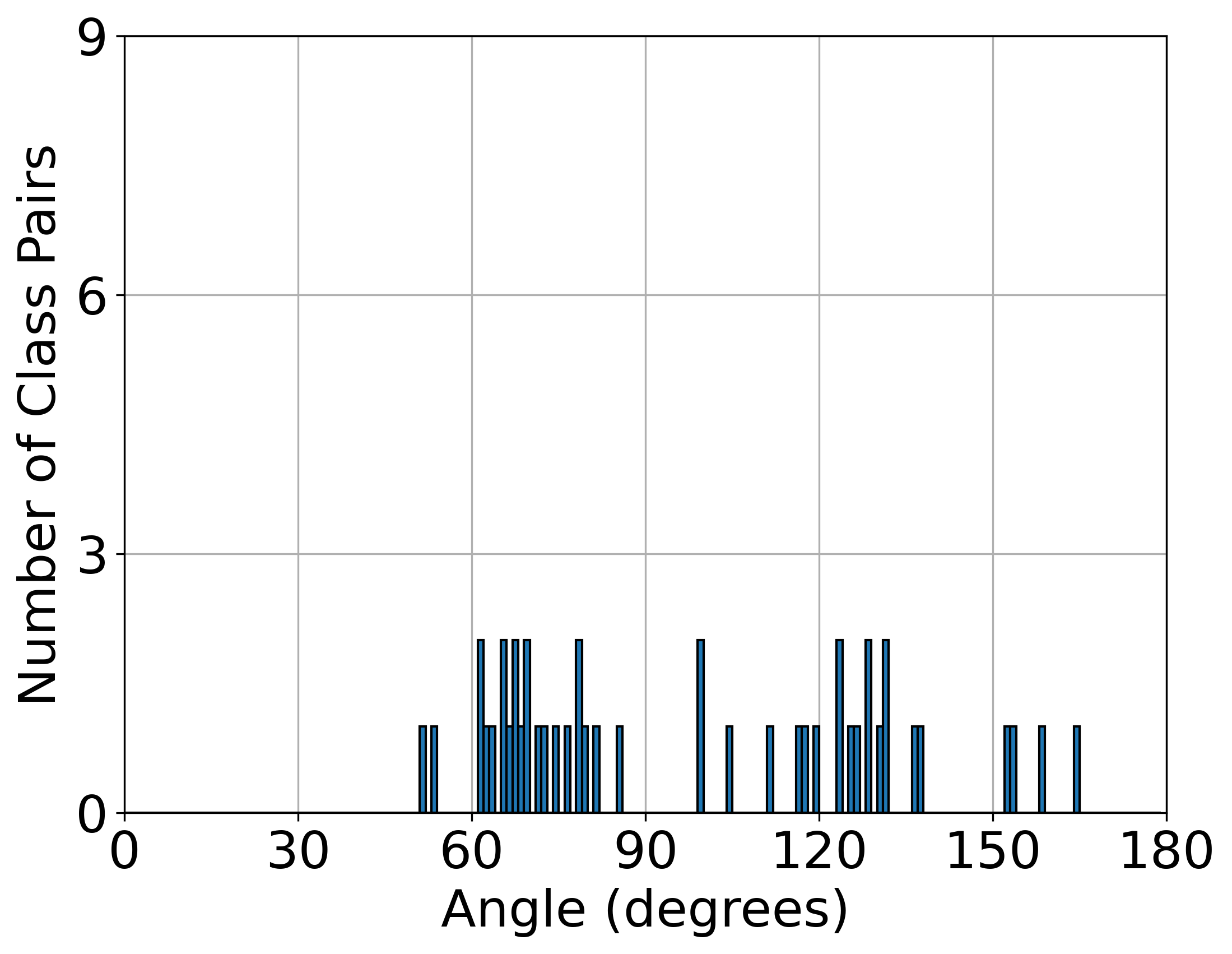}}
\hfil
\subfloat{\includegraphics[width=1in]{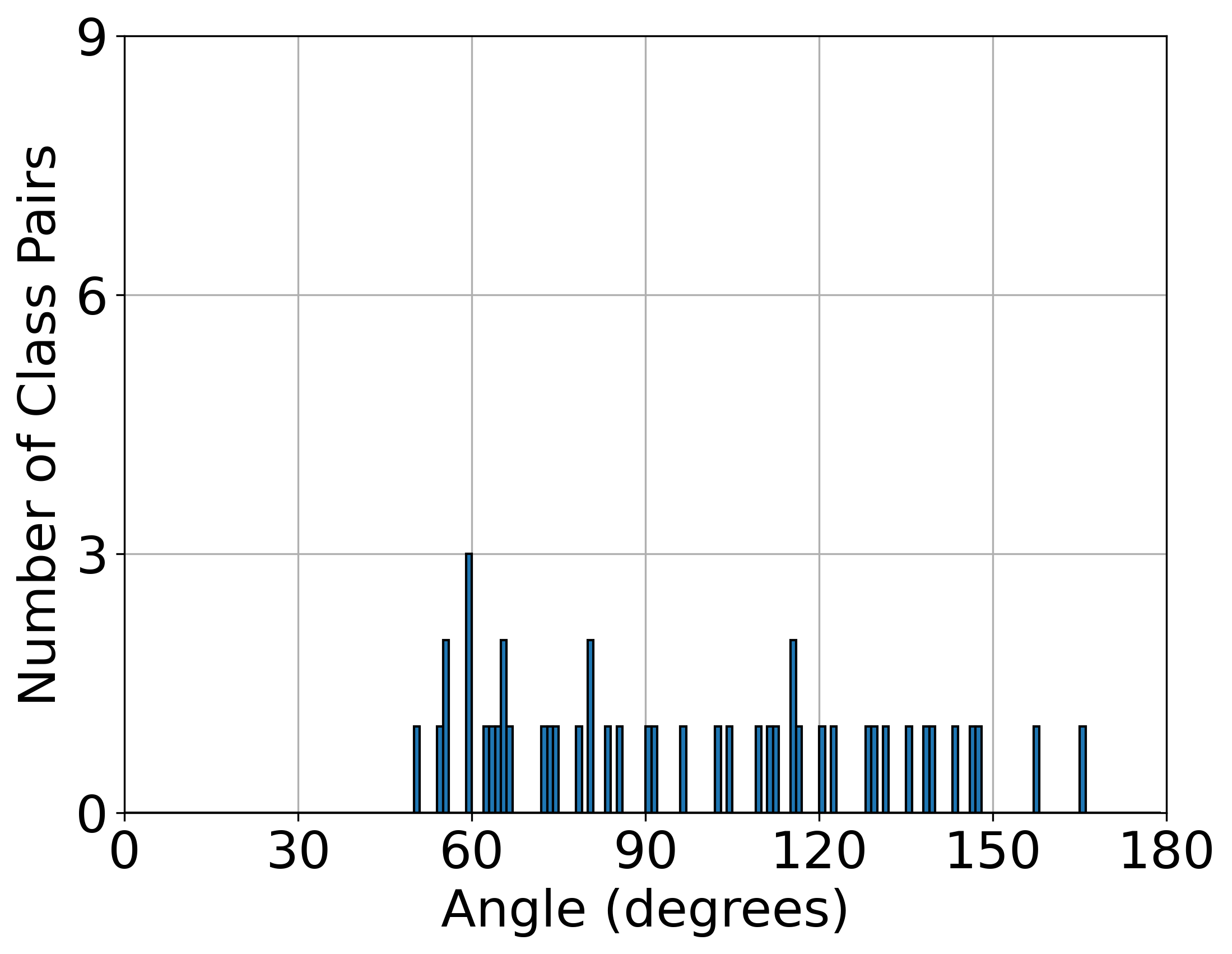}}
\hfil
\subfloat{\includegraphics[width=1in]{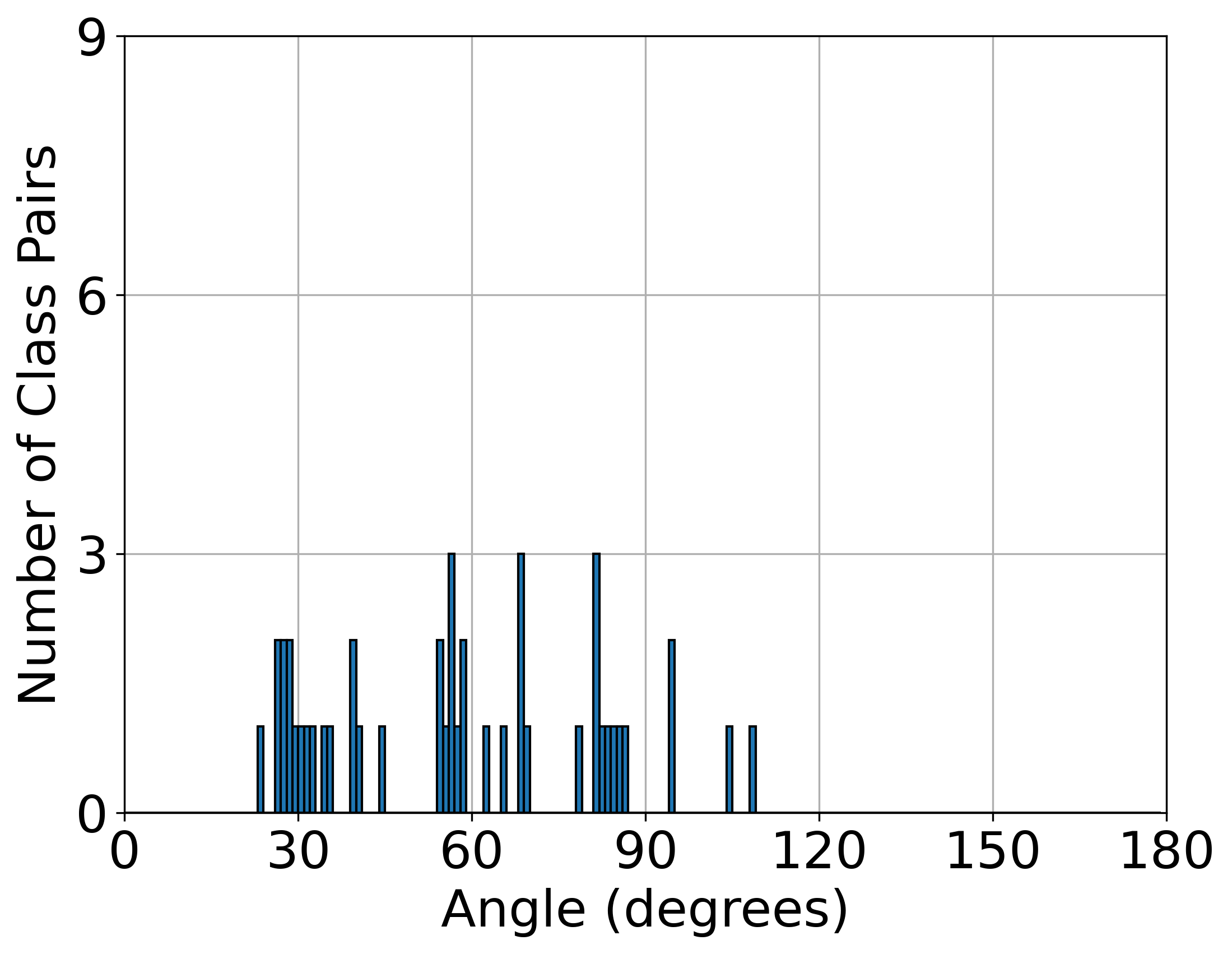}}
\hfil
\subfloat{\includegraphics[width=1in]{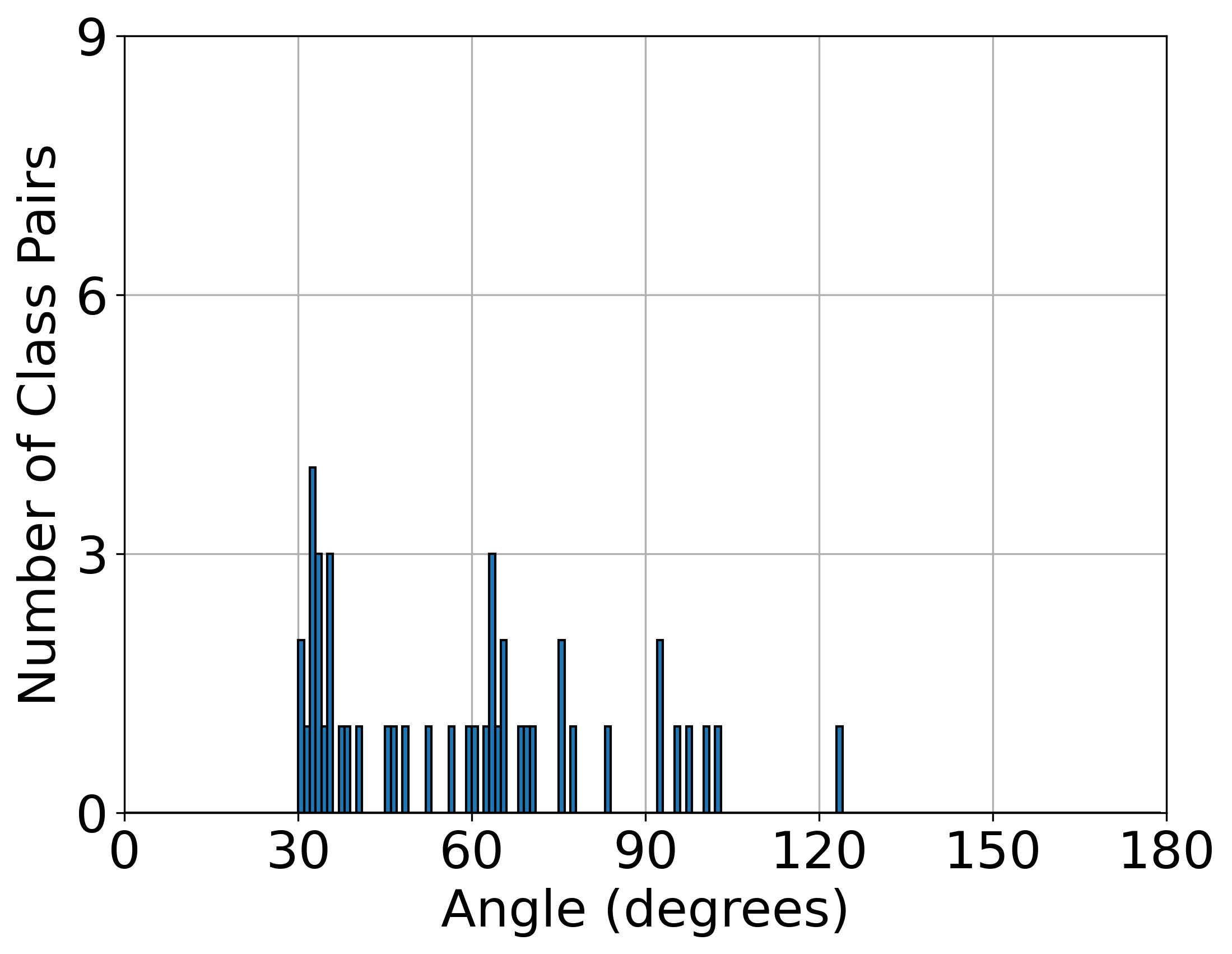}}
\hfil
\subfloat{\includegraphics[width=1in]{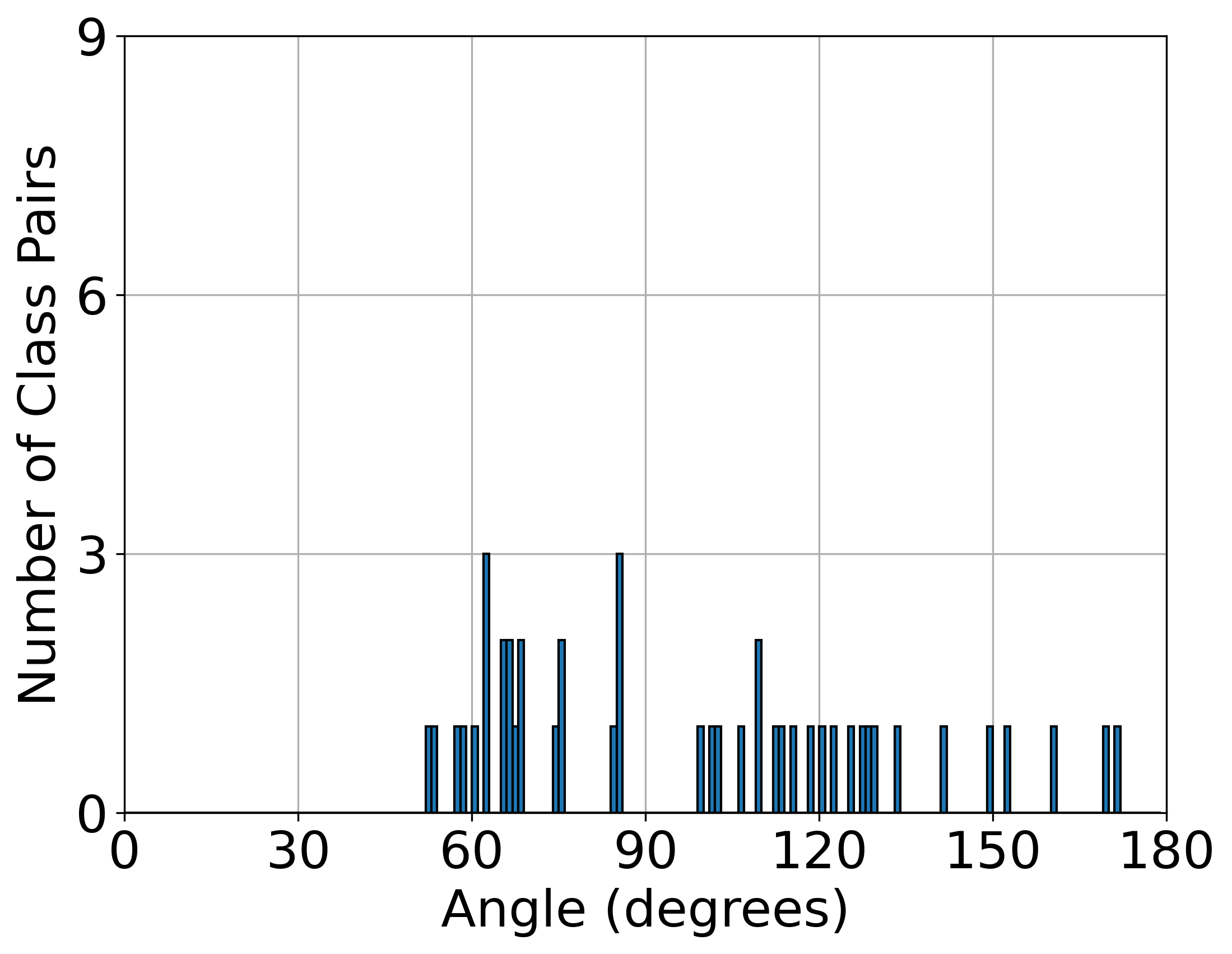}}
\hfil
\subfloat{\includegraphics[width=1in]{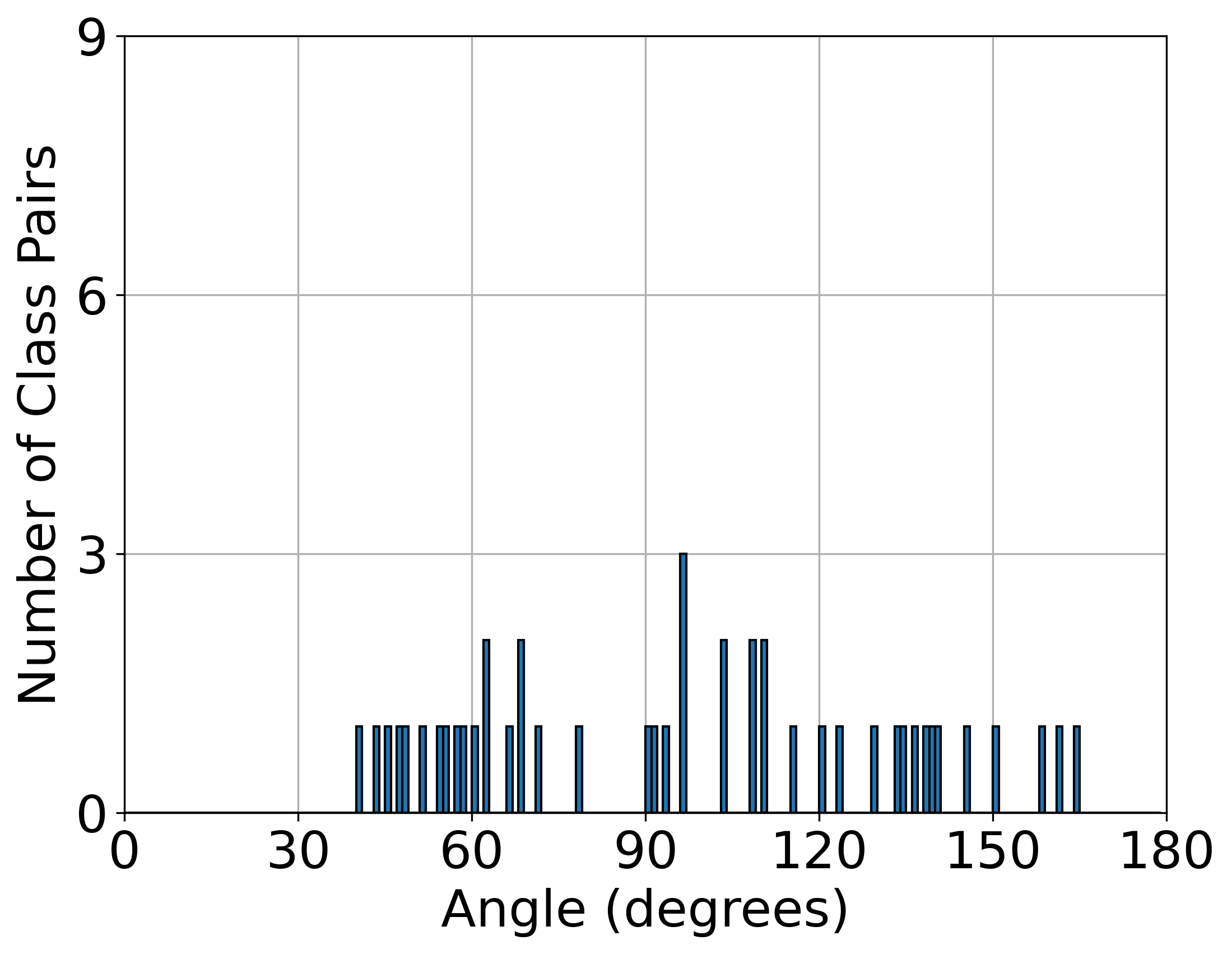}}
\hfil
\subfloat{\includegraphics[width=1in]{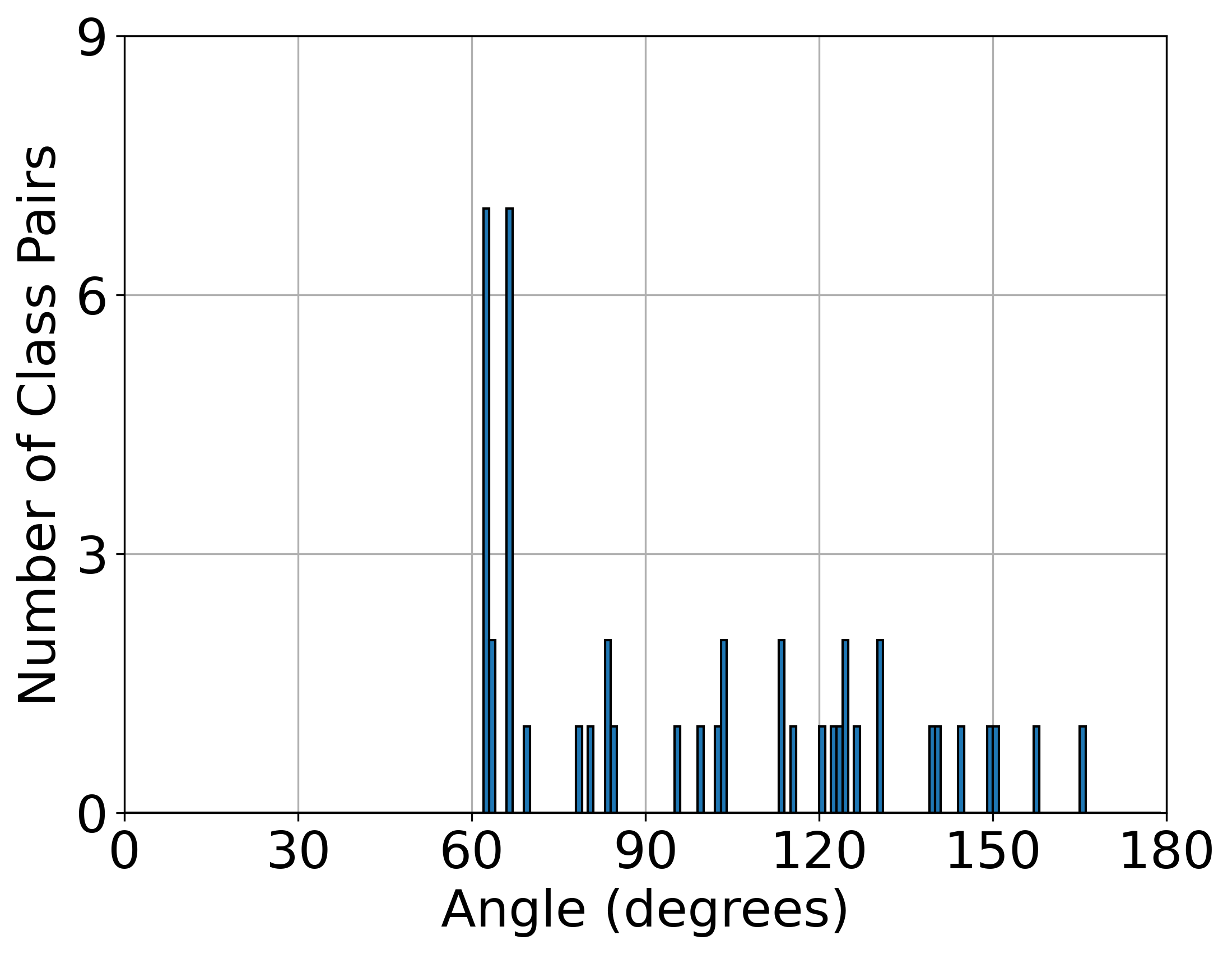}}
\\
\subfloat{\includegraphics[width=1in]{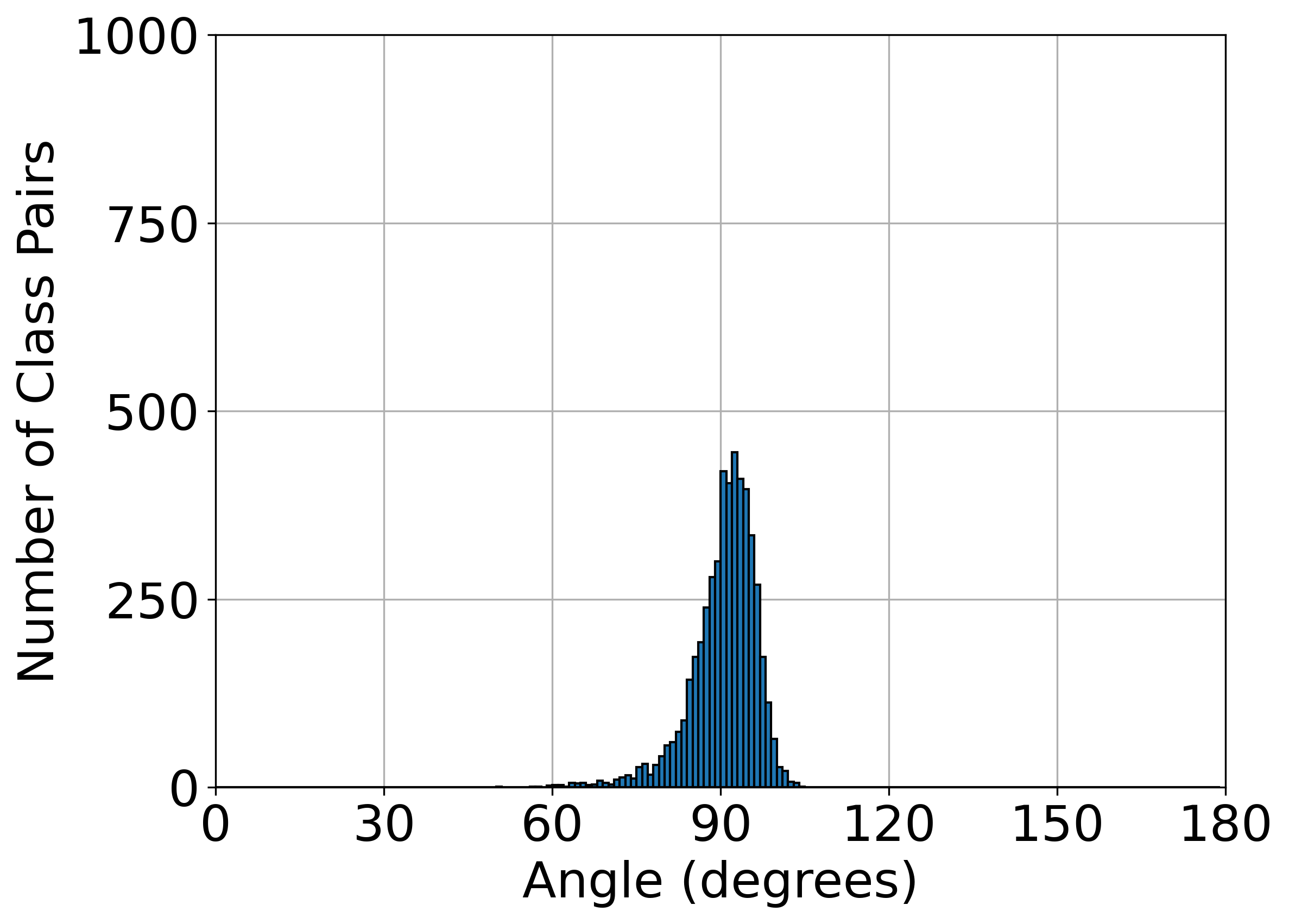}}
\hfil
\subfloat{\includegraphics[width=1in]{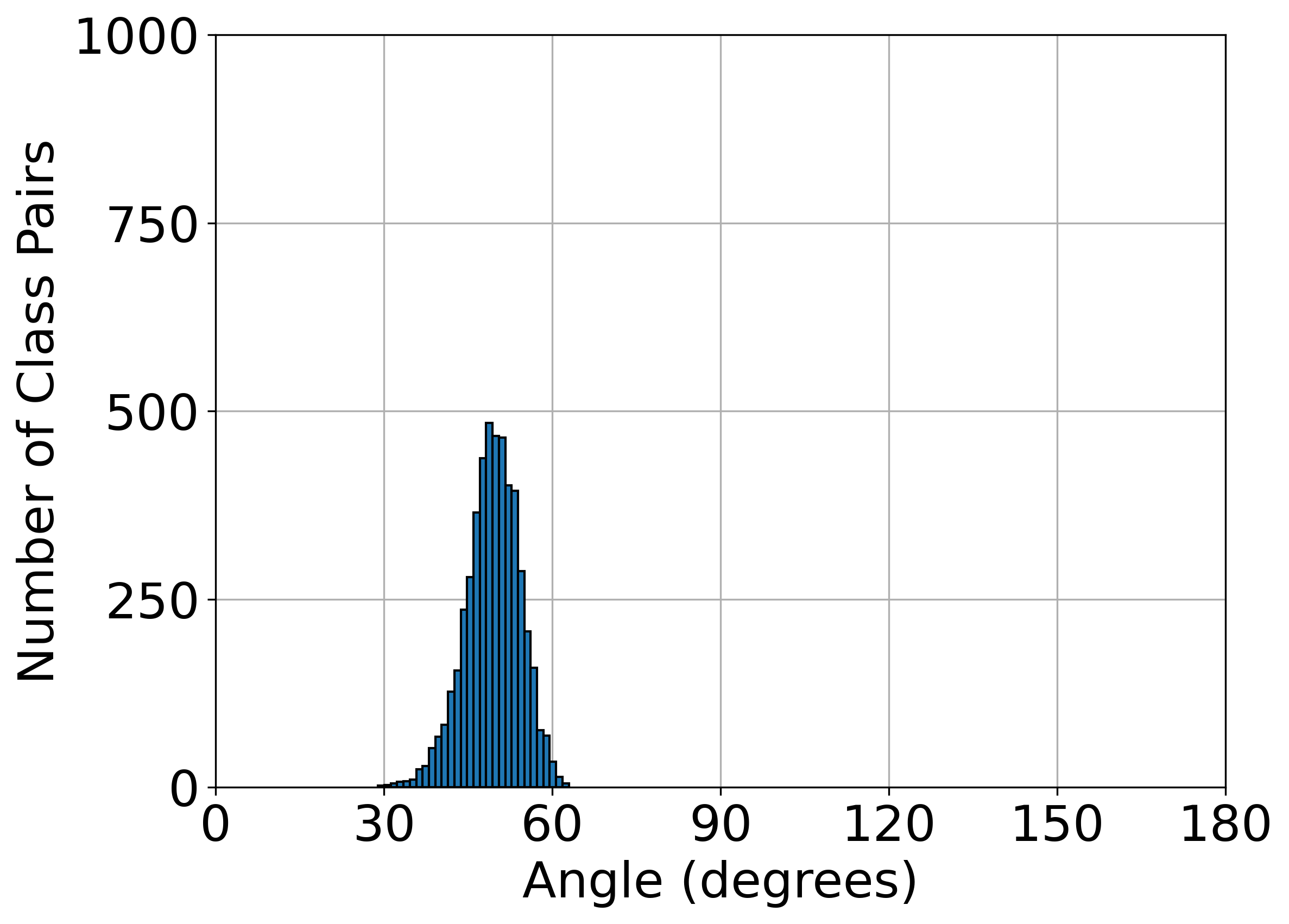}}
\hfil
\subfloat{\includegraphics[width=1in]{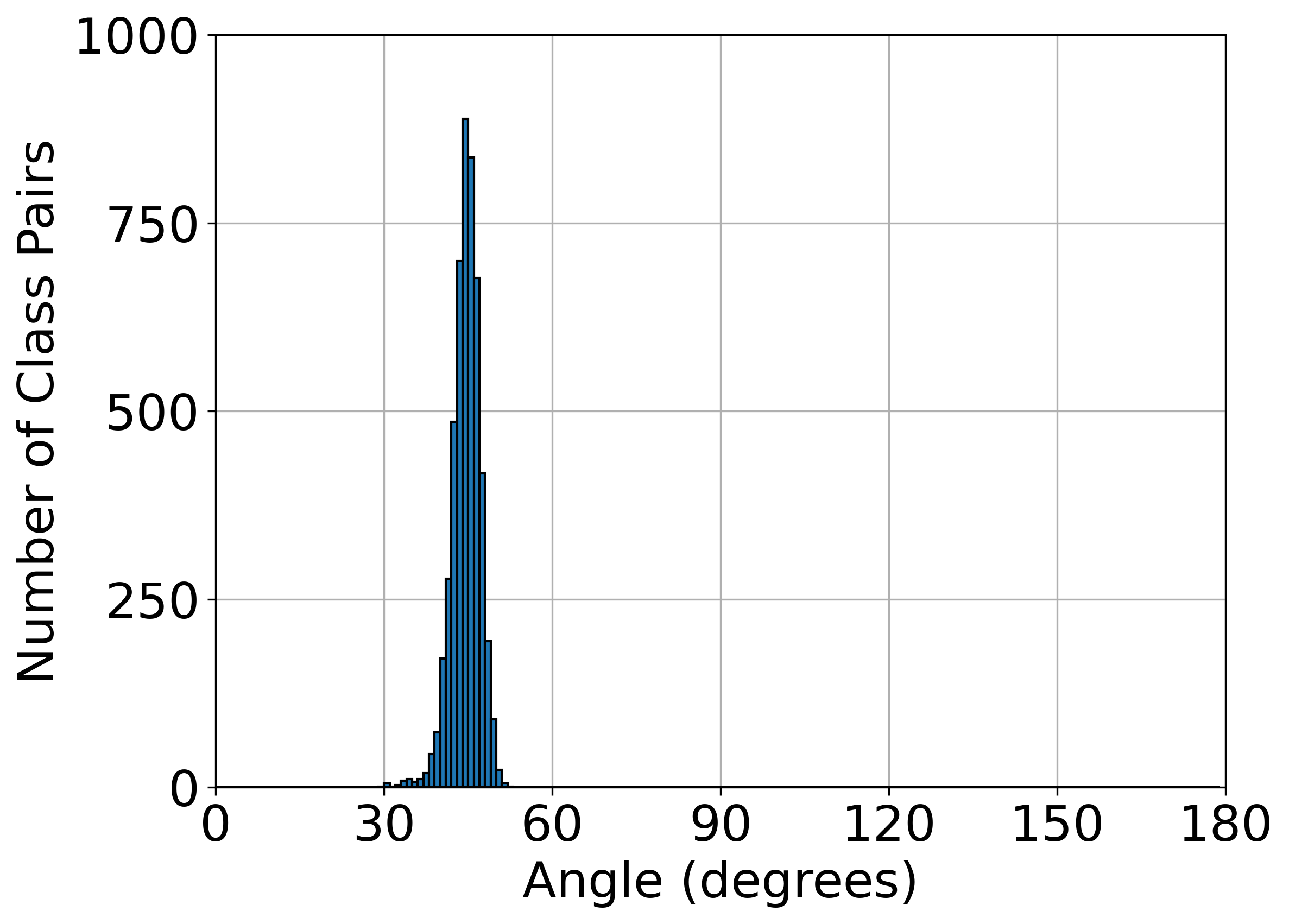}}
\hfil
\subfloat{\includegraphics[width=1in]{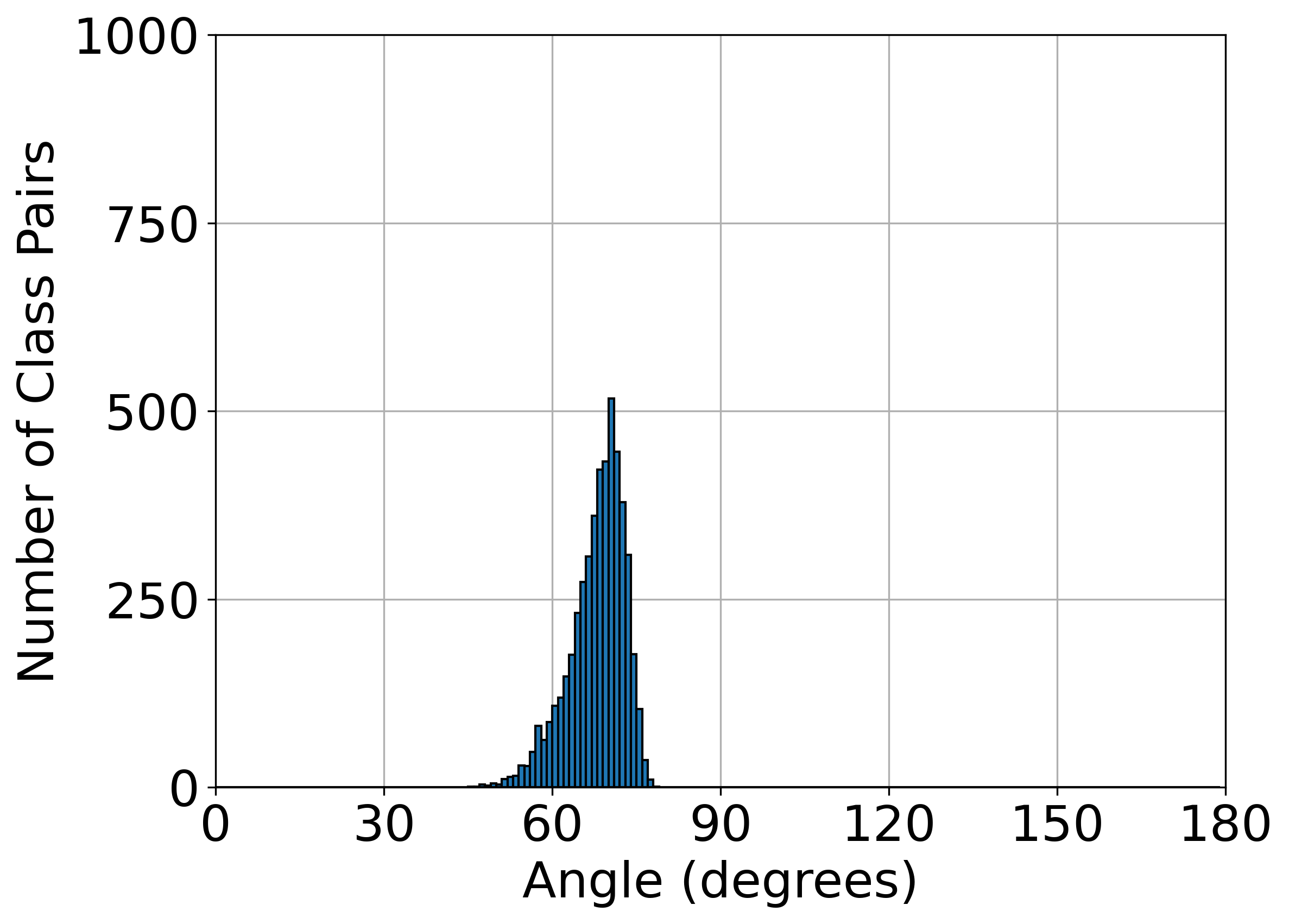}}
\hfil
\subfloat{\includegraphics[width=1in]{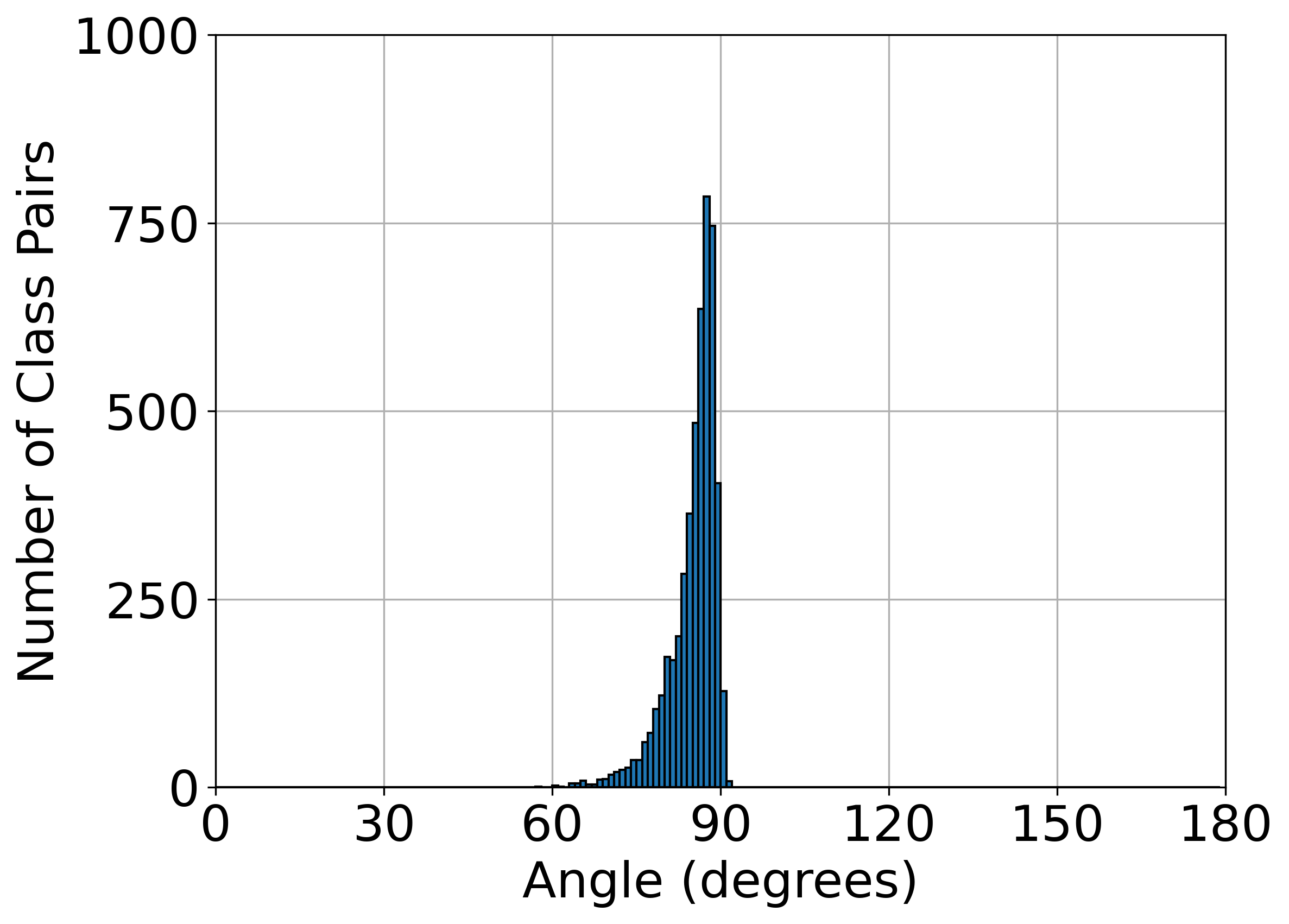}}
\hfil
\subfloat{\includegraphics[width=1in]{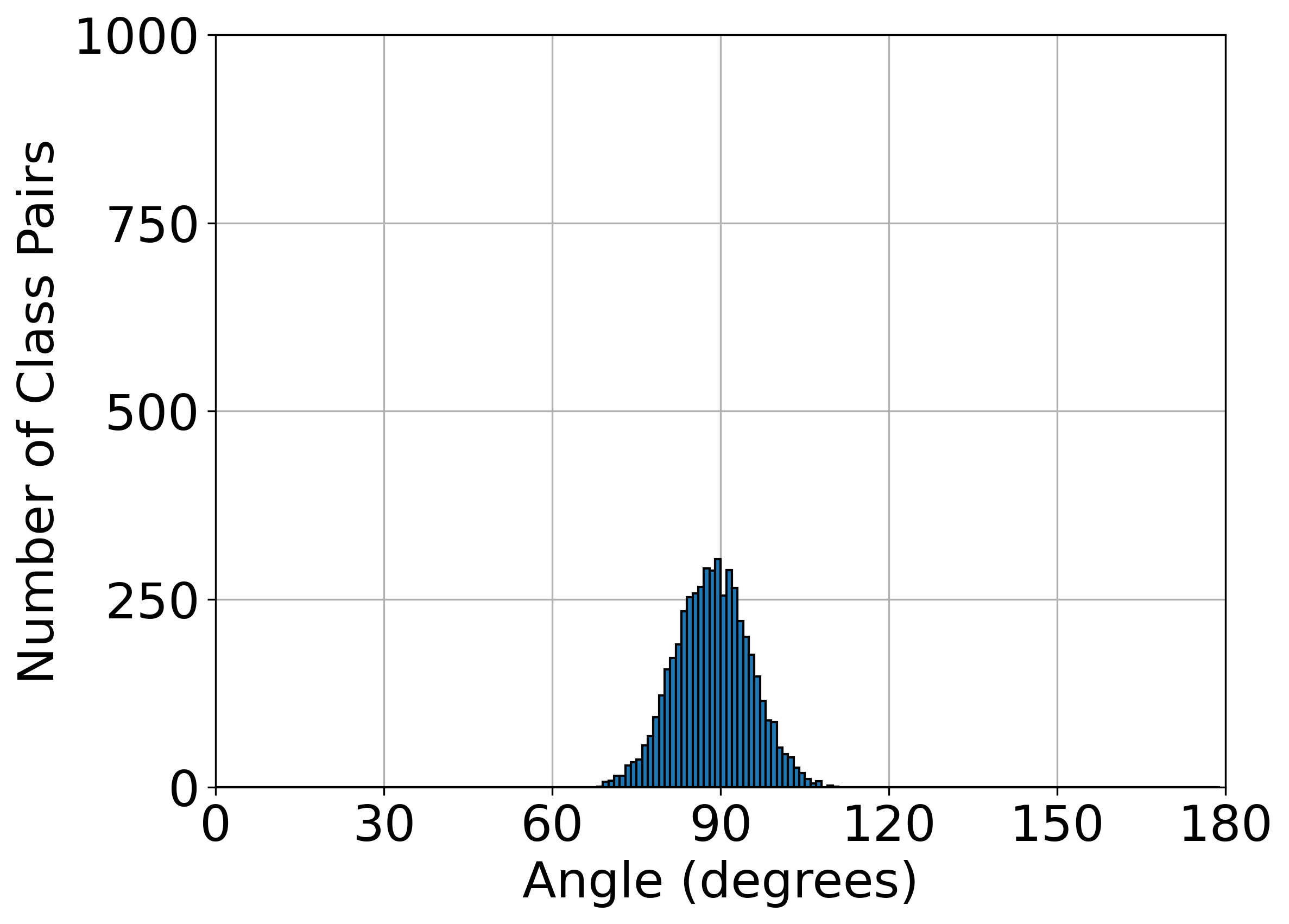}}
\hfil
\subfloat{\includegraphics[width=1in]{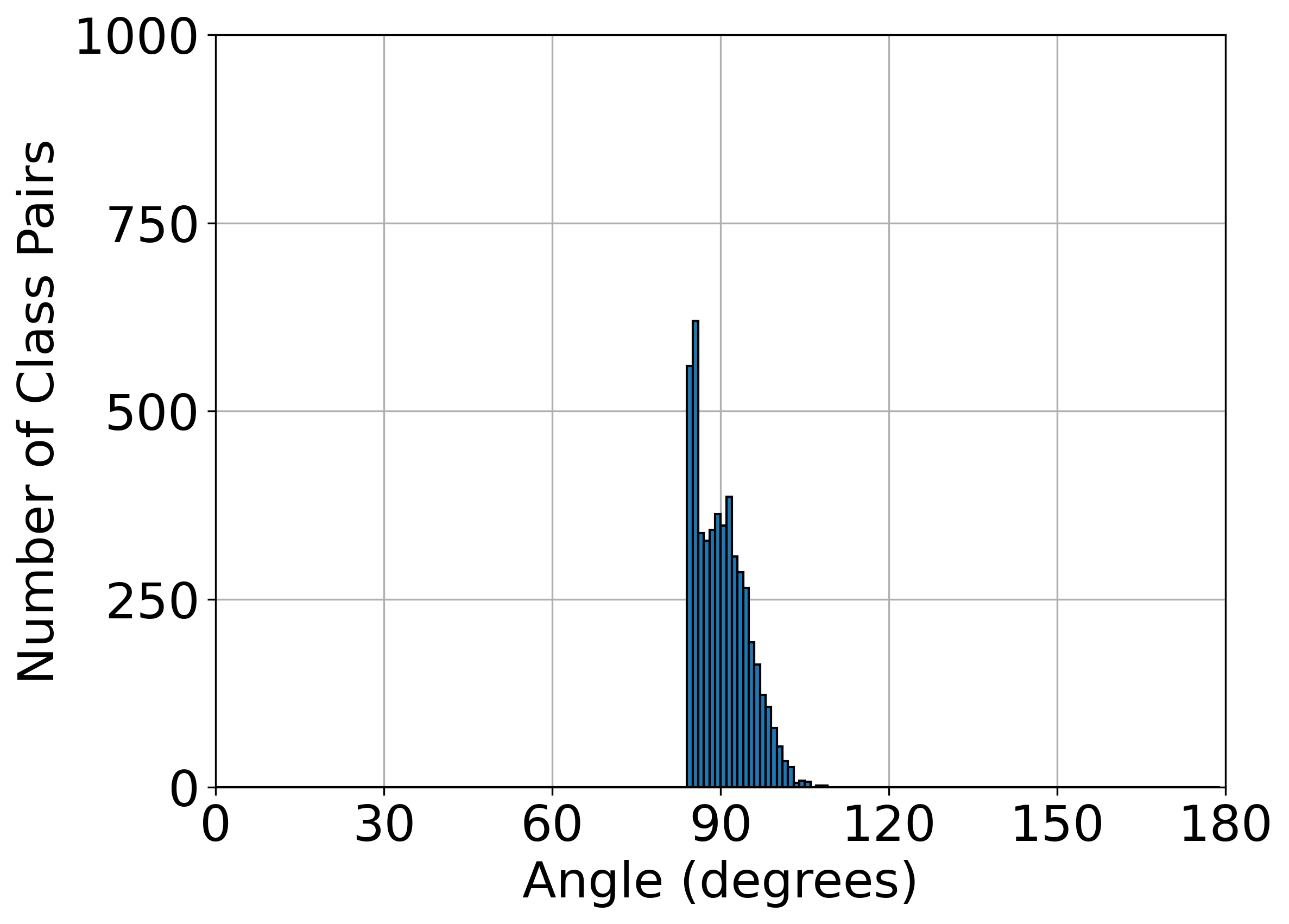}}
\\
\setcounter{subfigure}{0}
\subfloat[CE]{\includegraphics[width=1in]{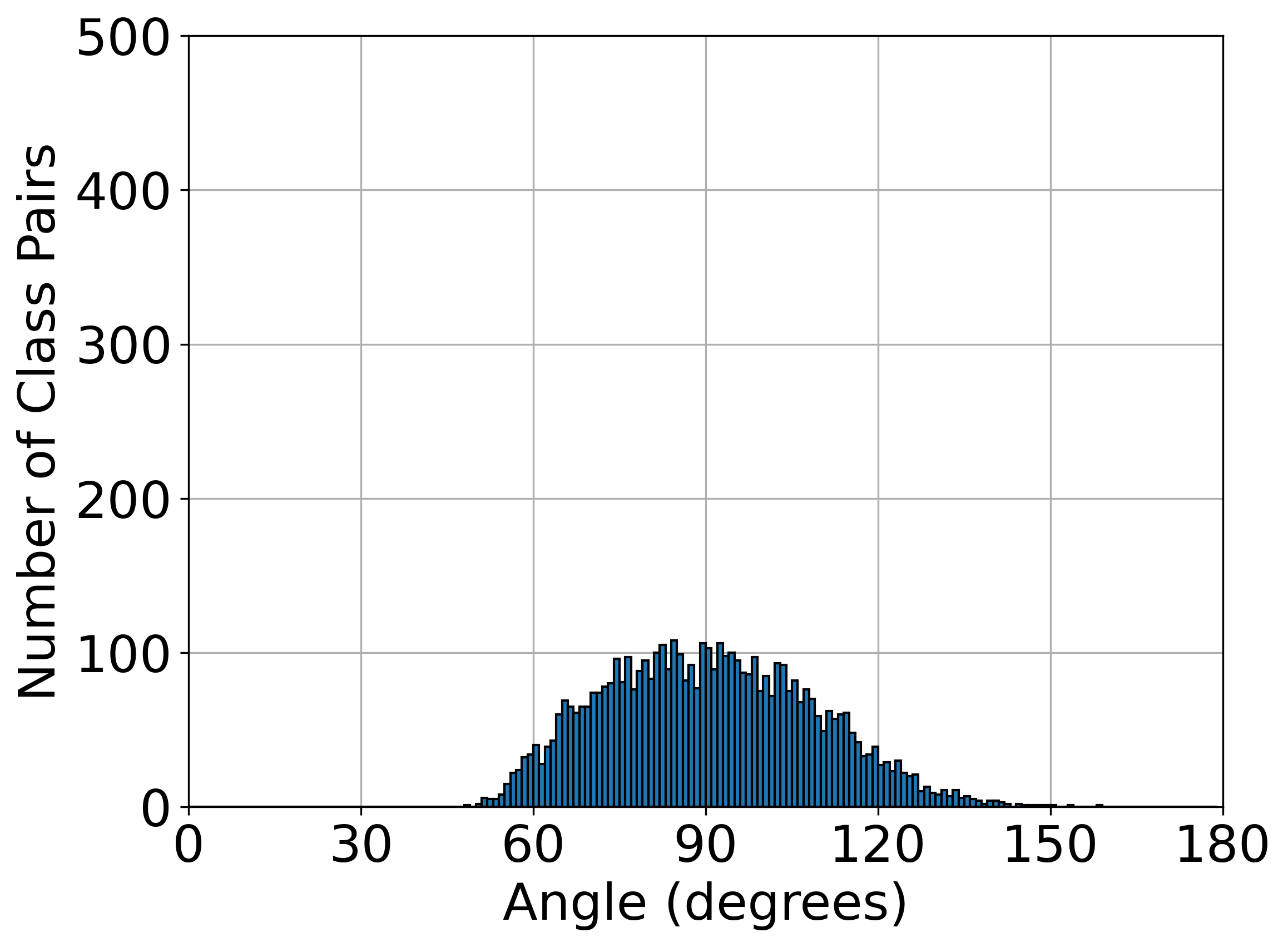}}
\hfil
\subfloat[CL]{\includegraphics[width=1in]{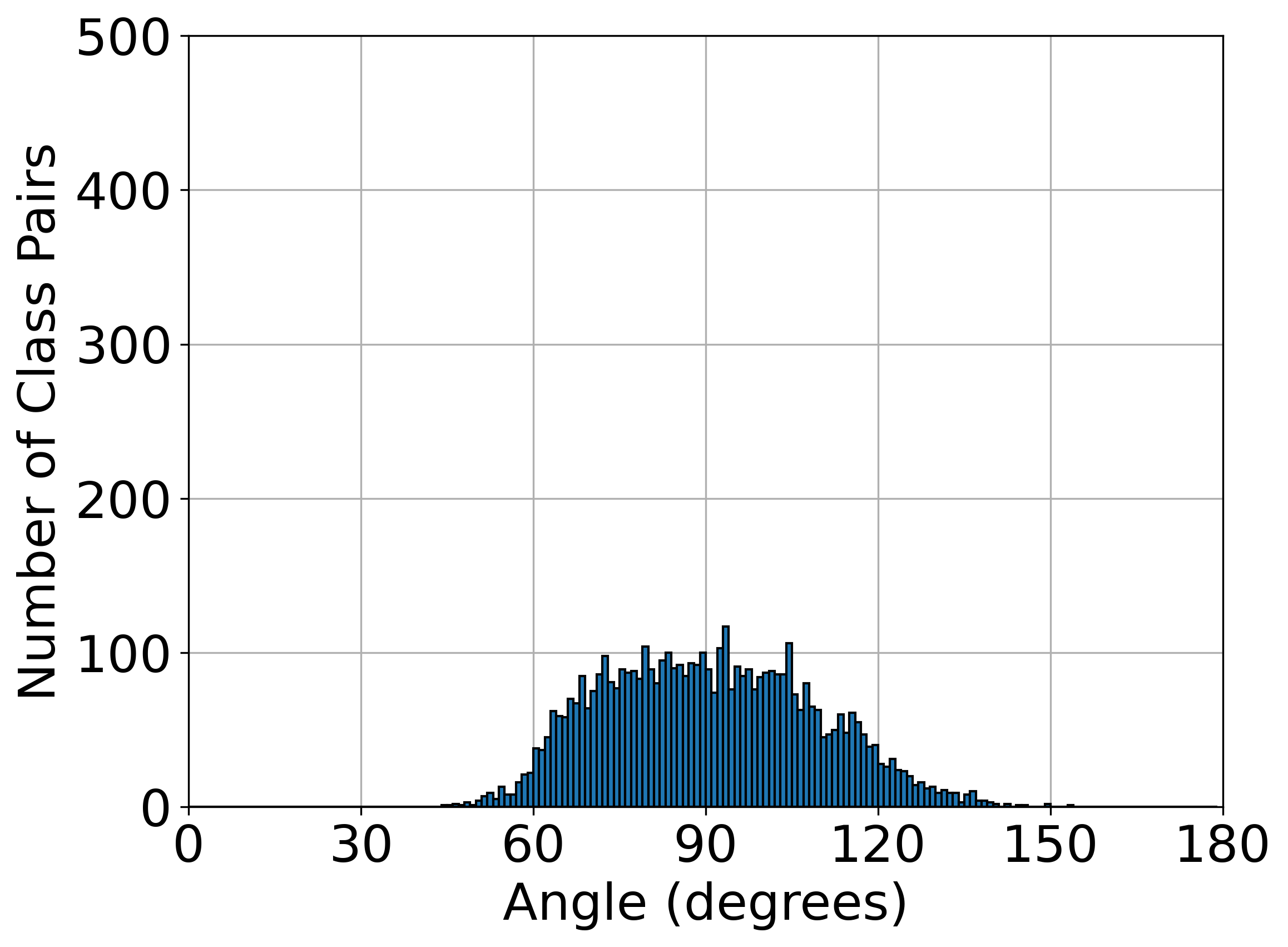}}
\hfil
\subfloat[CCL]{\includegraphics[width=1in]{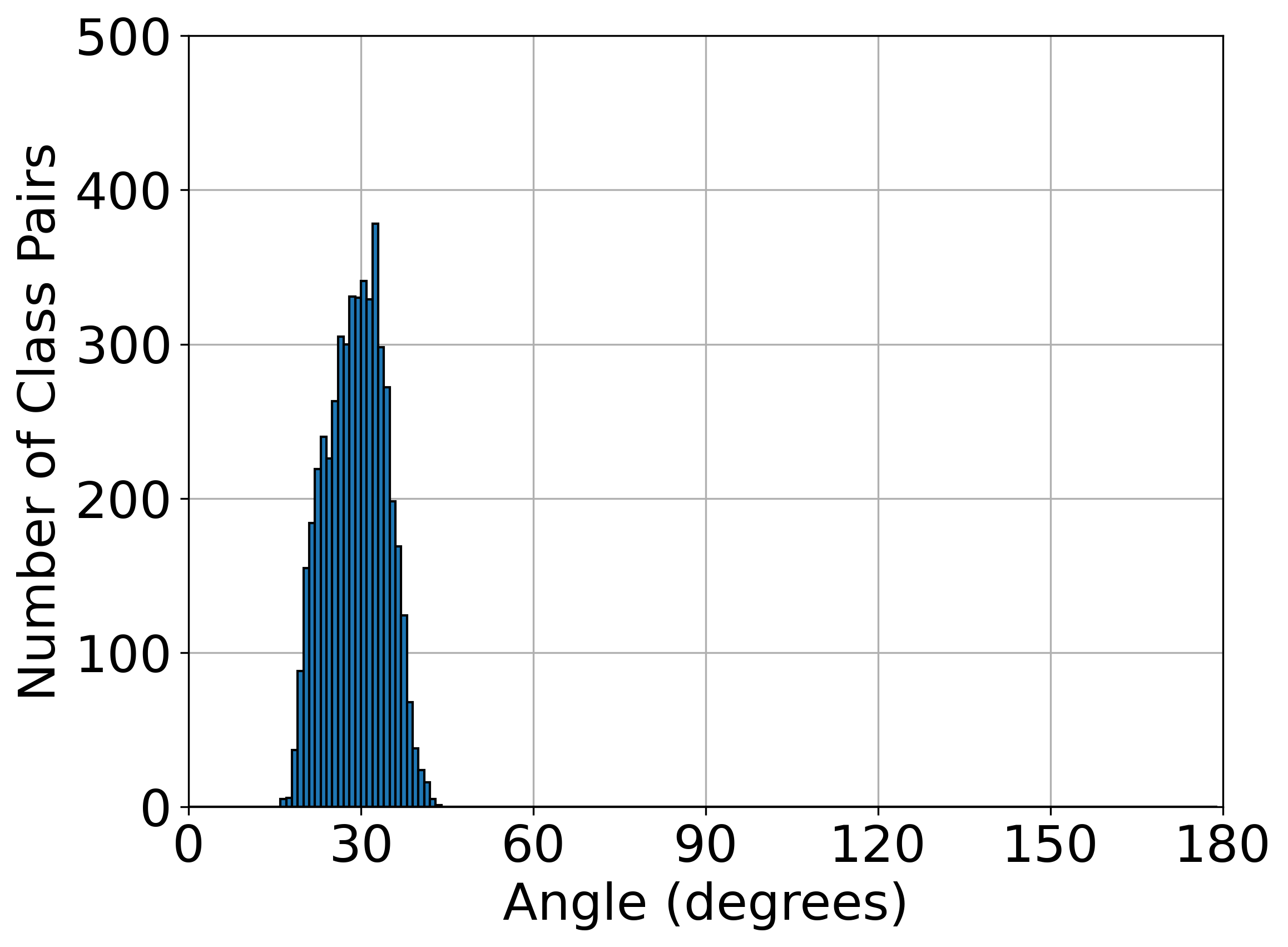}}
\hfil
\subfloat[SCCL]{\includegraphics[width=1in]{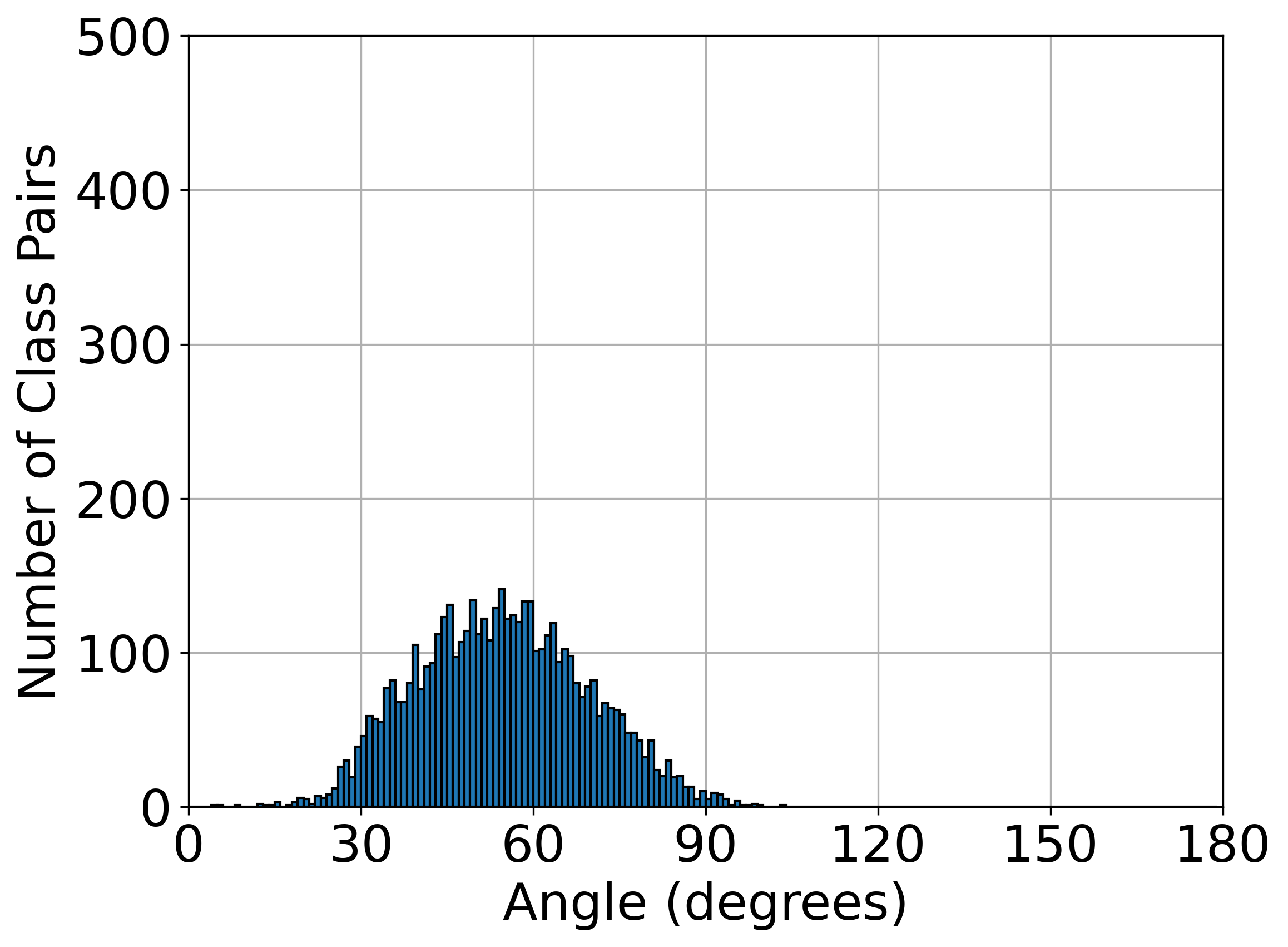}}
\hfil
\subfloat[CPN]{\includegraphics[width=1in]{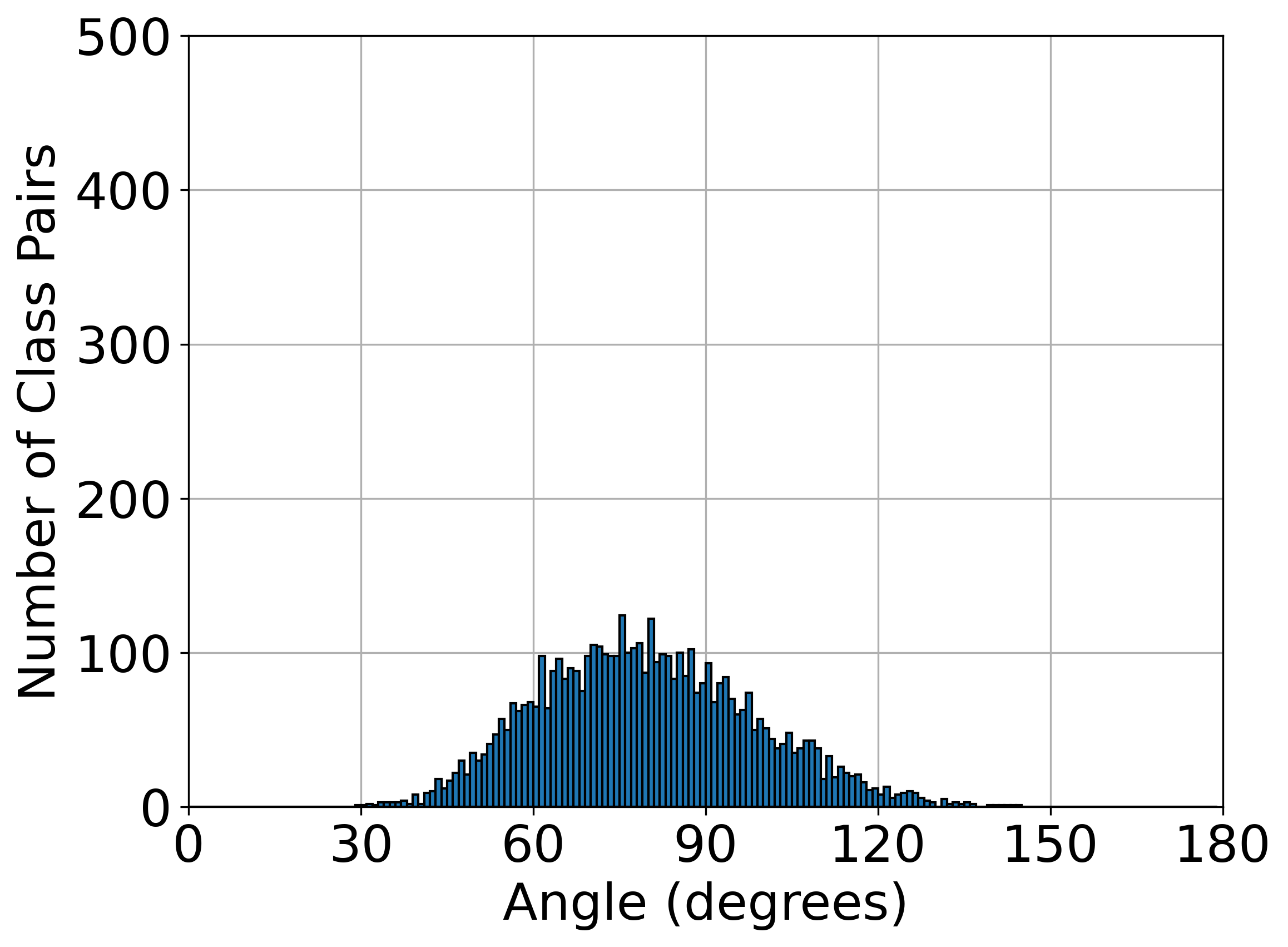}}
\hfil
\subfloat[DPP]{\includegraphics[width=1in]{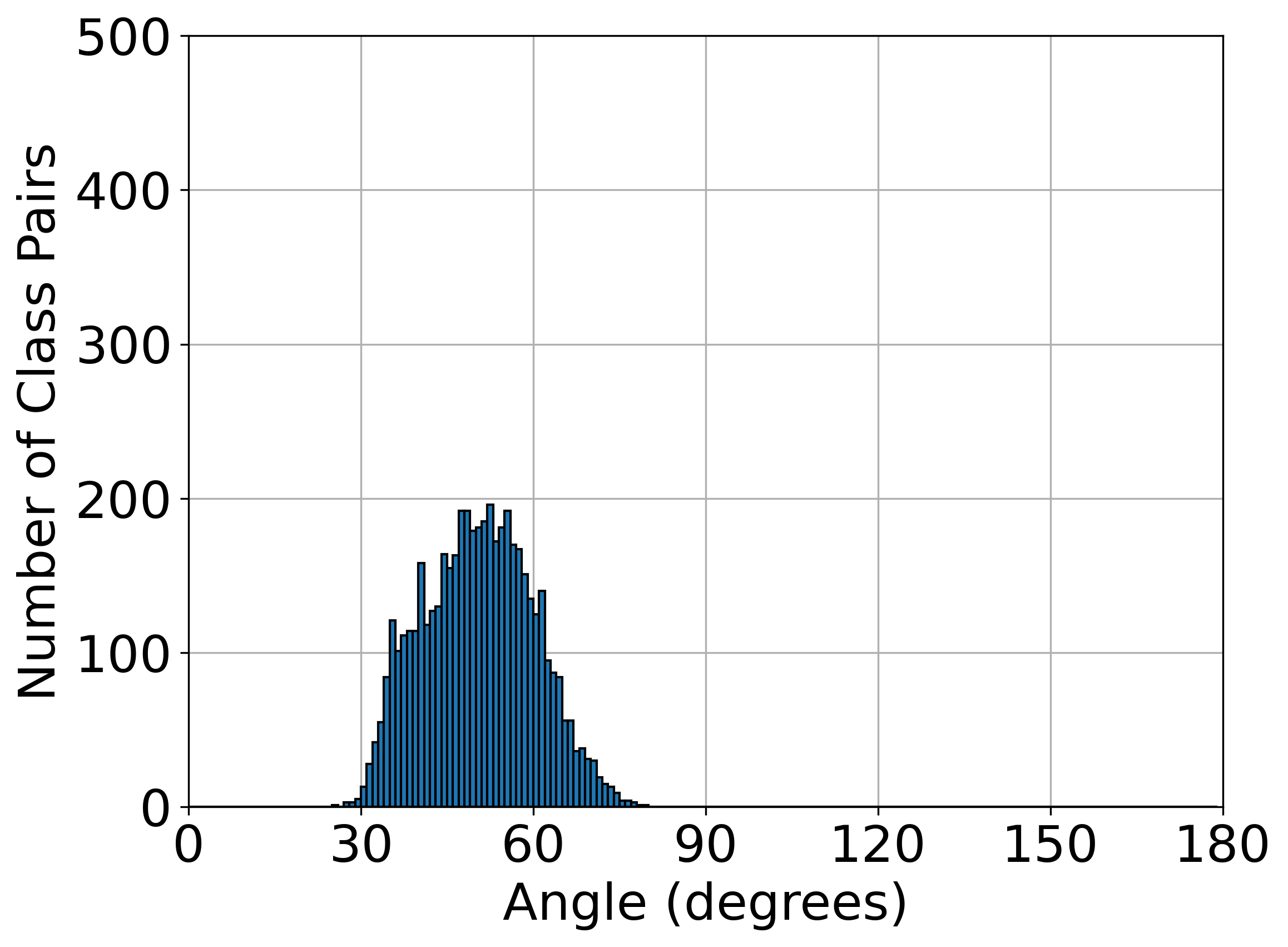}}
\hfil
\subfloat[DPNP]{\includegraphics[width=1in]{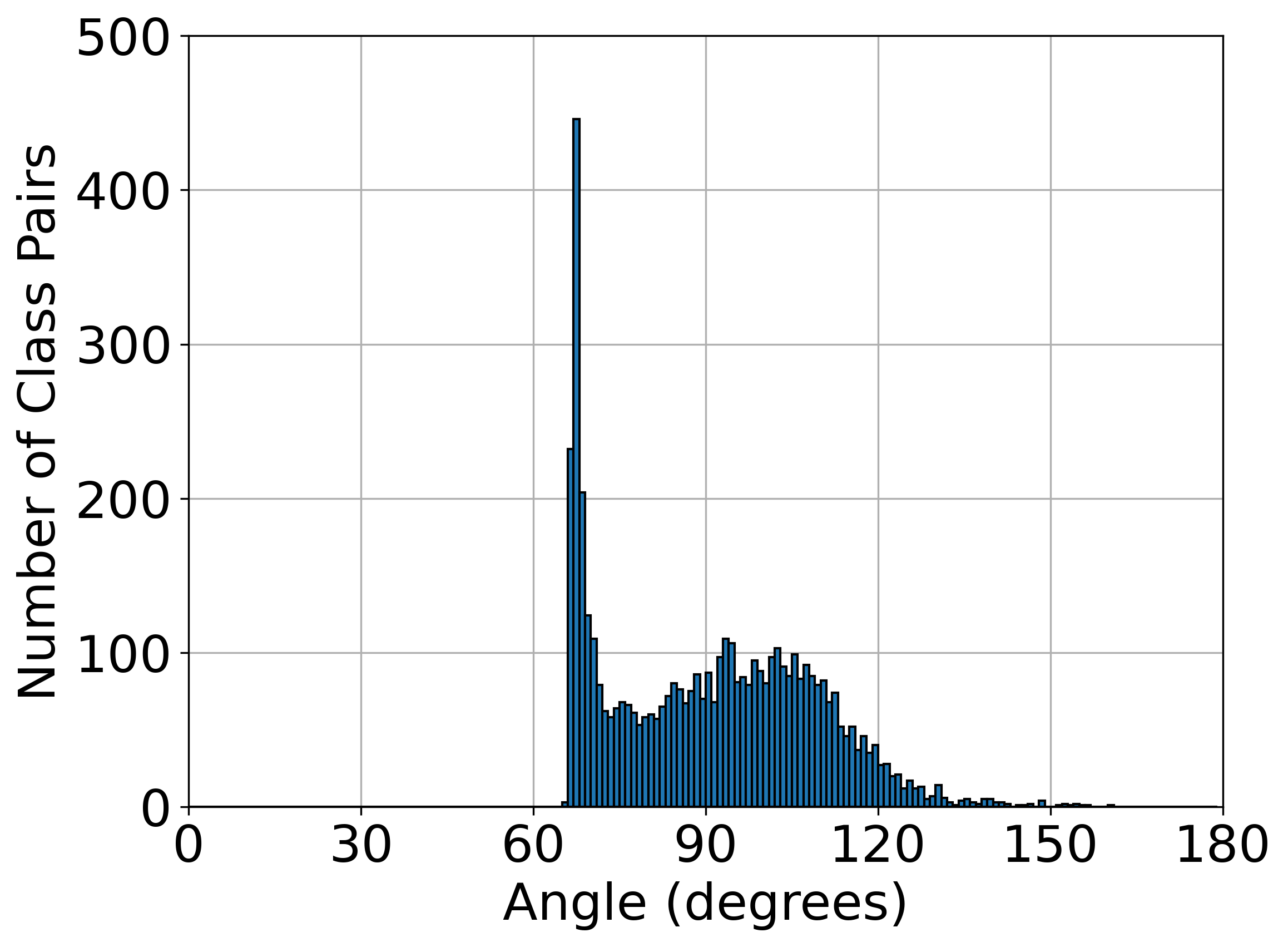}}
\caption{Histograms of inter-class angles for different methods: the first two rows correspond to CIFAR-10, and the bottom two rows show the results of CIFAR-100. Standard ResNet-18 architecture is used in odd rows, while reduced ResNet-18 is used in even rows. The proposed models show concentrated histograms with larger minimums which denote better separation.}
\label{fig_BC}
\end{figure*}

In addition to evaluating classification accuracy, it is crucial to assess the obtained feature space representations, and to check how well different classes are separated. A clear separation between classes enhances the model's generalization and robustness, particularly in complex classification tasks with overlapping class boundaries. 
To quantify inter-class separation, we analyze the angular distribution between DPPs in the feature space for our proposed models and compare them with baseline and state-of-the-art methods across different datasets and architectures. For each trained model, we extract the prototype of class $j$ in the feature space, denoted as $c_j$. The inter-class angle $\phi_{jk}$ between each two prototypes ${c}_{j}$ and ${c}_k$ in degrees, is then computed as follows:
\begin{equation}
\label{eq15}
\phi_{jk}=arccos({c_j}^{T}c_k)\times\left(\frac{180}{\pi}\right)
\end{equation}
For rival methods, CE and SCCL, the classification weights are used as class centers. In the case of CL and CCL, where two independent sets of parameters are present for classification and class centers, the class center parameters are used. Histograms of the inter-class angles resulted from different methods are visualized in Figure \ref{fig_BC} for the CIFAR-10 and CIFAR-100 datasets, using both the standard and reduced ResNet-18 architectures. Here, larger angles indicate better separation.

The noticeable peaks on the left side of DPNP angle histograms denote its success in the arrangement of DPPs in the feature space, as it has collected the left tail into this peak that occurs at larger angles. The injection of regularity by DPNP is even more evident and effective when the network size is reduced in even rows of Figure \ref{fig_BC}. This fact leads to the least accuracy drop, see Table \ref{tab:acc_reduced_resnet18}, when the network is shrunk into less than half of its original size.
The reported results in Table \ref{tab:BC_angle_comparison} quantitatively confirm that both DPP and DPNP models indeed can achieve better separation compared to baseline and state-of-the-art methods. Here, $MinSep$ denotes the minimum histogram value in each case and is a proper measure of separability, which guarantees an angular margin between class prototypes/centers. DPNP surely surpasses the DPP model in this regard as it benefits from extra repulsive terms in its loss.

Since $MinSep$ is an order statistic, it may vary in each run and does not offer a stable measure. One may prefer to use histogram mean instead. However, it should be noted that since far away class centers are also included in the inter-class angle histograms, mean cannot be directly calculated from the histogram itself. 
In fact, if the number of classes is relatively large with respect to the dimensionality of the feature space, as it is the case with even rows in Figure \ref{fig_BC}, then the angle histograms are more dispersed since class centers are placed between each other. In these cases, only the nearest neighbors should be considered in assessing separability. Therefore, we also report $MeanSep$, the average of the angles between class centers and their nearest neighbor.  A good classification model which can inject enough regularity into the feature space, leads to larger $MinSep$ as an angular margin and $MeanSep$, as an inter-class separation measure; see Table \ref{tab:BC_angle_comparison}.

On the other hand, when the number of classes is lower than the feature space dimensionality, such as the reported experiments with standard ResNet18, $MeanSep$, although smaller, nearly follows the histogram mean; see the odd rows of Figure \ref{fig_BC} and Table \ref{tab:BC_angle_comparison}. In these cases a 90-degree or more angle margin is feasible, though it may not be obtained due to the limitations of the feature extractor or learning algorithm. In Table \ref{tab:BC_angle_comparison}, DPNP offers the best inter-class separation, followed by the DPP model as the second winner.

Finally, the standard deviation of the angles between each DPP/class center and its nearest neighbor is also reported as $Std$, to see how uniform the desired regularity is present in the feature space. Again, our proposed models are performing best by the least variation in the delivered inter-class separation.

Models that use separate parameters for classification weights and class centers, i.e., CL and CCL, cannot directly benefit from the dispersion CE creates and hence achieve less separation, especially in higher-dimensional spaces, where centers may diverge more from the corresponding weights. A second affecting element is how the centers are learned or adapted. CL, CPN and our proposed models DPP and DPNP, all propagate the loss gradient and update the centers correspondingly, while CCL and SCCL update the network parameters based on the gradient and then use feature space representation of the class members to recalculate the centers. Therefore, the latter methods lose the CE initiated separation. Hopefully, DPNP not only gain better separation from these two aspects, but also explicitly use the negative prototypes to even further increase the regularity of its feature space.

The enhanced inter-class separation of DPNP correlates with the improvements observed in classification accuracy (as detailed in the previous subsection). By maximizing the angles between class prototypes, our model reduces overlap between classes in the feature space, leading to more confident and accurate predictions. This in turn can be attributed to the incorporation of negative prototypes at both class and instance levels; while removing the redundancies in parametrization via unification of classification weights and class centers plays also a crucial role in letting the corresponding terms in the loss function do their job in organizing the topology of the feature space.

\begin{table*}[t]
\centering
\caption{Comparison of inter-class angles (in degrees) on CIFAR-10 and CIFAR-100}
\label{tab:BC_angle_comparison}
\begin{tabular}{c|ccc|ccc||ccc|ccc}
\toprule
 & \multicolumn{6}{c||}{\textbf{Standard ResNet-18}}  & \multicolumn{6}{c}{\textbf{Reduced ResNet-18}} \\
\cmidrule(lr){2-7} \cmidrule(lr){8-13} 
\textbf{Model} & \multicolumn{3}{c|}{\textbf{CIFAR-10 512D}} & \multicolumn{3}{c||}{\textbf{CIFAR-100 512D}} & \multicolumn{3}{c|}{\textbf{CIFAR-10 3D}} & \multicolumn{3}{c}{\textbf{CIFAR-100 10D}} \\
\cmidrule(lr){2-4} \cmidrule(lr){5-7} \cmidrule(lr){8-10} \cmidrule(lr){11-13}
 & MinSep & MeanSep & Std & MinSep & MeanSep & Std & MinSep & MeanSep & Std & MinSep & MeanSep & Std\\
\midrule
CE                & 69.60 & 79.54 & 7.38 & 50.29 & 68.83 & 6.35 & 51.77 & \underline{59.20} & 5.62 & \underline{48.15} & \underline{55.29} & 2.71\\
CL                & 39.24 & 44.03 & 2.47 & 28.94 & 36.38 & 3.24 & 50.16 & 55.88 & 3.91 & 39.48 & 53.88 & 4.16\\
CPN               & 60.18 & 68.71 & 4.70 & 57.10 & 68.15 & 3.70 & \underline{52.59} & 57.97 & 4.62 & 29.89 & 43.92 & 7.61 \\
CCL               & 34.60 & 36.54 & 1.10 & 29.80 & 36.12 & 2.95 & 23.70 & 27.93 & 3.62 & 16.76 & 18.68 & 0.81 \\
SCCL              & 50.43 & 51.38 & 0.71 & 45.82 & 52.56 & 2.70 & 30.19 & 31.69 & 1.08 & 4.02  & 23.01 & 6.04 \\
DPP (Ours)        & \underline{82.44} & \underline{86.34} & 2.40 & \underline{68.64} & \underline{72.25} & 1.94 & 40.12 & 48.60 & 6.18 & 25.16 & 31.38 & 1.81 \\
\textbf{DPNP (Ours)} & \textbf{91.83} & \textbf{91.97} & 0.18 & \textbf{84.39} & \textbf{84.72} & 0.31 & \textbf{63.71} & \textbf{64.02} & 0.22 & \textbf{65.82} & \textbf{66.78} & 0.62\\
\bottomrule
\end{tabular}
\end{table*}

\begin{figure*}[!b]
\centering
\subfloat{\includegraphics[width=1in]{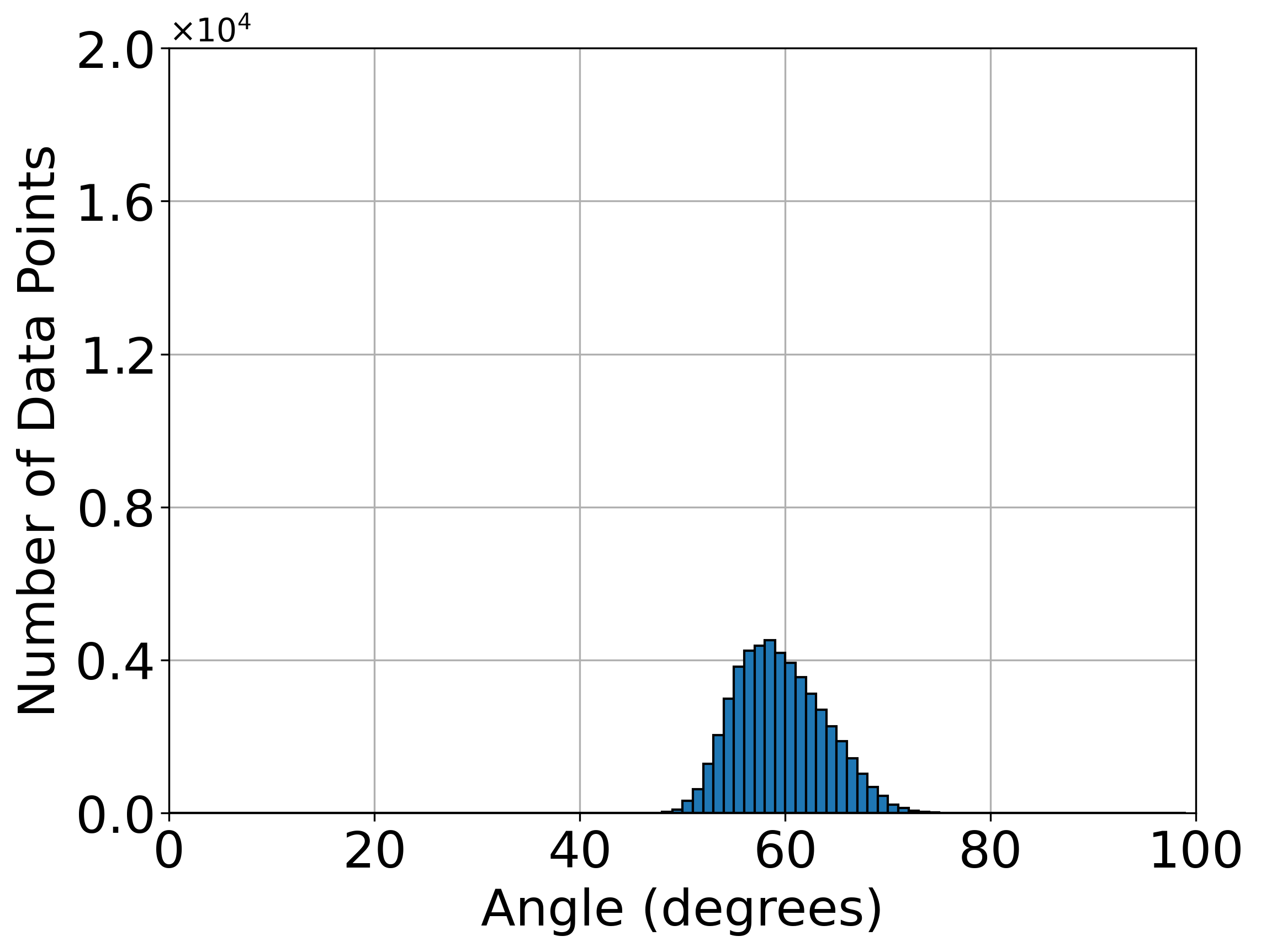}}
\hfil
\subfloat{\includegraphics[width=1in]{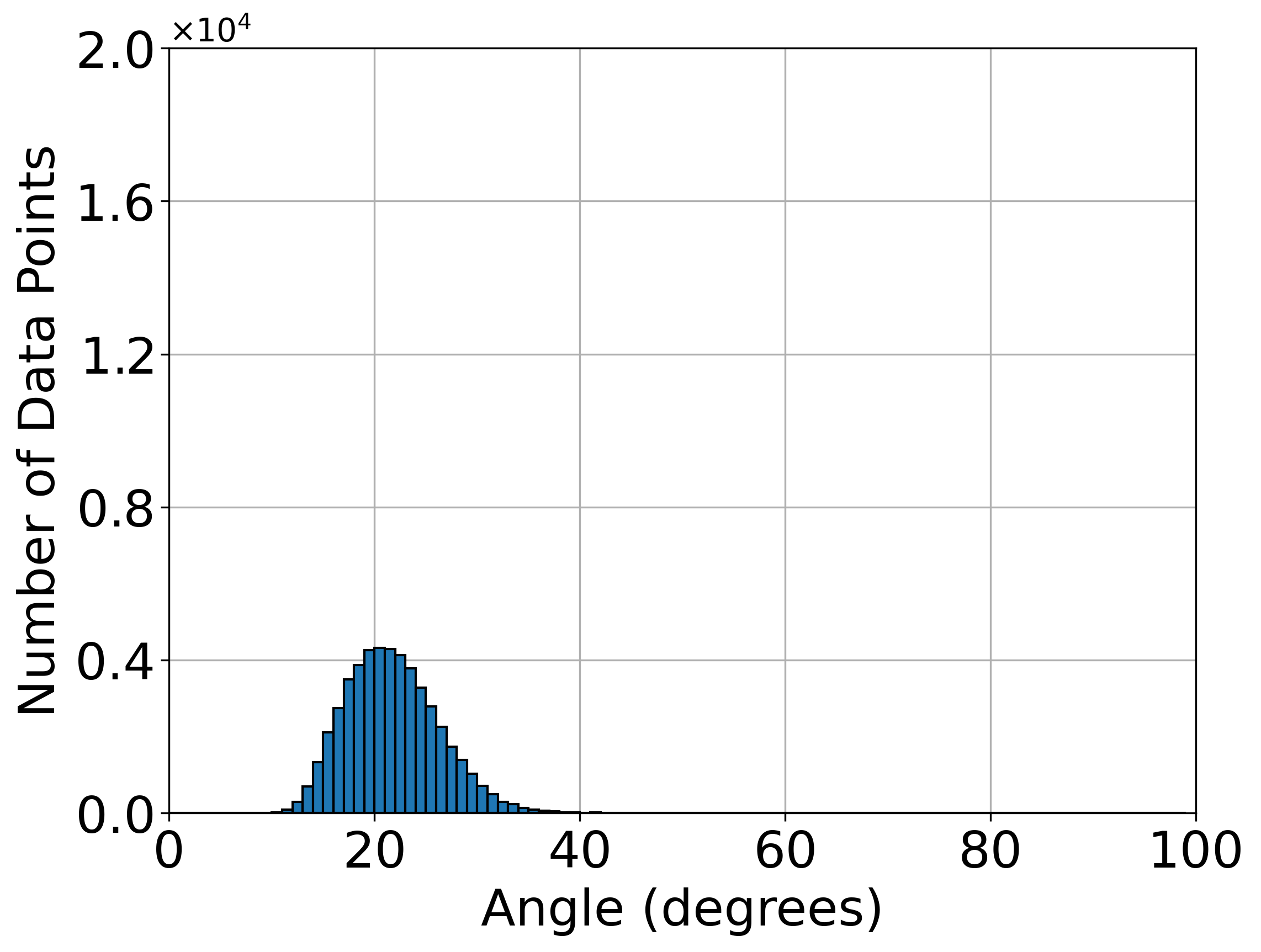}}
\hfil
\subfloat{\includegraphics[width=1in]{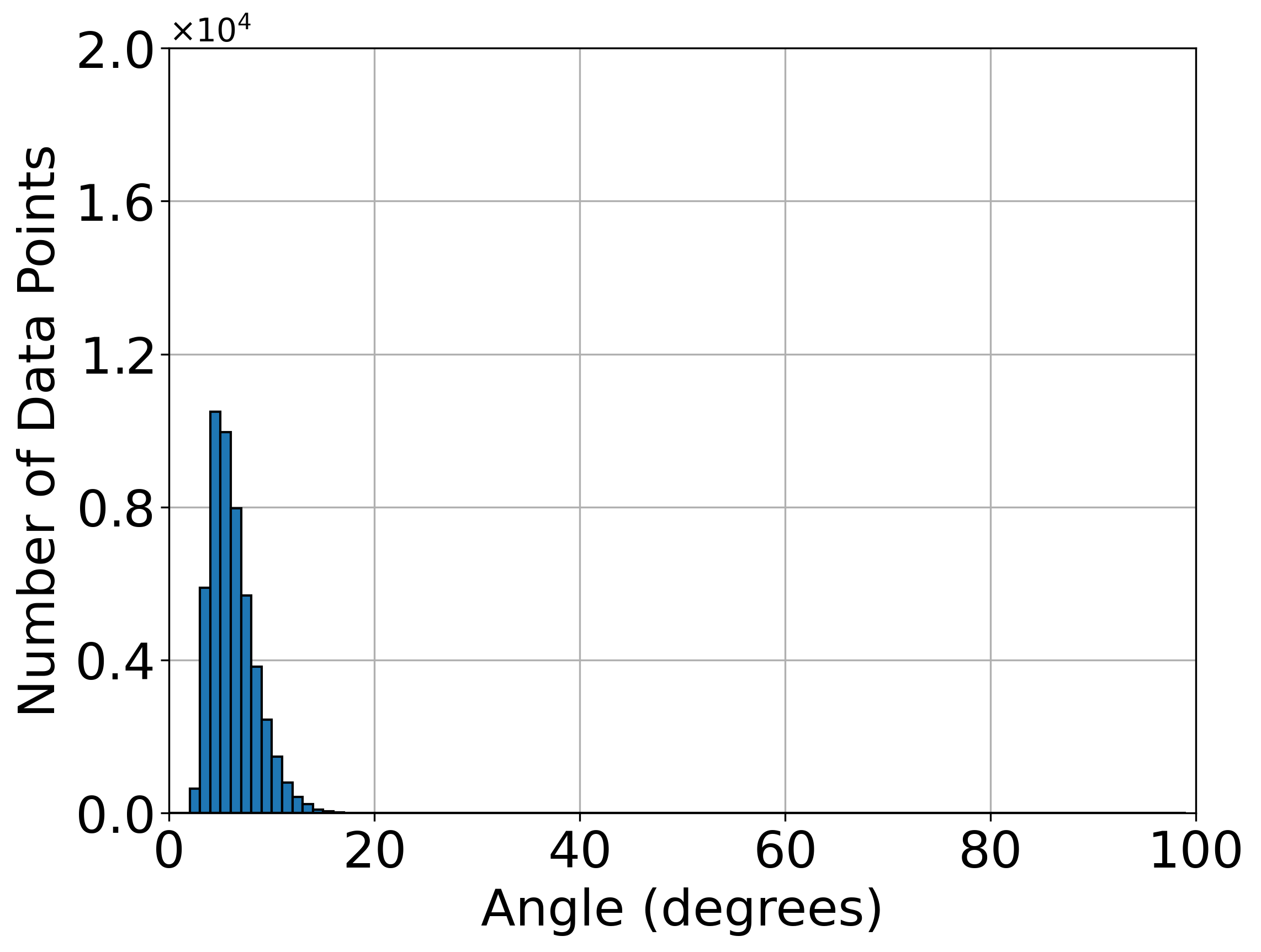}}
\hfil
\subfloat{\includegraphics[width=1in]{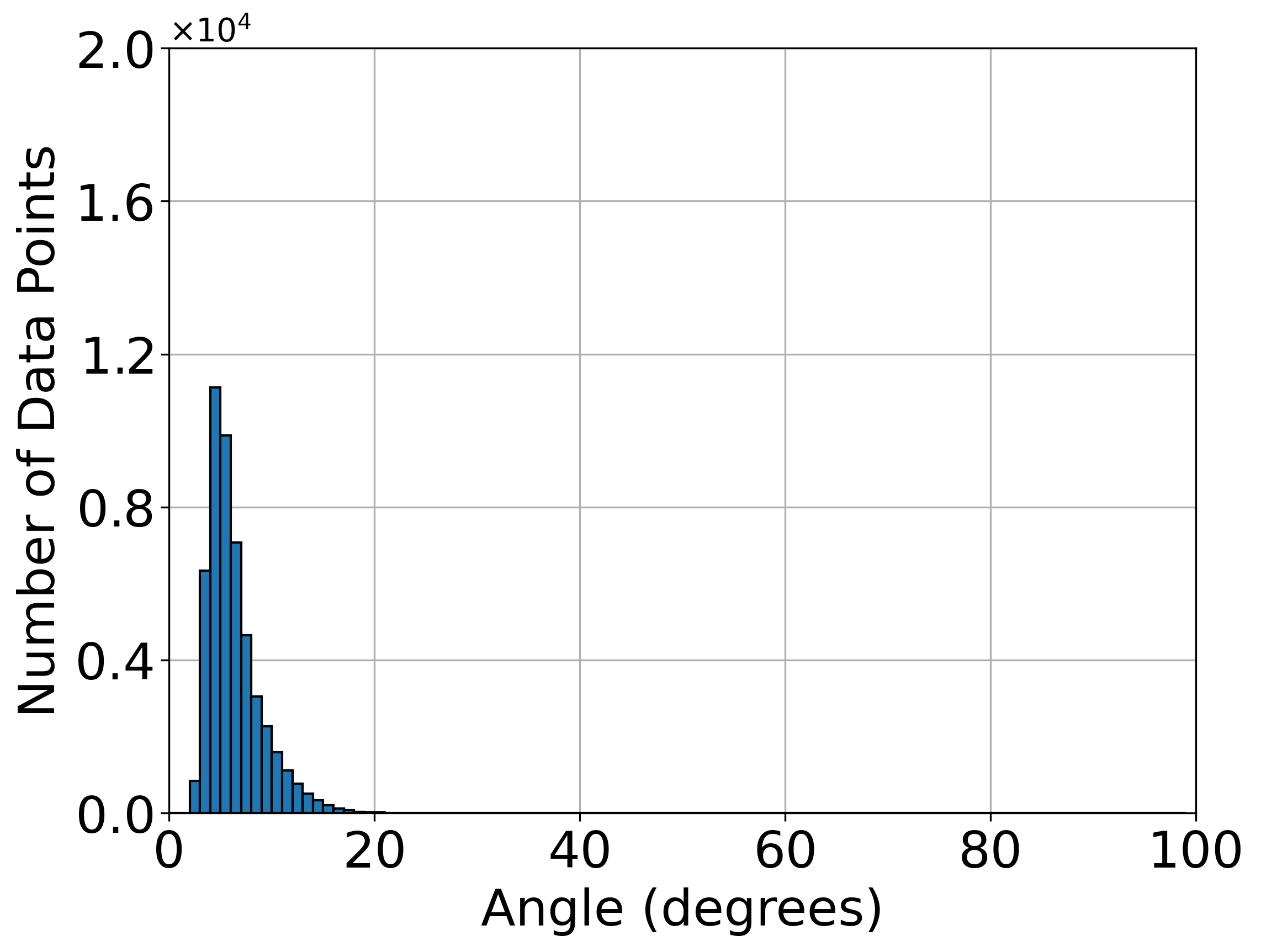}}
\hfil
\subfloat{\includegraphics[width=1in]{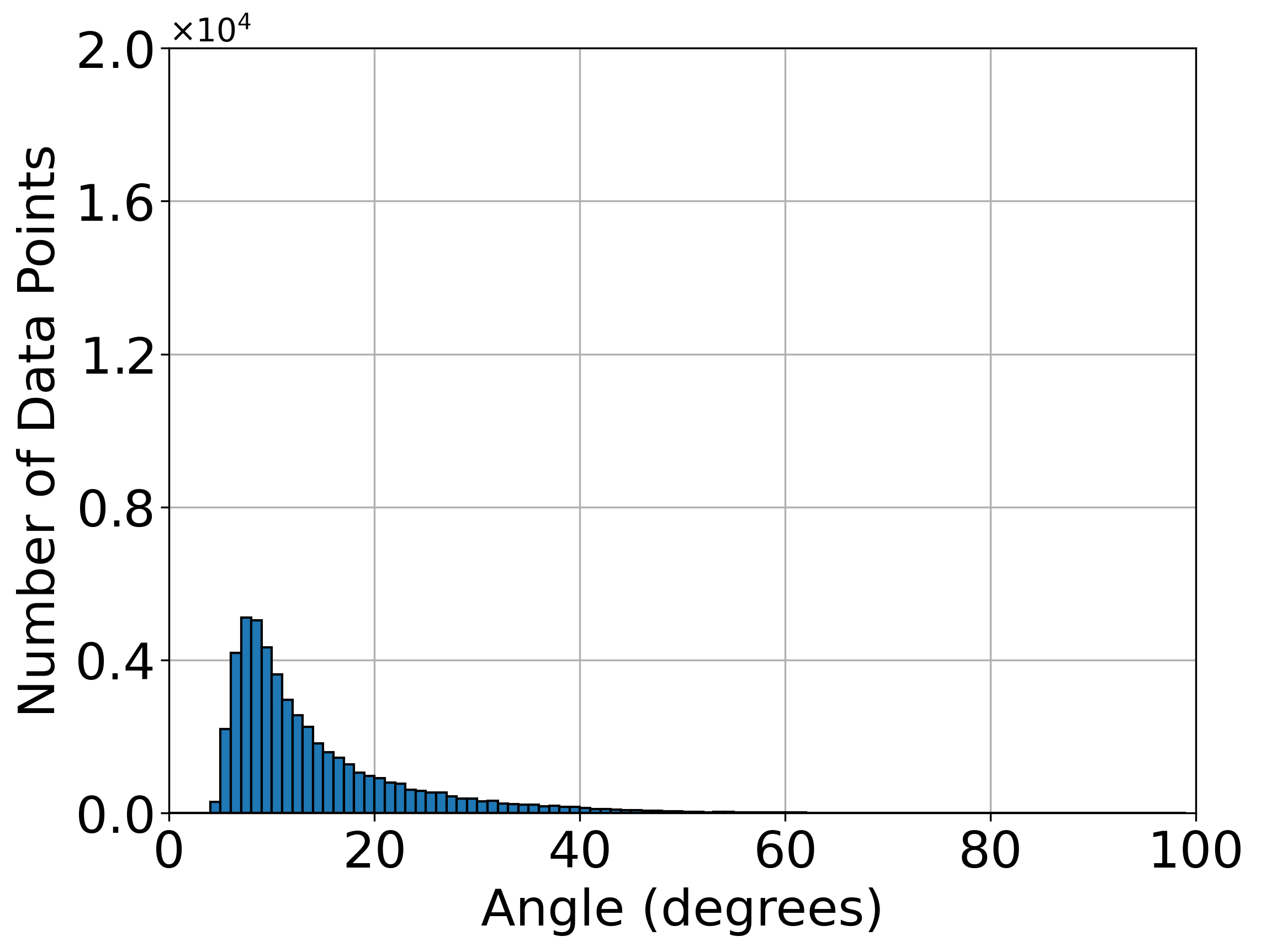}}
\hfil
\subfloat{\includegraphics[width=1in]{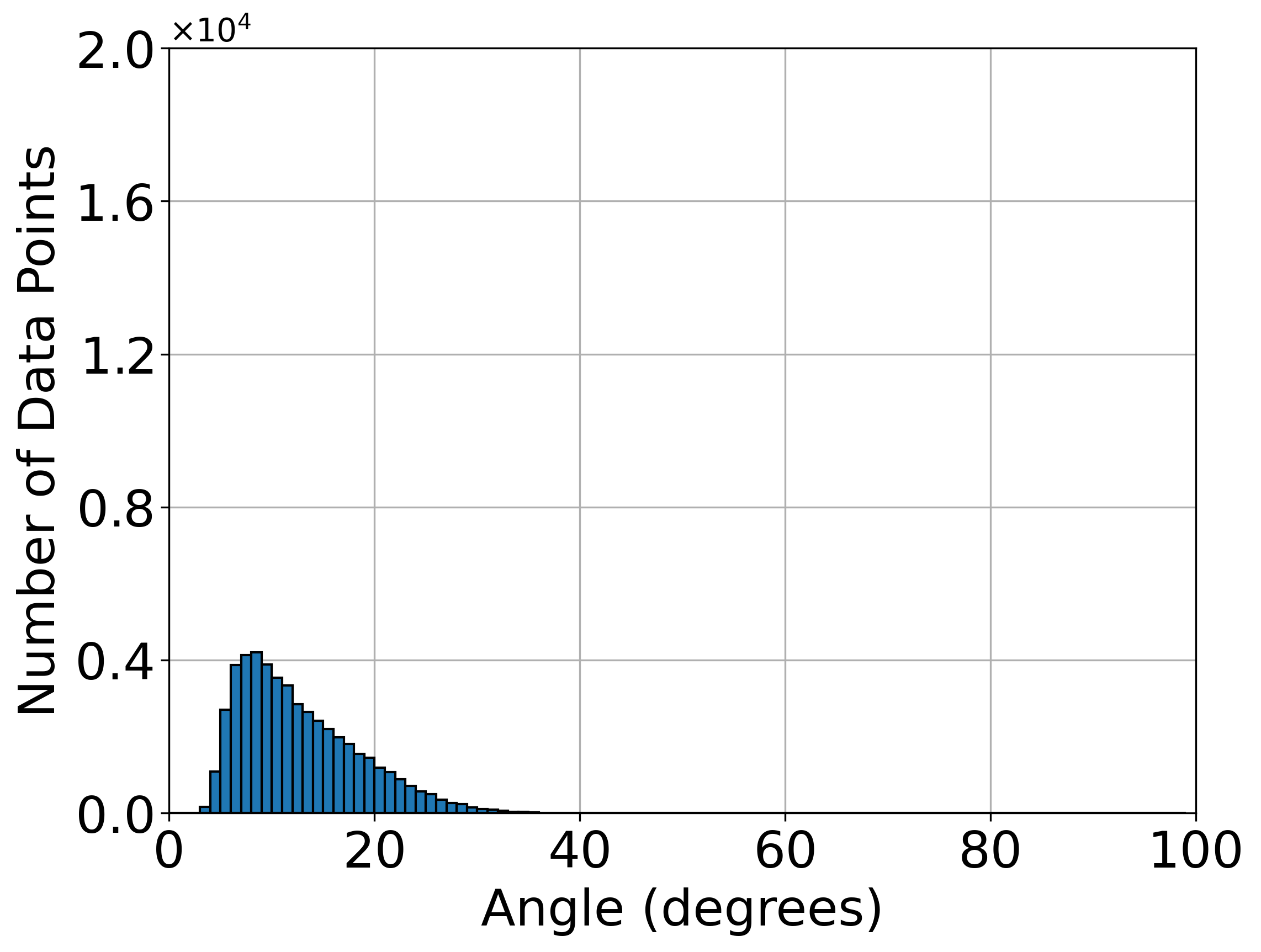}}
\hfil
\subfloat{\includegraphics[width=1in]{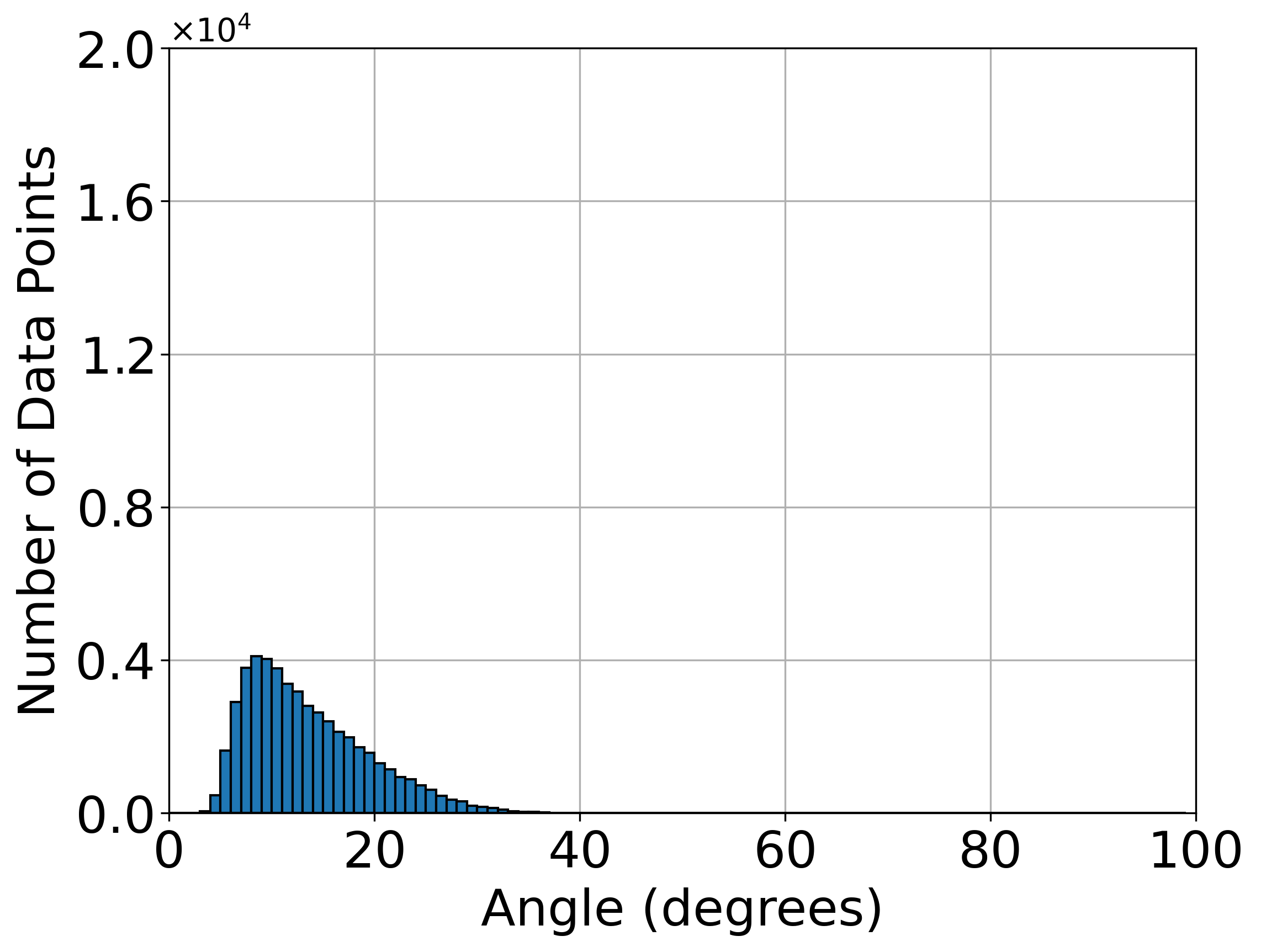}}
\\
\subfloat{\includegraphics[width=1in]{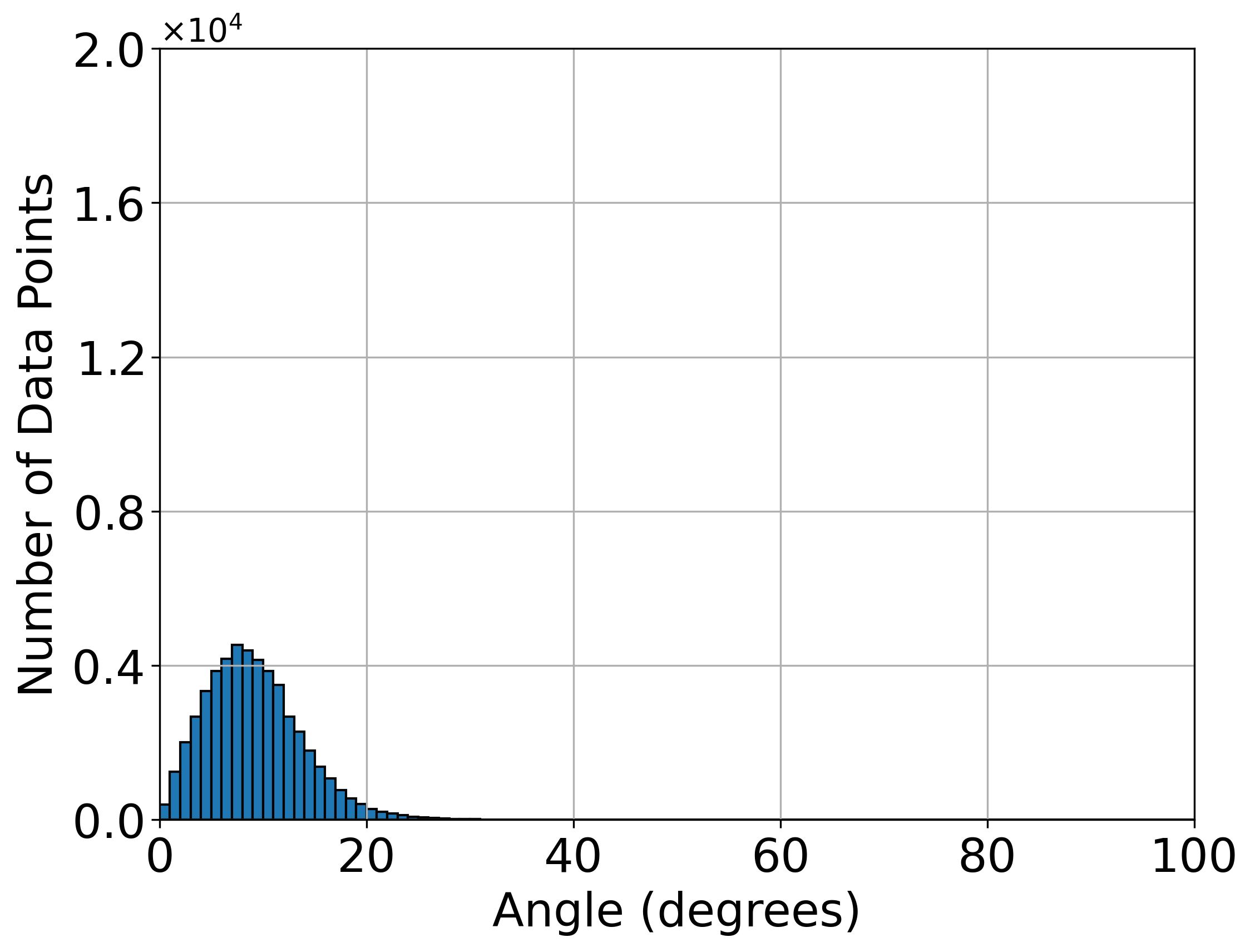}}
\hfil
\subfloat{\includegraphics[width=1in]{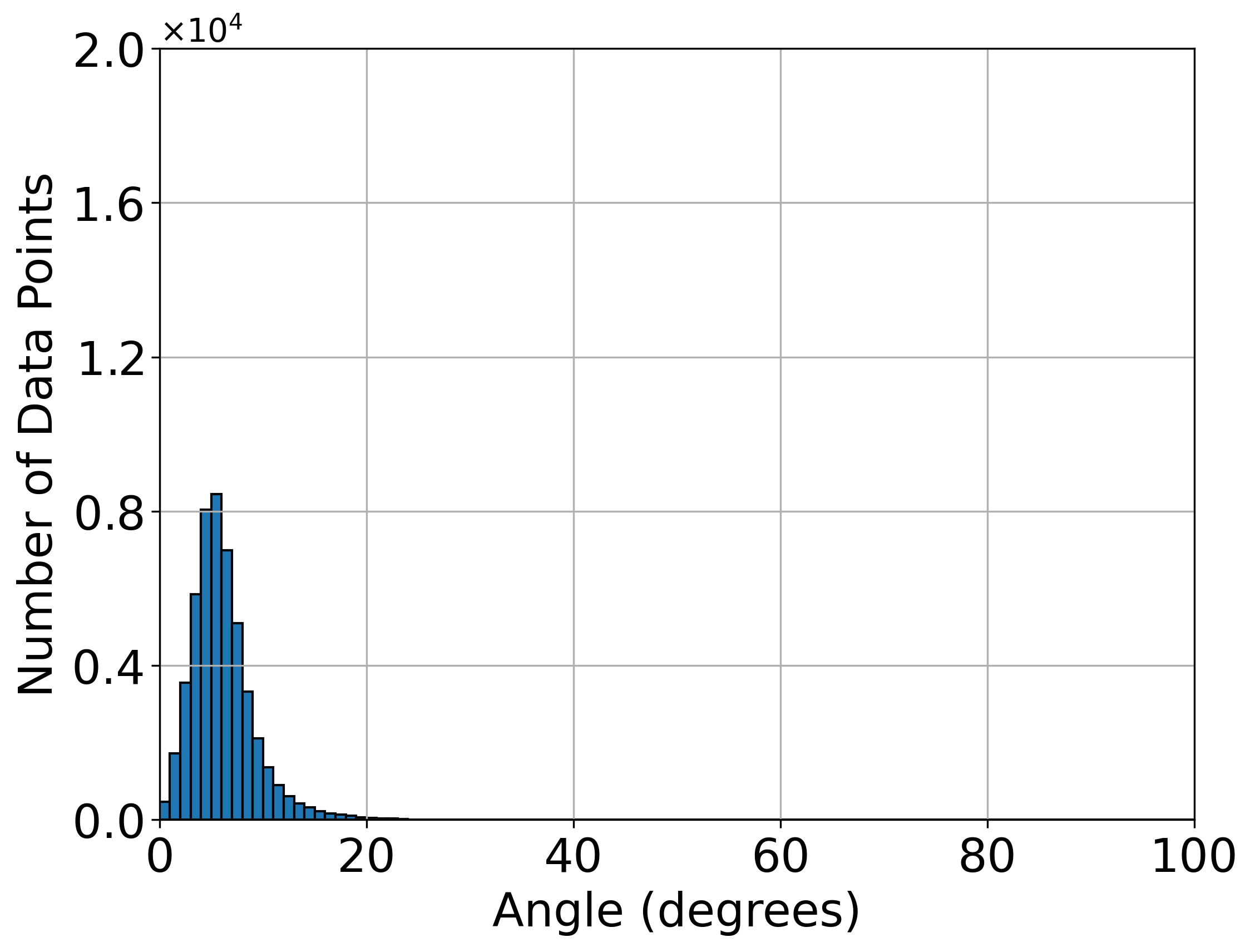}}
\hfil
\subfloat{\includegraphics[width=1in]{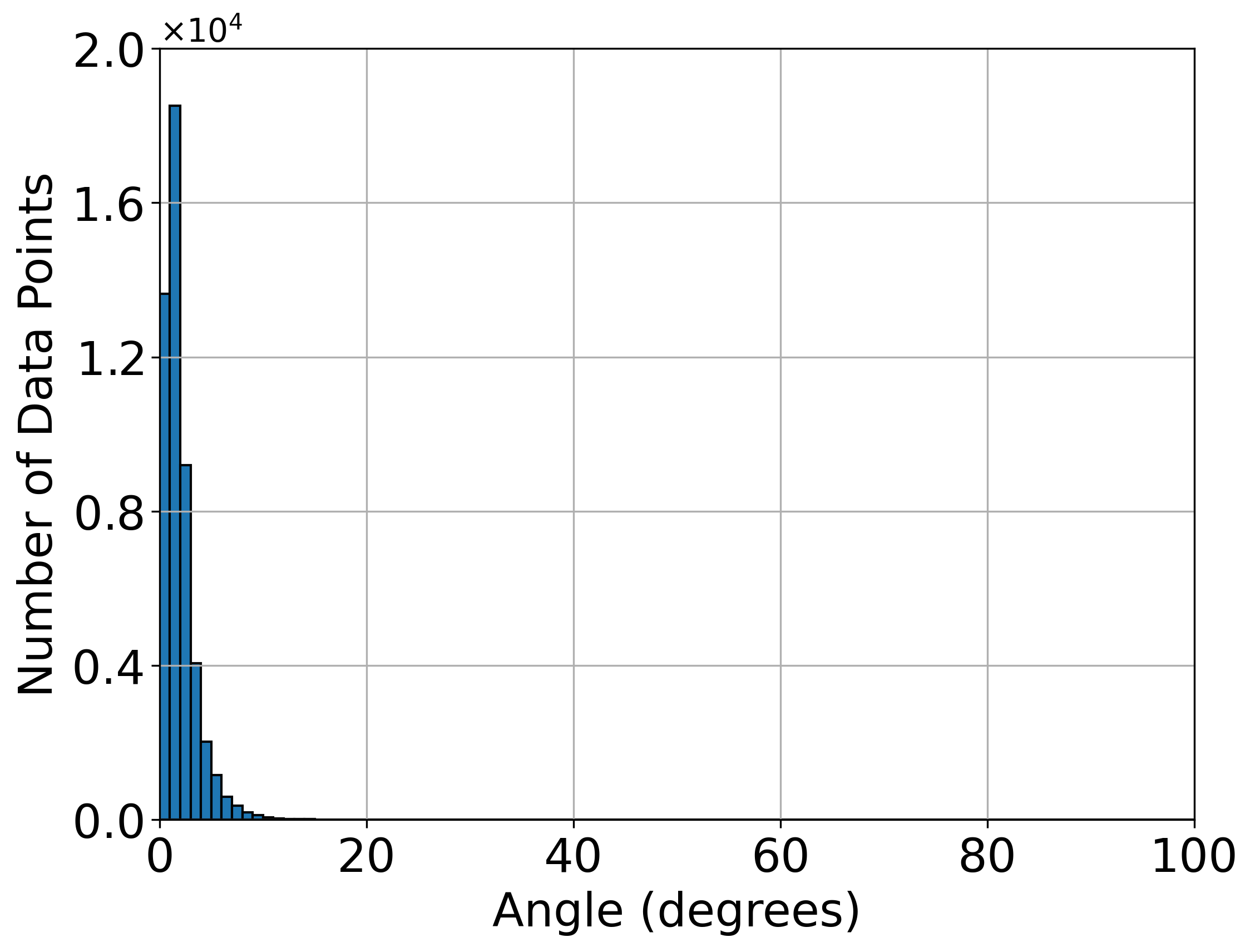}}
\hfil
\subfloat{\includegraphics[width=1in]{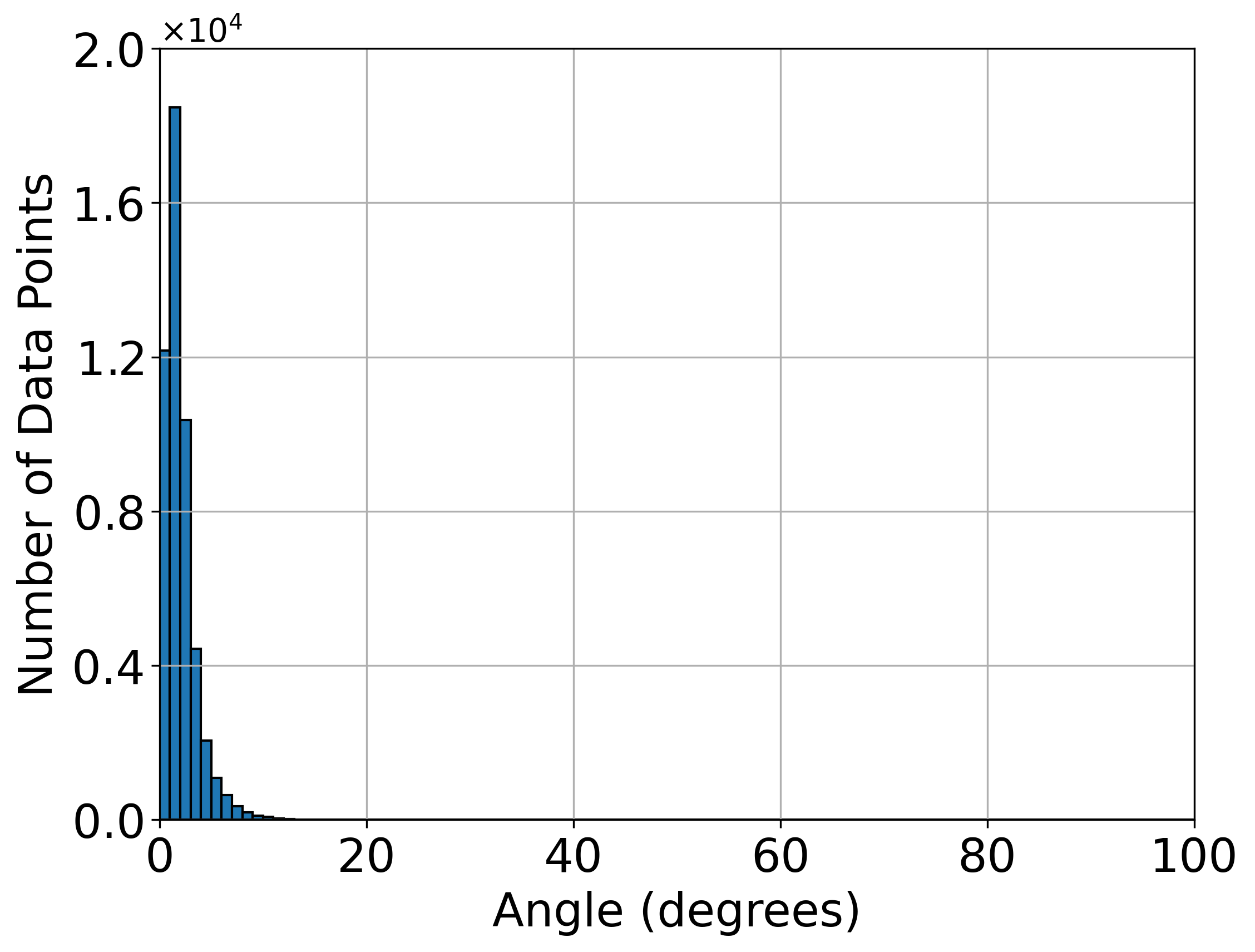}}
\hfil
\subfloat{\includegraphics[width=1in]{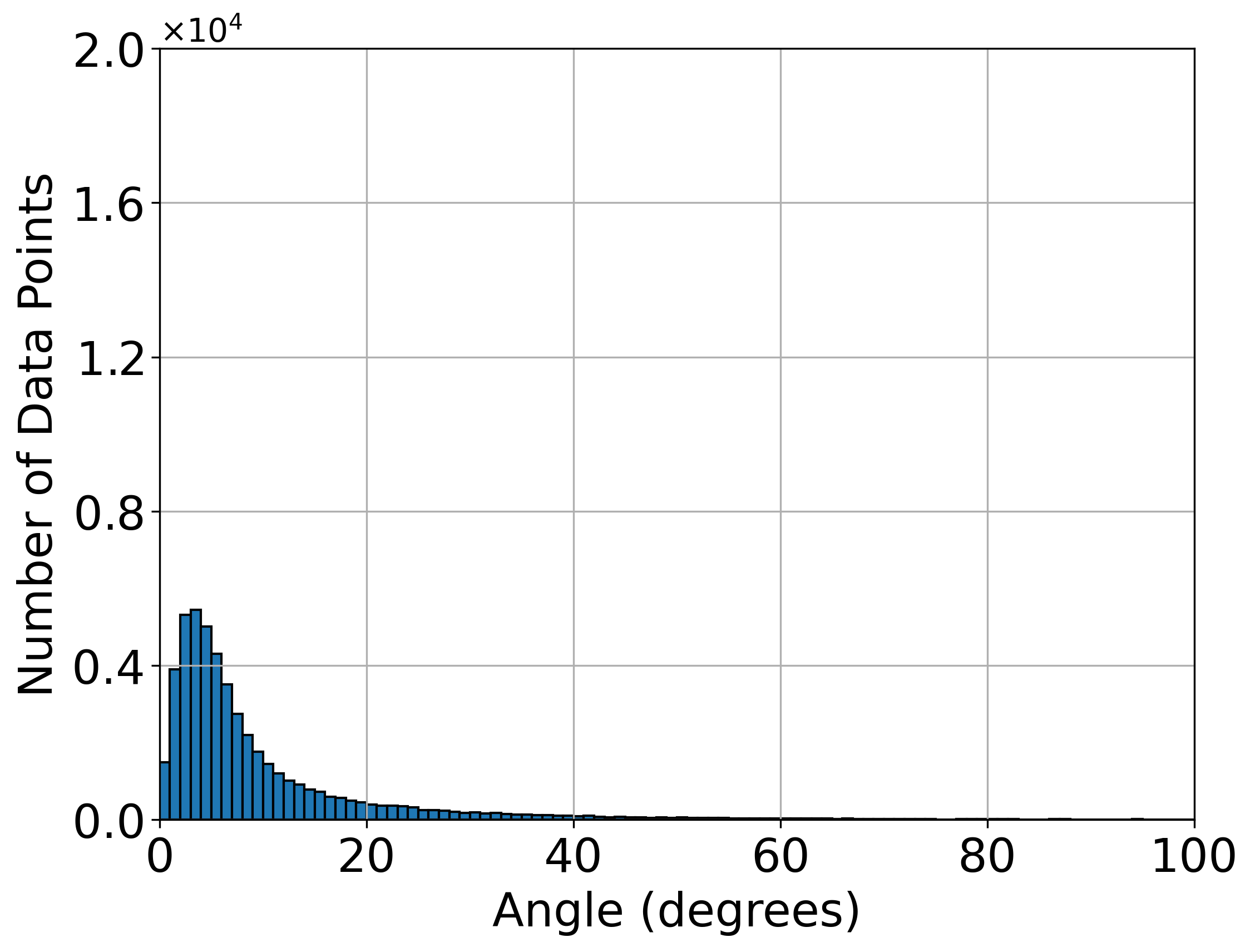}}
\hfil
\subfloat{\includegraphics[width=1in]{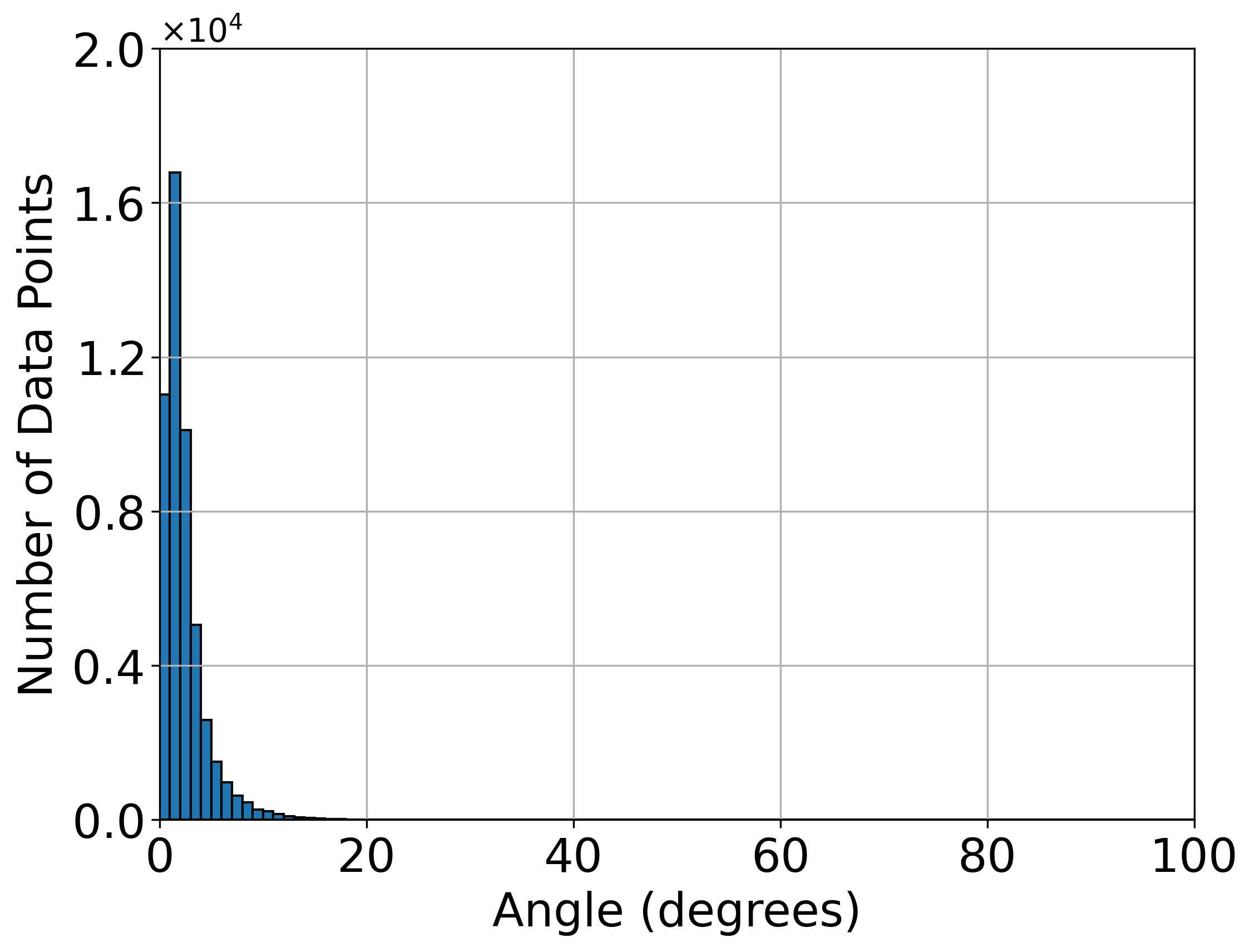}}
\hfil
\subfloat{\includegraphics[width=1in]{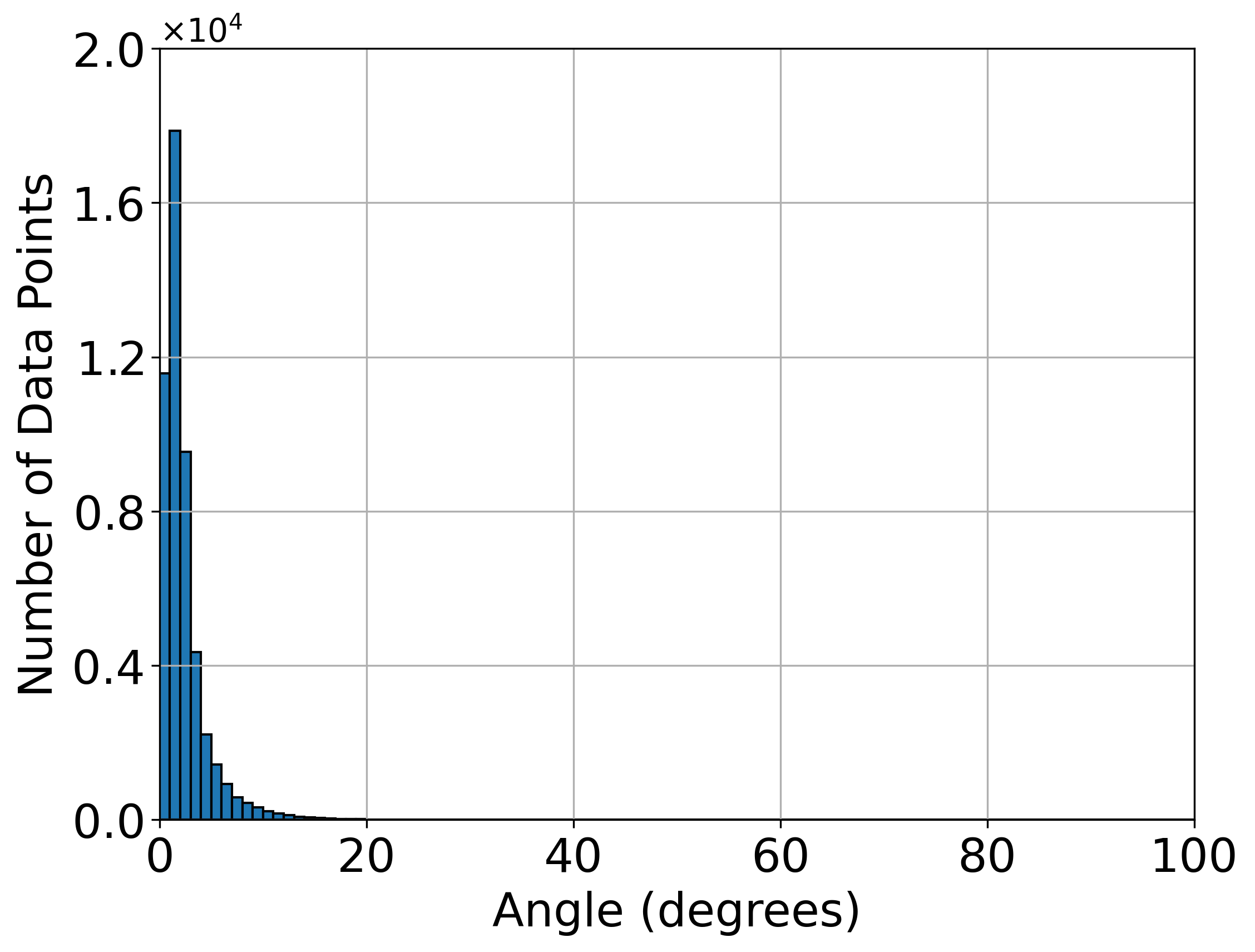}}
\\
\subfloat{\includegraphics[width=1in]{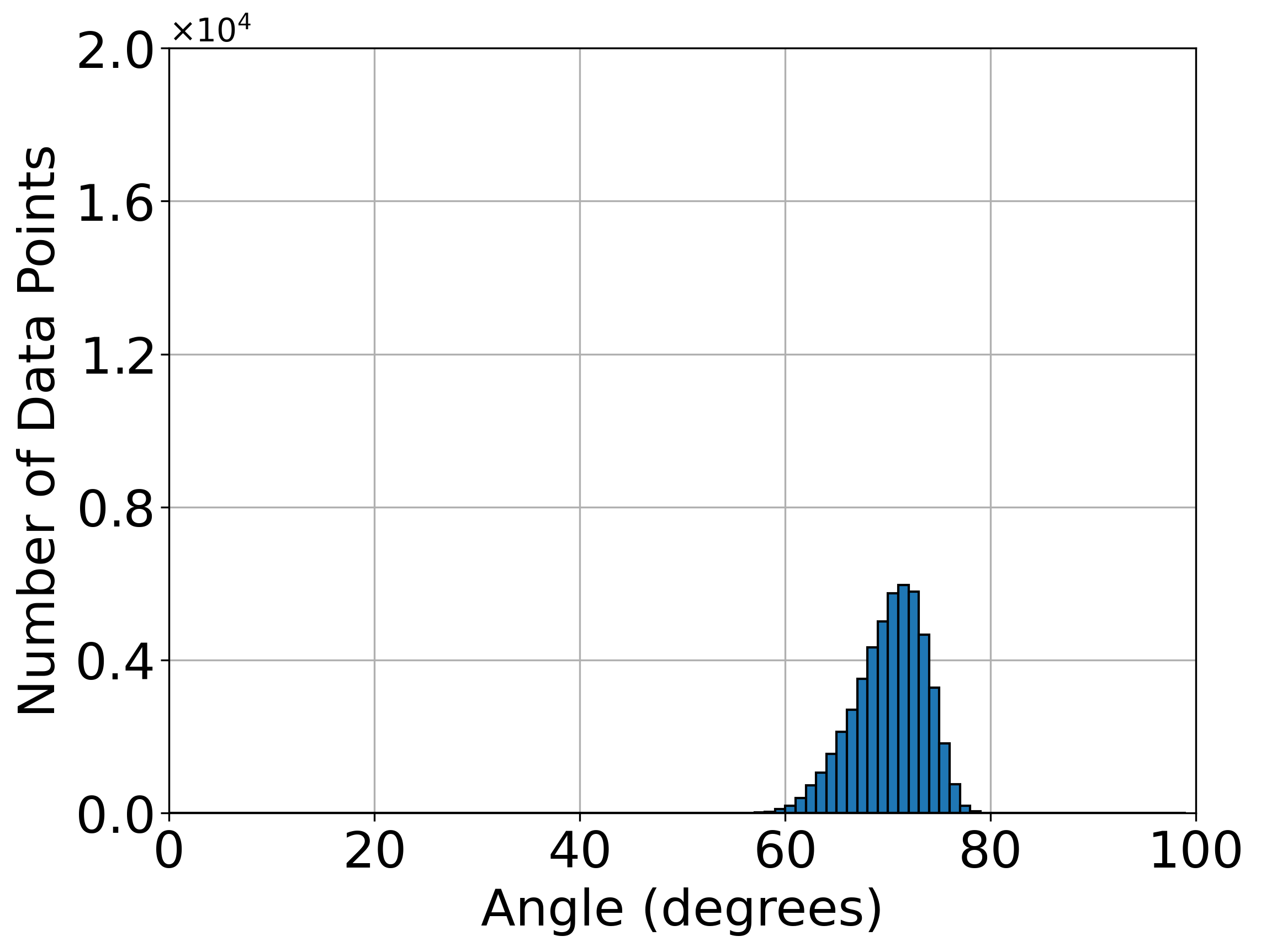}}
\hfil
\subfloat{\includegraphics[width=1in]{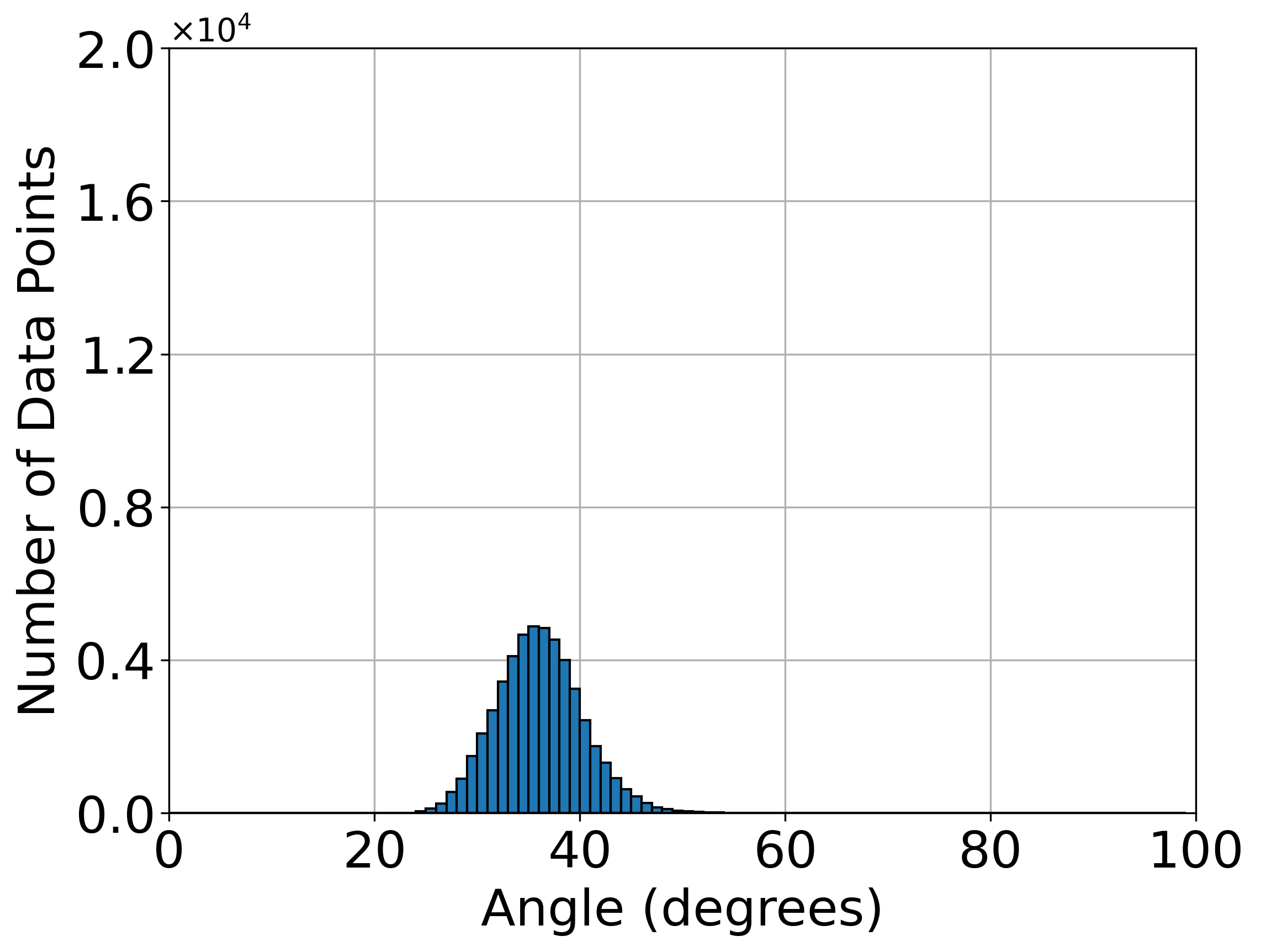}}
\hfil
\subfloat{\includegraphics[width=1in]{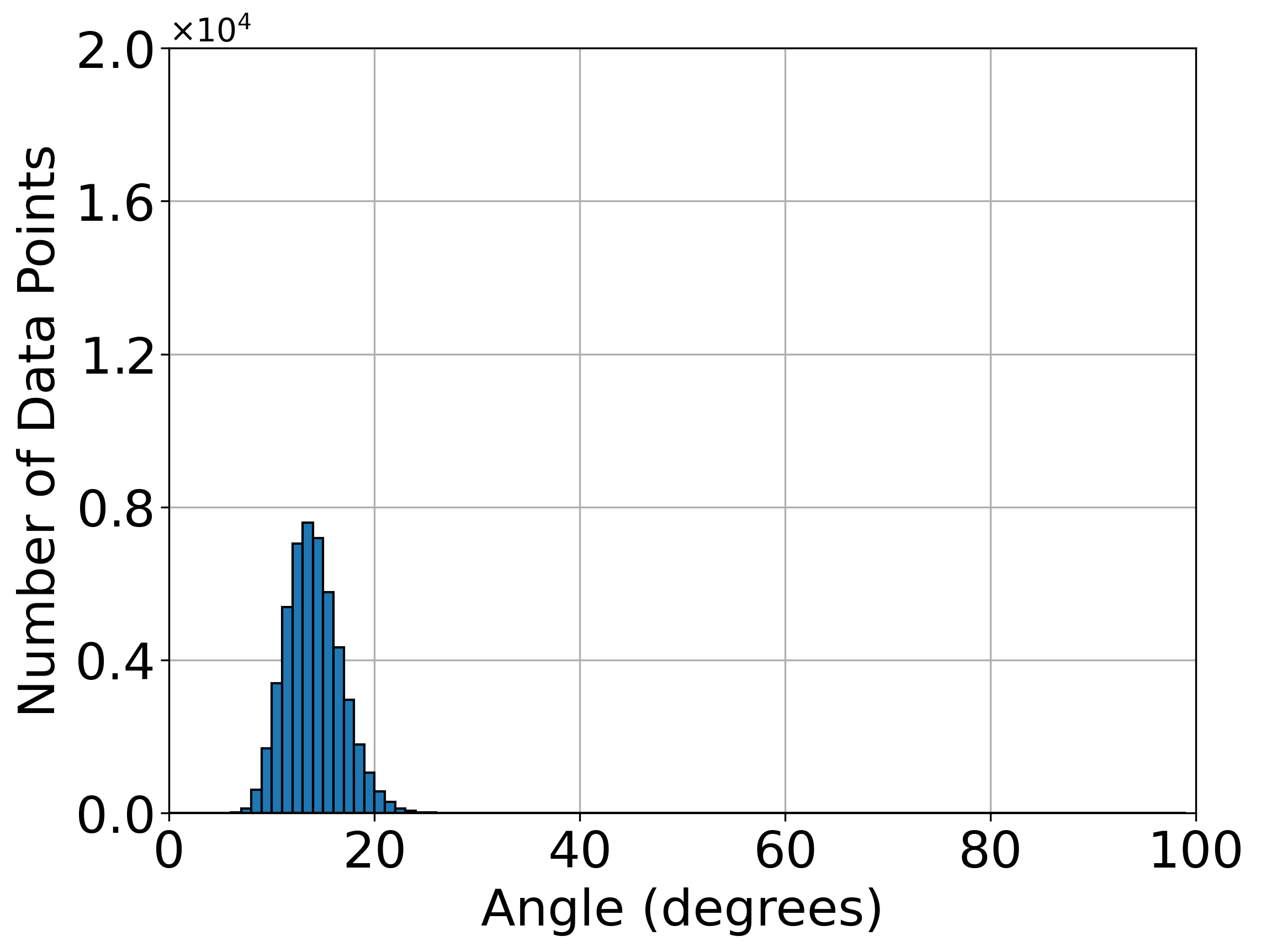}}
\hfil
\subfloat{\includegraphics[width=1in]{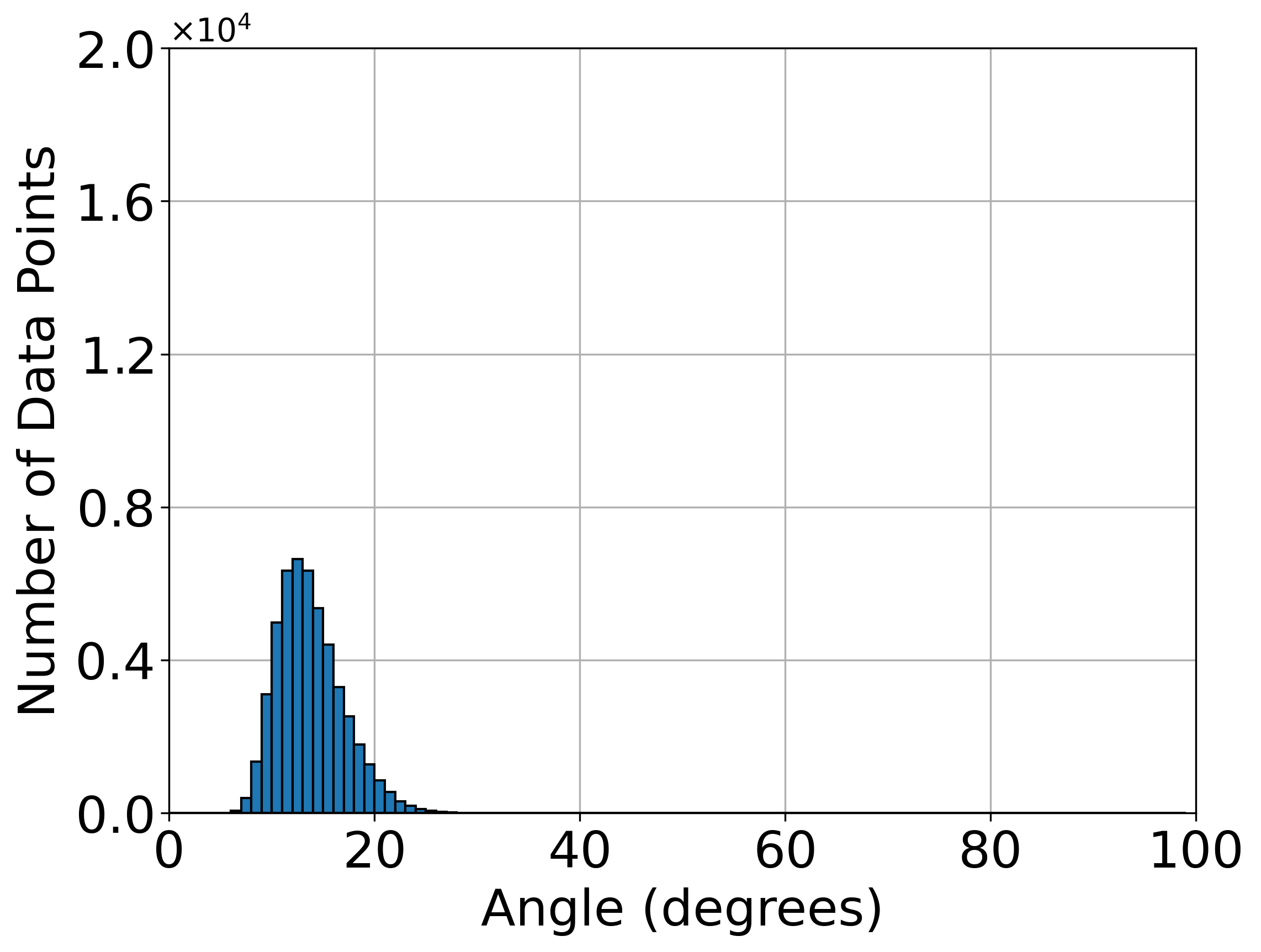}}
\hfil
\subfloat{\includegraphics[width=1in]{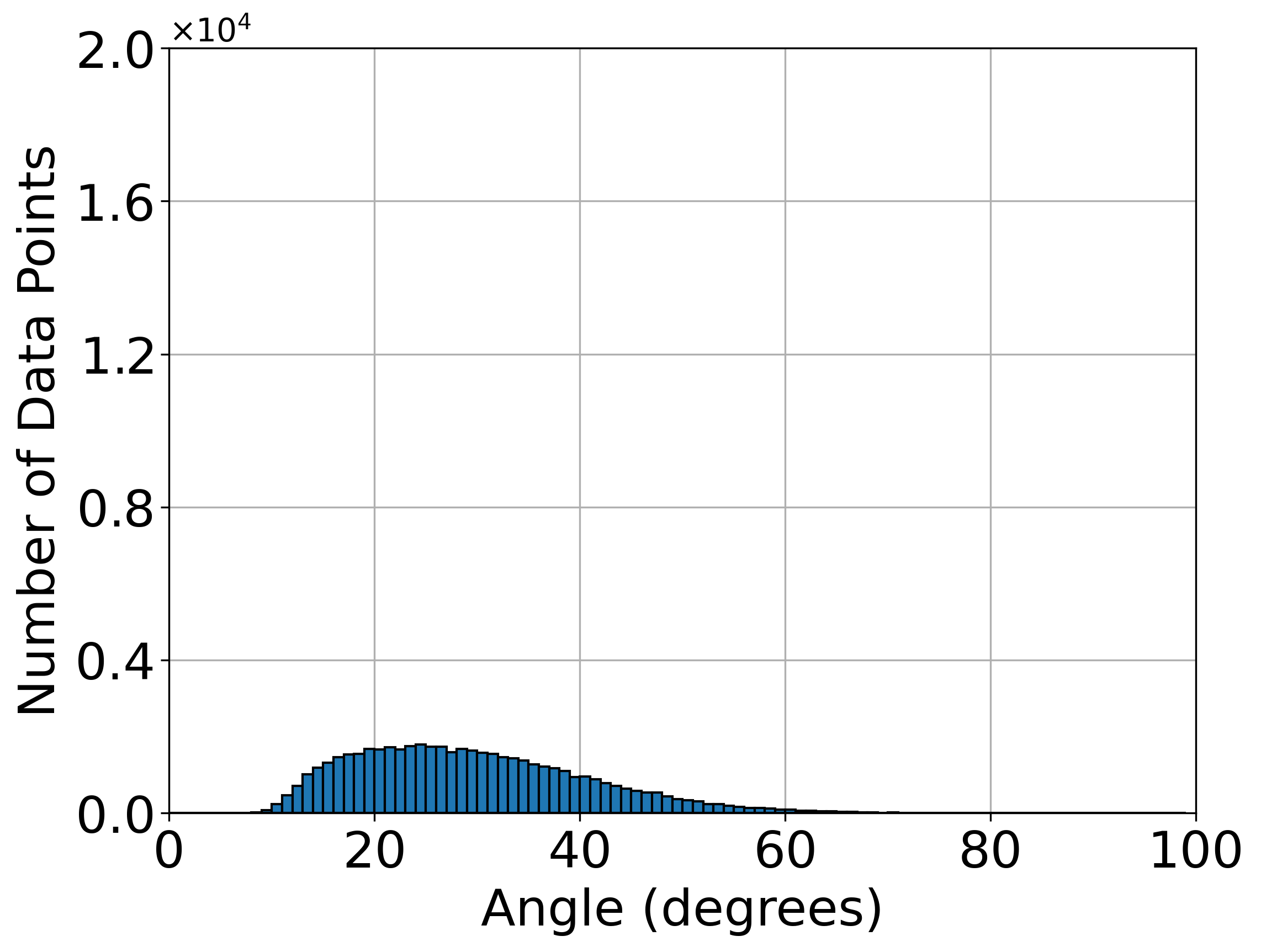}}
\hfil
\subfloat{\includegraphics[width=1in]{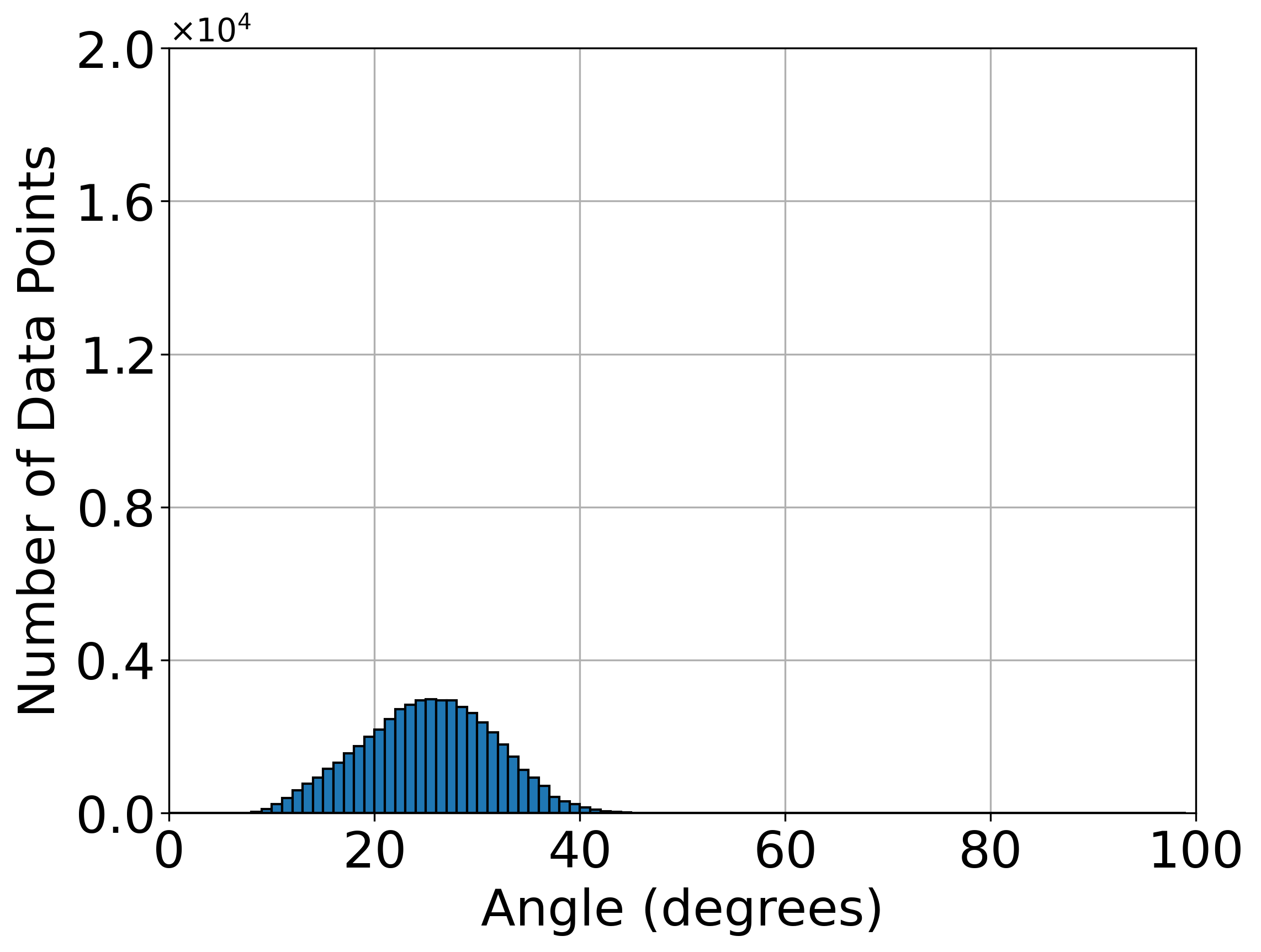}}
\hfil
\subfloat{\includegraphics[width=1in]{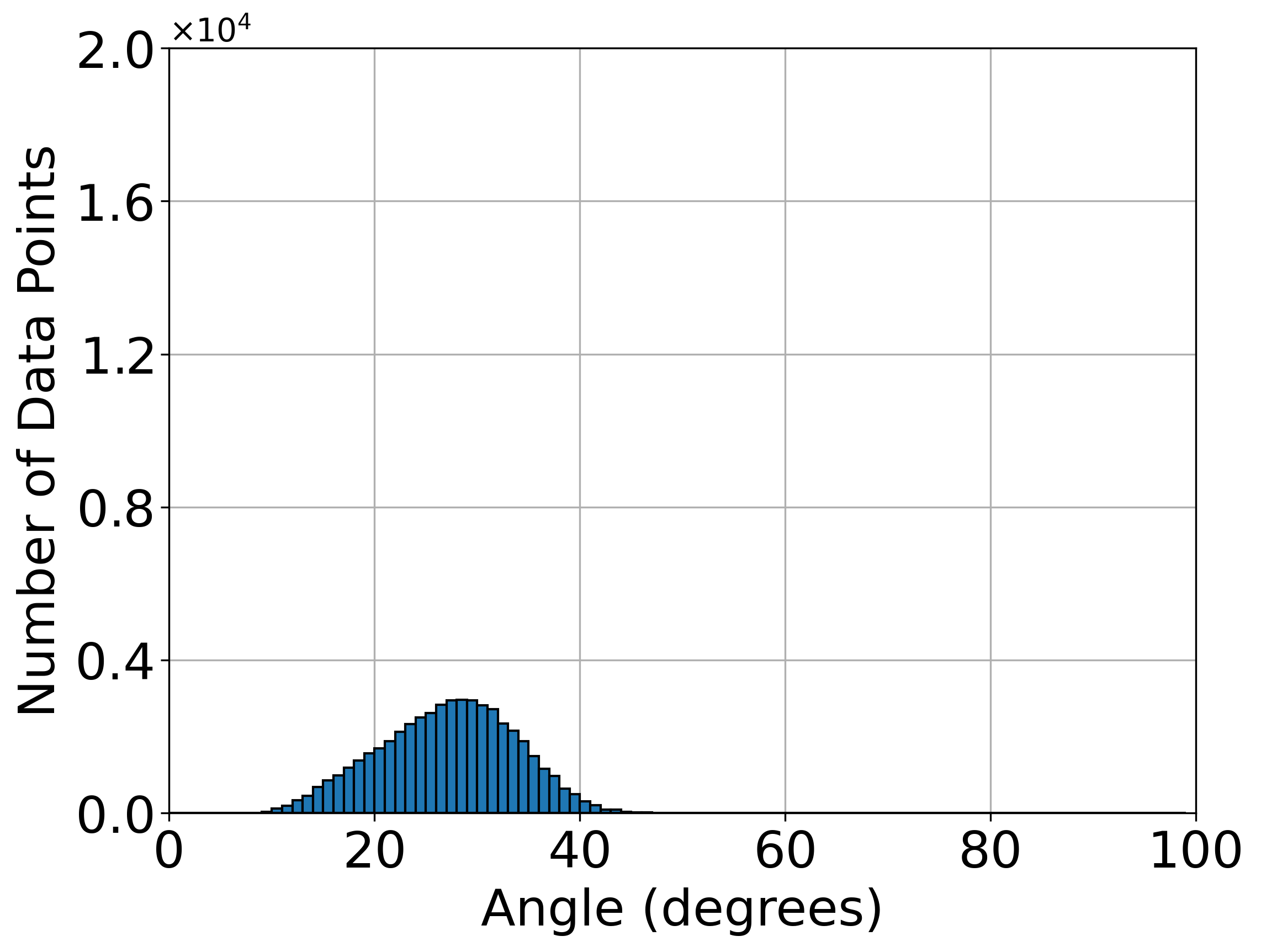}}
\\
\setcounter{subfigure}{0}
\subfloat[CE]{\includegraphics[width=1in]{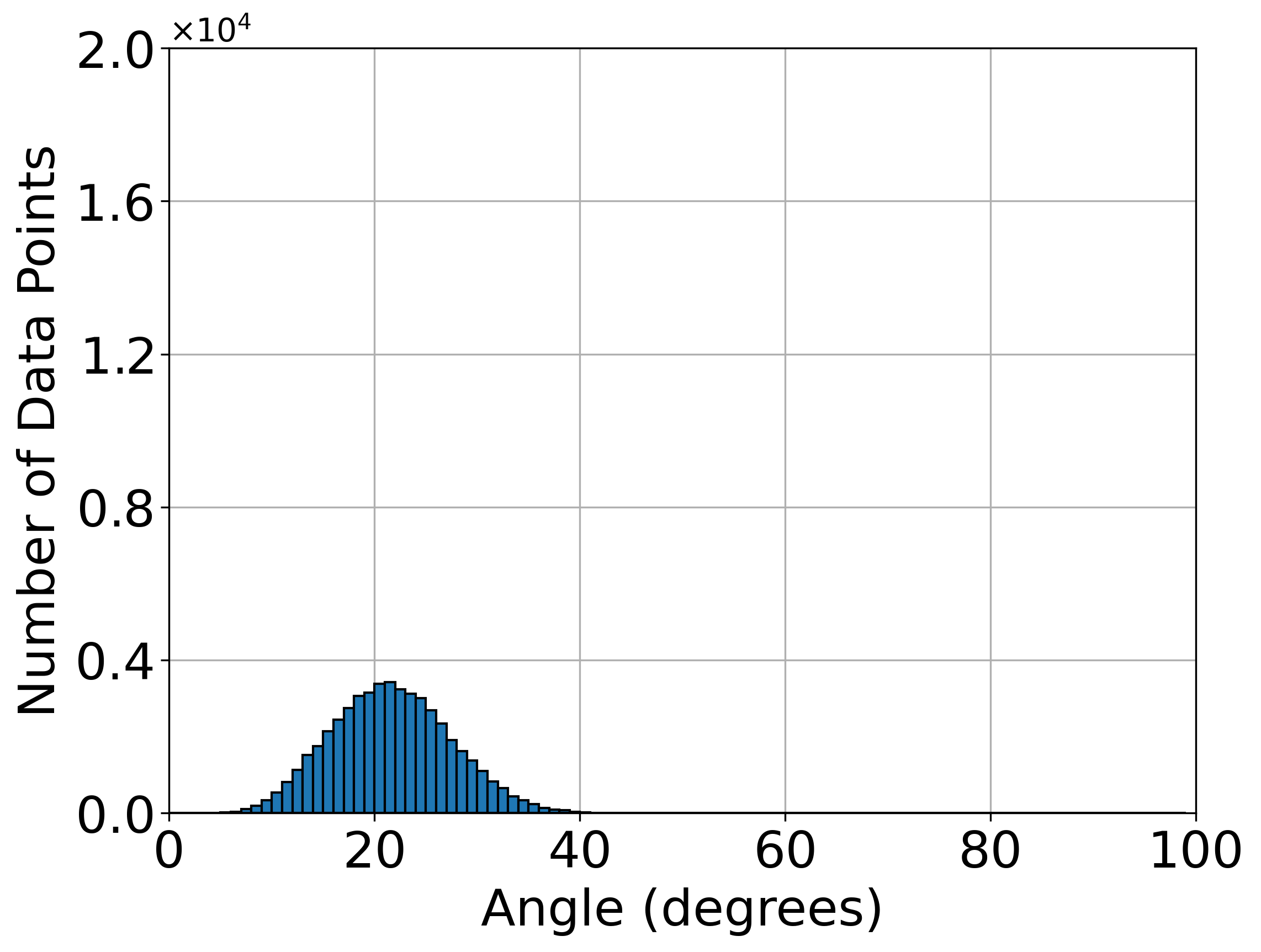}}
\hfil
\subfloat[CL]{\includegraphics[width=1in]{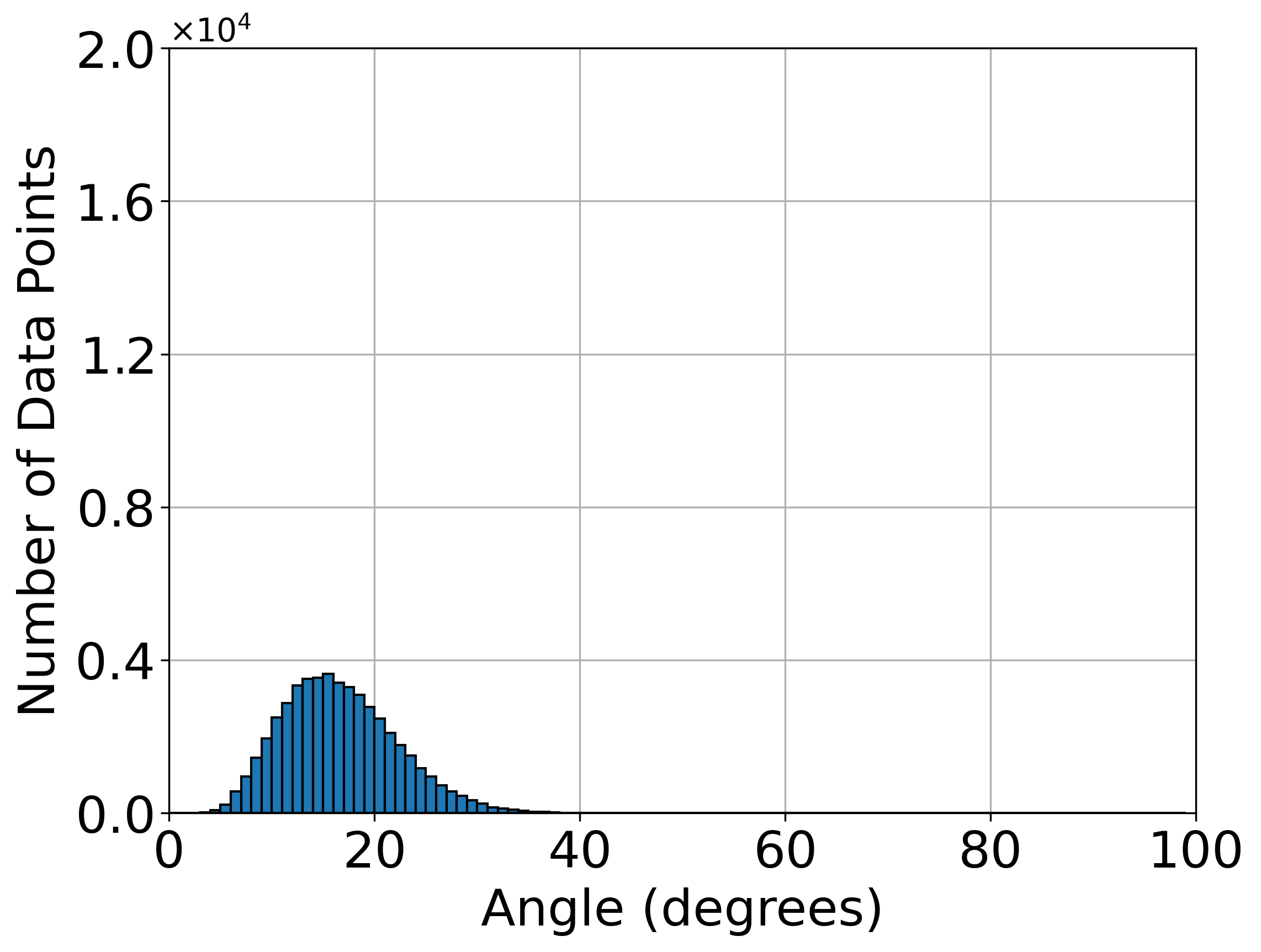}}
\hfil
\subfloat[CCL]{\includegraphics[width=1in]{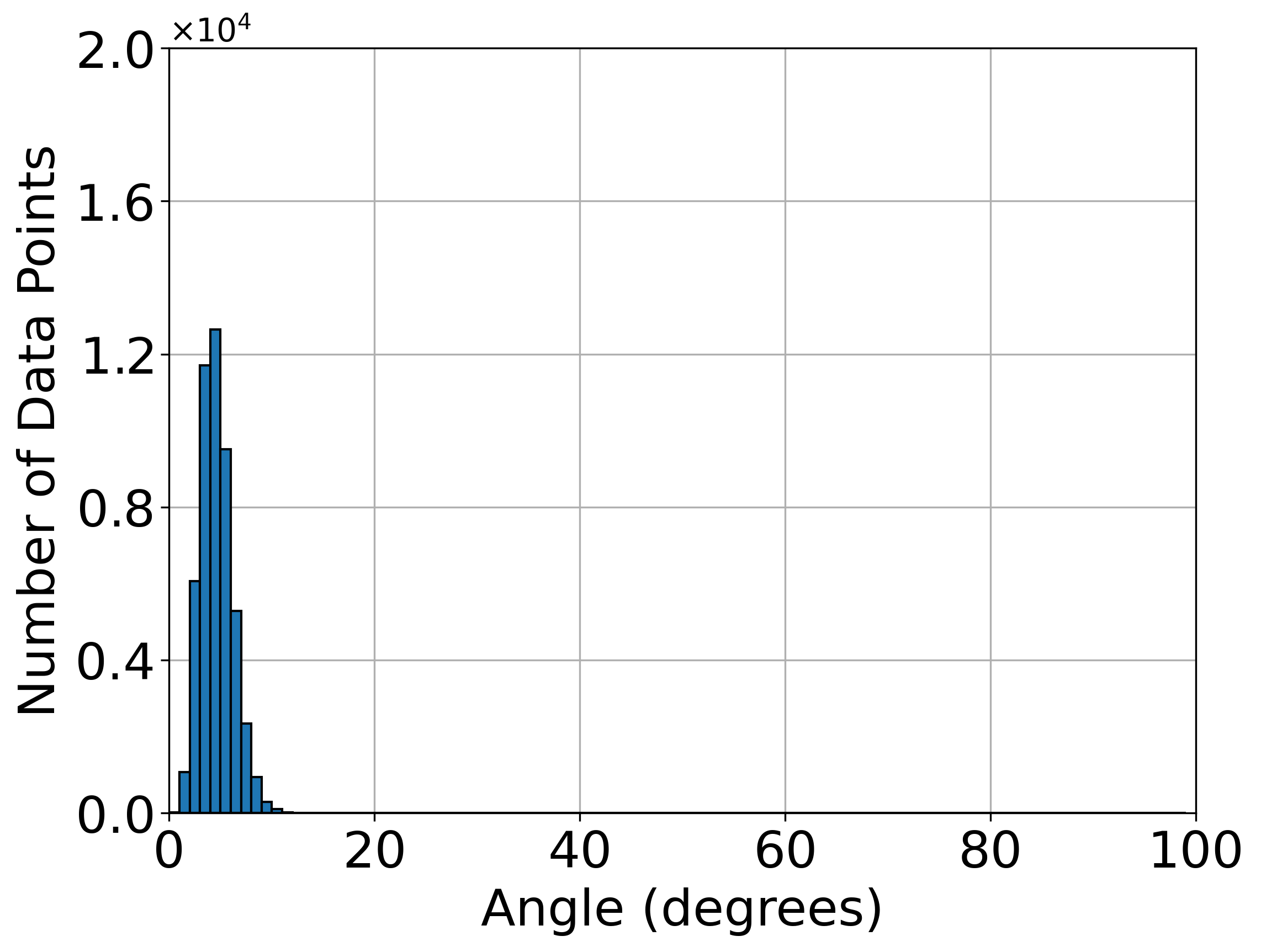}}
\hfil
\subfloat[SCCL]{\includegraphics[width=1in]{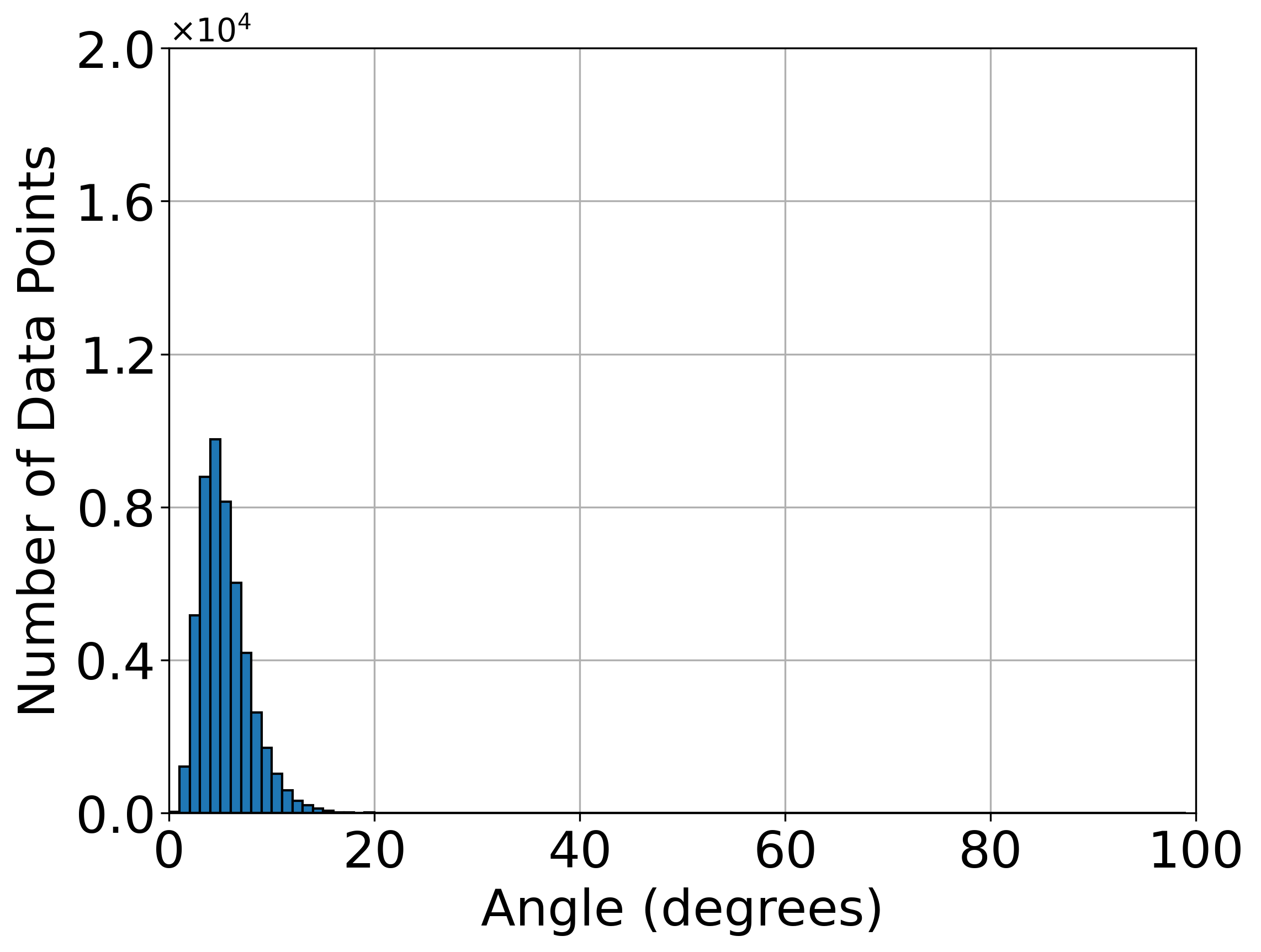}}
\hfil
\subfloat[CPN]{\includegraphics[width=1in]{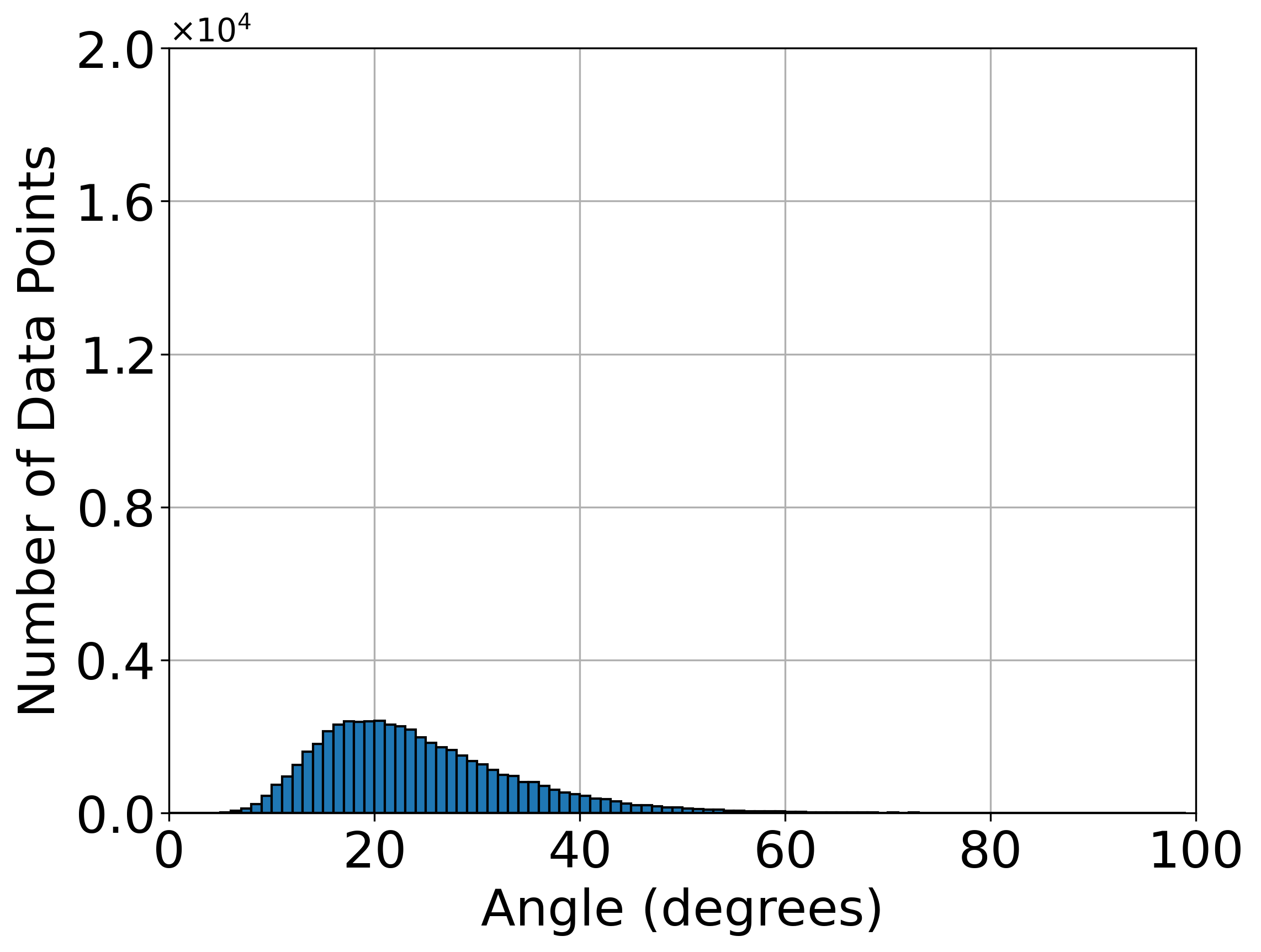}}
\hfil
\subfloat[DPP]{\includegraphics[width=1in]{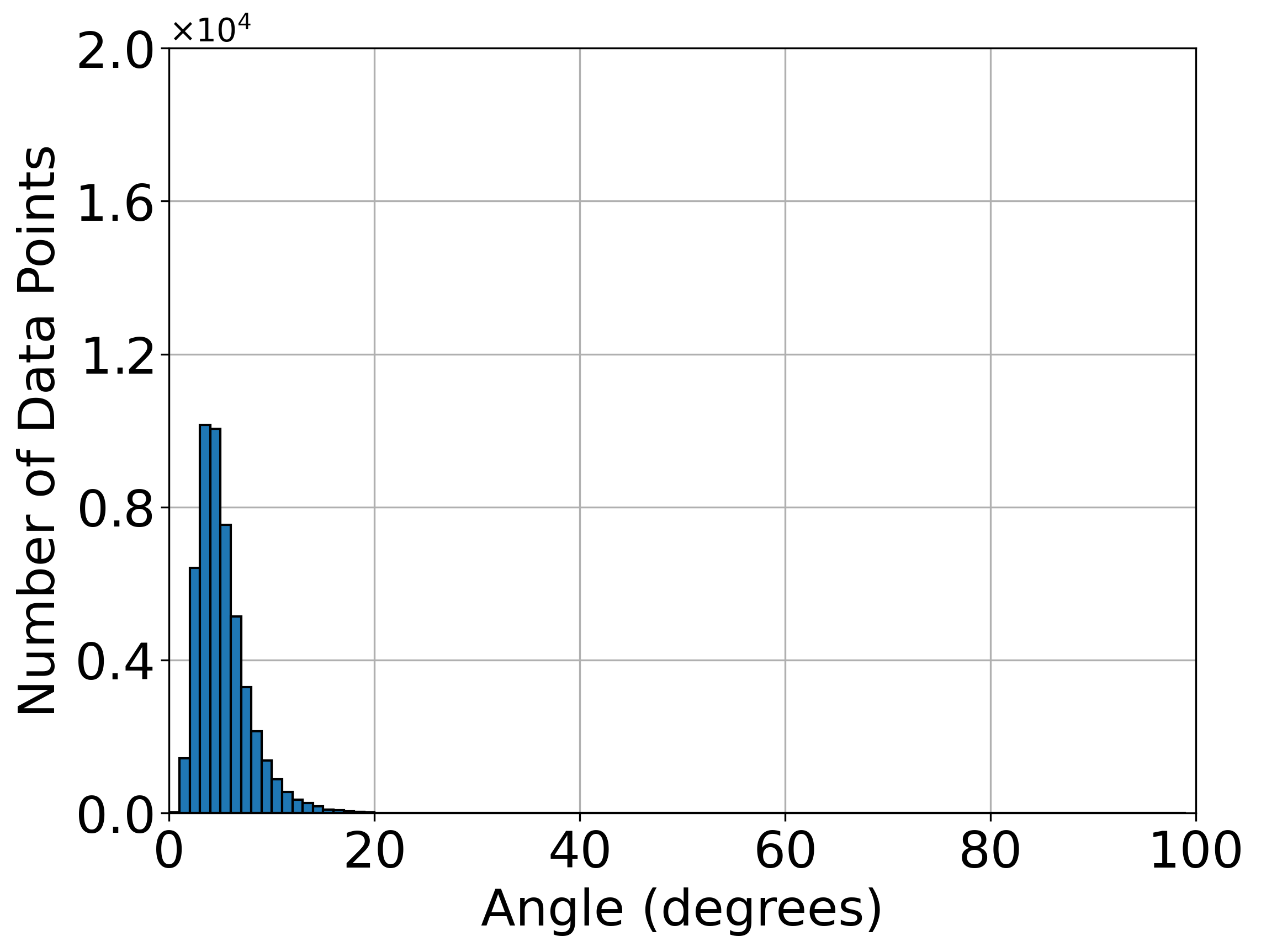}}
\hfil
\subfloat[DPNP]{\includegraphics[width=1in]{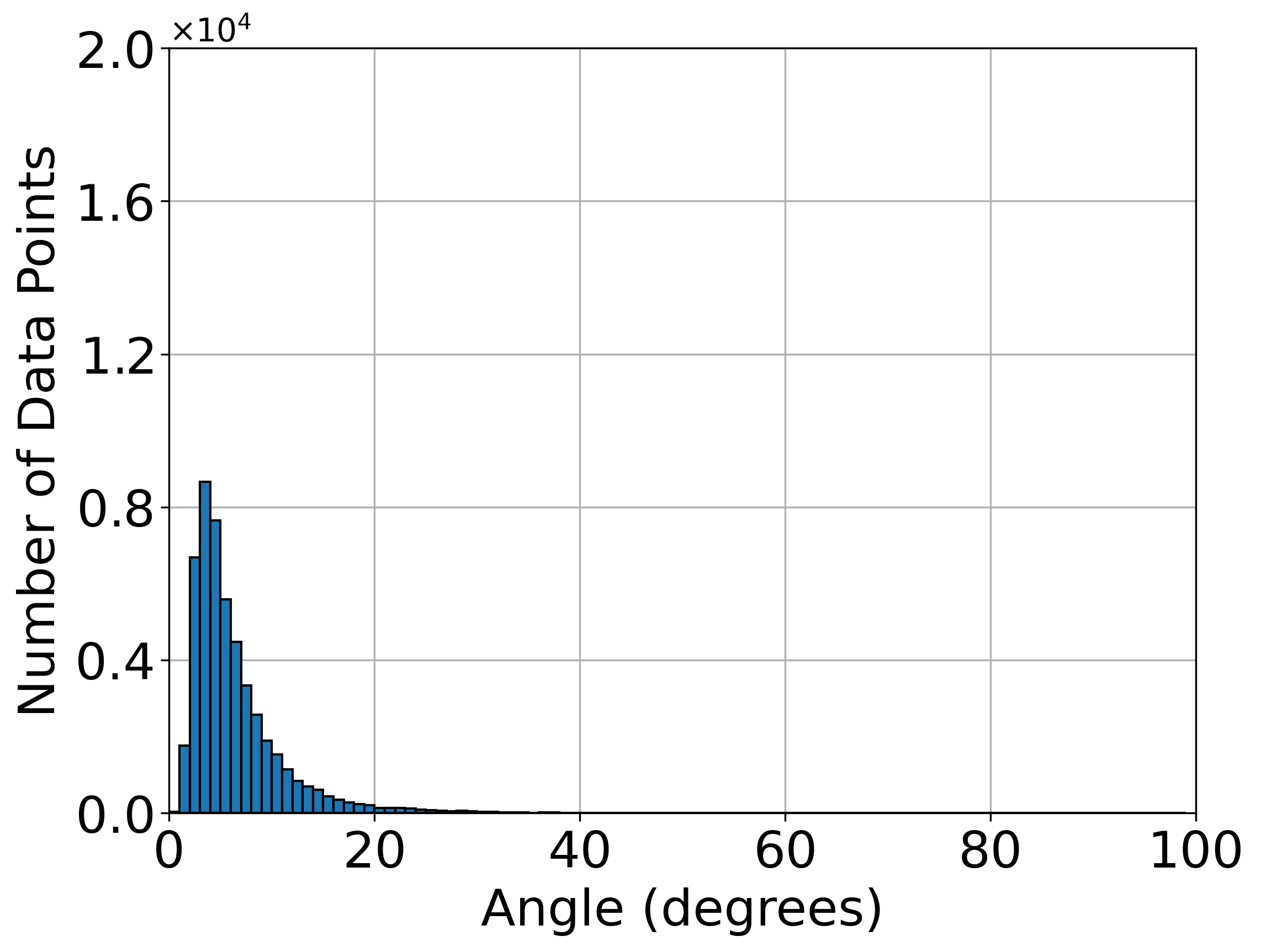}}
\caption{Histogram of intra-class angles denotes the resulted compactness from different methods: the first two rows correspond to CIFAR-10 and the bottom two rows to CIFAR-100. Standard ResNet-18 architecture is used in odd rows, while reduced ResNet-18 is used in even rows. The proposed models show concentrated histograms similar to CL, CCL, and SCCL which all use center pulling terms in the loss function to obtain better compactness.}
\label{fig:WC}
\end{figure*}

To assess intra-class compactness resulted from different methods, the intra-class angle $\psi_{i}$ (in degrees) between $i^{th}$ training data point and its respective class center is first calculated as:
\begin{equation}
\label{eq16}
\psi_{i}=arccos(c_{y_i}^{T}h(x_i; \theta))\times\left(\frac{180}{\pi}\right)
\end{equation}
Then, the histogram of these intra-class angles is plotted in Figure \ref{fig:WC} to complement the previously discussed inter-class separation results. Here, a more left-concentrated histogram is preferred. 

CE shows a bell-shaped angle distribution with a larger mean compared to the others, indicating weaker intra-class compactness. This is expected since all other methods use center pulling terms in their loss function. However, the most obvious difference is between odd and even rows which compare network architectures. It can be seen that using standard ResNet-18 and working in higher-dimensional feature spaces does not lead to compactness, as the classifier is not that much in pressure toward center alignment of the data. This is directly affected by the value of $\lambda_{\text{center}}$ or its equivalent $\lambda_{\text{pos}}$ in our case; which is usually kept small enough to let CE increase the accuracy. This fact is apparent in the output of all models with standard ResNet-18 architecture (odd rows of Figure \ref{fig:WC}), as the mean of the histogram is shifted to the left with respect to the base case of CE; but still the histogram's mean is noticeably larger than the corresponding value in the next row. When the reduced architecture is used and the feature space is squeezed into much lower dimensions, then different classes are more easily mixed up and the classifier models have to compress data samples much more toward their centers; as it can be seen in the even rows of Figure \ref{fig:WC}. DPP and DPNP models act very similar in the sense of compactness across different data sets and architectures, which shows adding negative prototypes and the repulsion terms in DPNP did not hurt the desired compactness of the DPP model. However, CCL and SCCL marginally obtain better compactness in higher dimensions which is at the price of less separation already discussed above.

Different terms in the loss function may interfere with each other and better results in the sense of separation, compactness or even accuracy may not lead to the best outcomes in the others. 
Therefore, to evaluate how each model manages to reach a proper trade-off between separation and compactness, inspired by Fisher Linear Discriminant, the Separation to Compactness Ratio (SCR) is calculated as follows:
\begin{equation}
\label{eq17}
\text{SCR} = \frac{1}{M} \sum_{j=1}^M \frac{\min_{k \neq j} \| c_k - c_j \|}{\frac{1}{n_{j}} \sum_{i=1}^N \mathbbm{1}\{y_i=j \}\| h(x_i; \theta) - c_j \|}
\end{equation}
where \( n_j \) is the number of samples in class \( j \). Table \ref{tab:BC_WC} reports SCR across different datasets and network architectures. Here, our proposed DPNP model uniformly performs as the best followed by the DPP model as the second best, suggesting a successful trade-off between desired properties of the feature space. Again, it can be seen that in higher-dimensional spaces which are frequently used, models are not pushed enough to generate the best organization in the latent space, which may weaken the generalization.

\begin{table}[t]
\centering
\caption{Separation to Compactness Ratio (SCR) on CIFAR-10 and CIFAR-100, higher values indicate better clustering in the feature space}
\label{tab:BC_WC}
\begin{tabular}{c|cc||cc}
\toprule
\multirow{2.5}{*}{\textbf{Model}} & \multicolumn{2}{c||}{\textbf{CIFAR-10}} & \multicolumn{2}{c}{\textbf{CIFAR-100}} \\
\cmidrule(lr){2-3} \cmidrule(lr){4-5}
 & \textbf{512D} & \textbf{3D} & \textbf{512D} & \textbf{10D} \\
\midrule
CE                & 1.26 & 3.17   & 1.12 & 1.87 \\
CL                & 1.77 & 5.77   & 1.19 & 2.62 \\
CPN               & 1.36 & 3.62   & 1.62 & 1.67 \\
CCL               & 1.33 & 11.25  & 1.50 & 3.92 \\
SCCL              & 1.89 & 12.22  & 1.88 & 3.93 \\
DPP (Ours)        & \underline{1.94} & \underline{13.70}  & \underline{1.88} & \underline{5.38} \\
\textbf{DPNP (Ours)} & \textbf{2.03} & \textbf{17.55} &   \textbf{1.99} & \textbf{6.62} \\
\bottomrule
\end{tabular}
\end{table}

\subsection{Visualization of the Feature Space on CIFAR-10 Dataset}
\begin{figure*}[!t]
\centering
\subfloat[CE]{\includegraphics[width=1.75in]{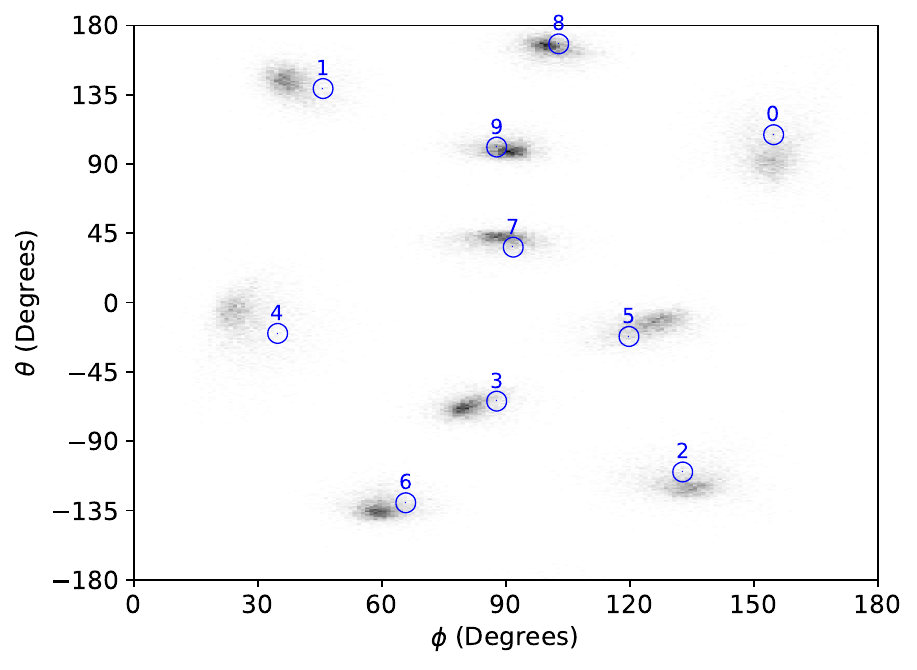}\label{fig_CE_histogram}}
\hfil
\subfloat[CL]{\includegraphics[width=1.75in]{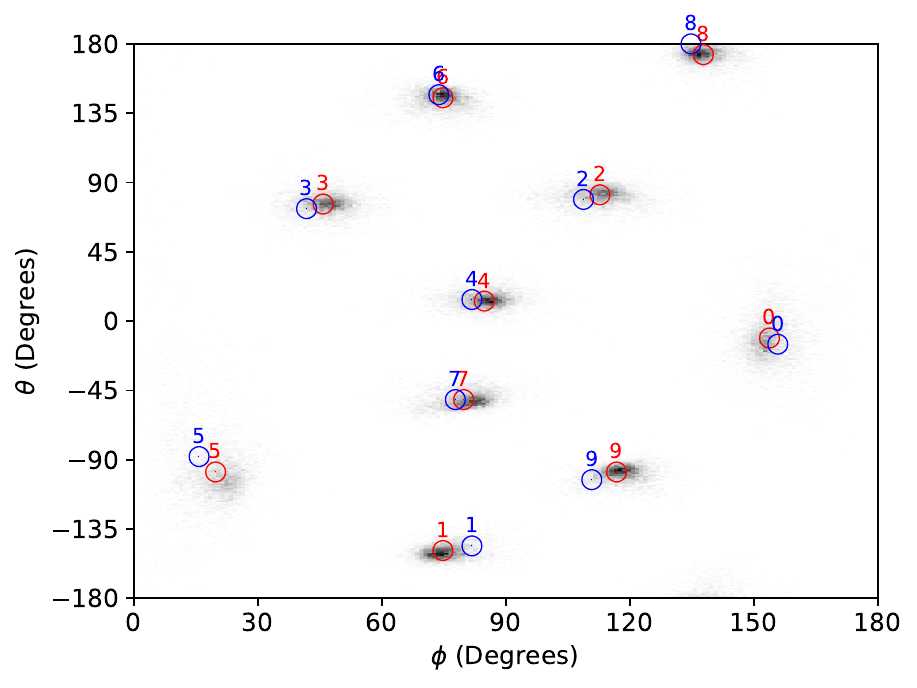}\label{fig_CL_histogram}}
\hfil
\subfloat[CCL]{\includegraphics[width=1.75in]{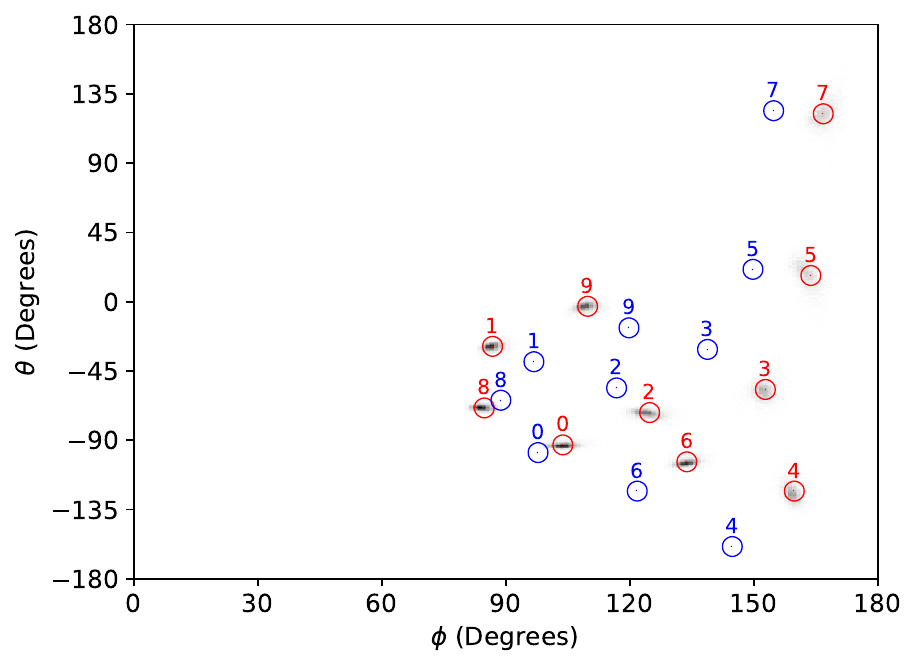}\label{fig_CCL_histogram}}
\hfil
\subfloat[SCCL]{\includegraphics[width=1.75in]{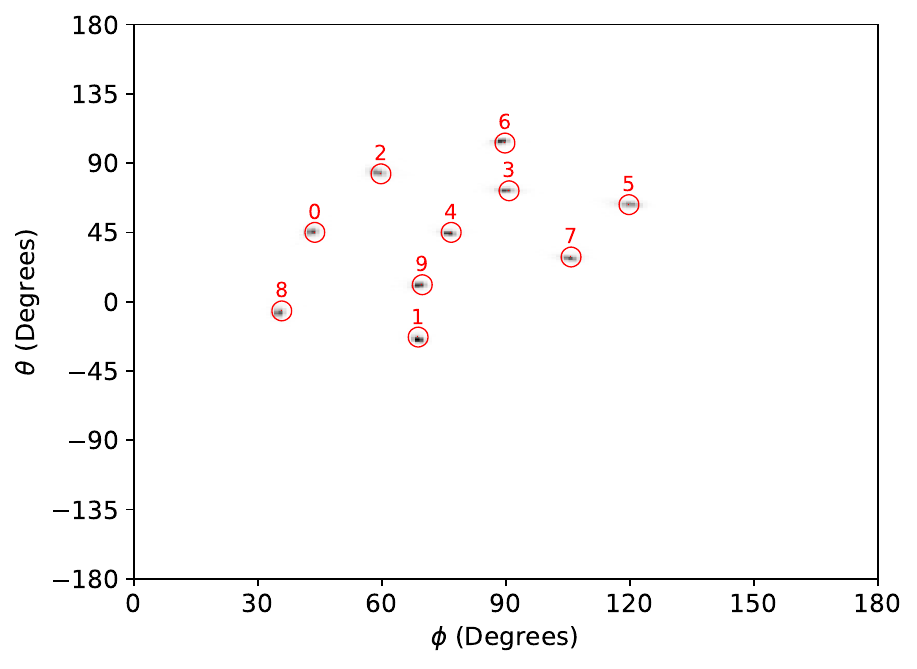}\label{fig_SCCL_histogram}}
\hfil
\\
\subfloat[CPN]{\includegraphics[width=1.75in]{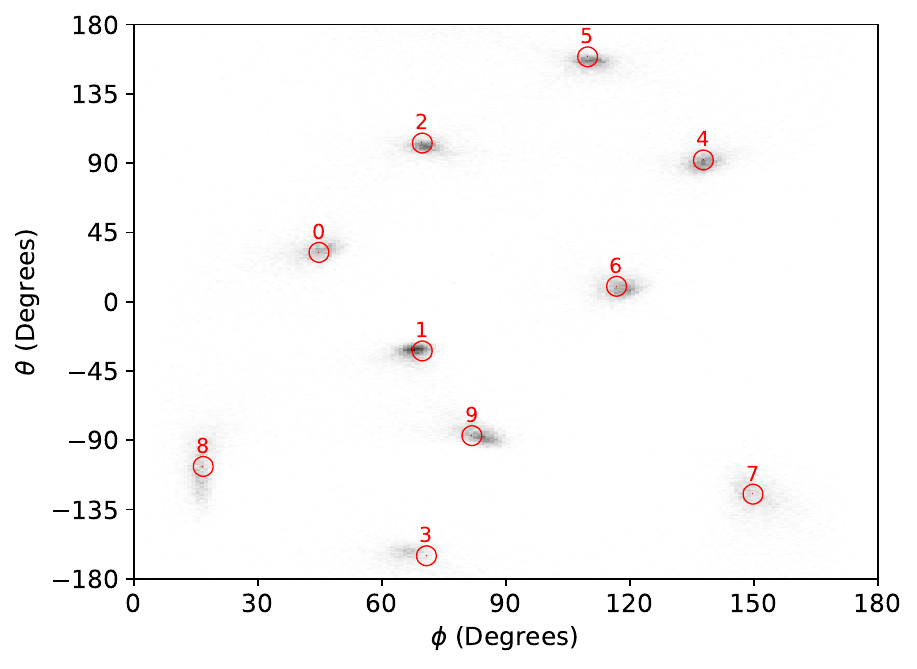}\label{fig_CPN_histogram}}
\hfil
\subfloat[DPP]{\includegraphics[width=1.75in]{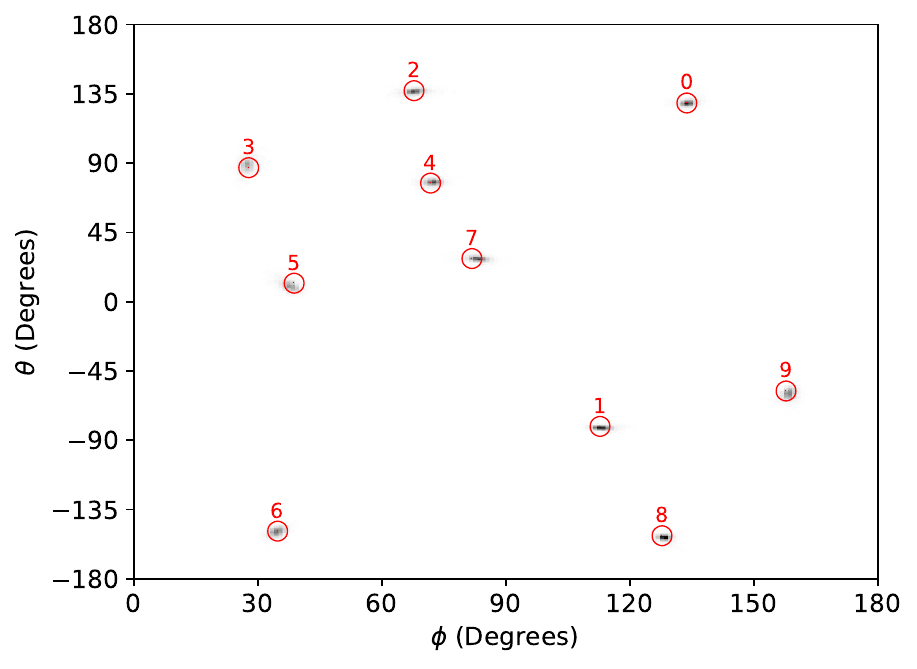}\label{fig_DPP_histogram}}
\hfil
\subfloat[DPNP]{\includegraphics[width=1.75in]{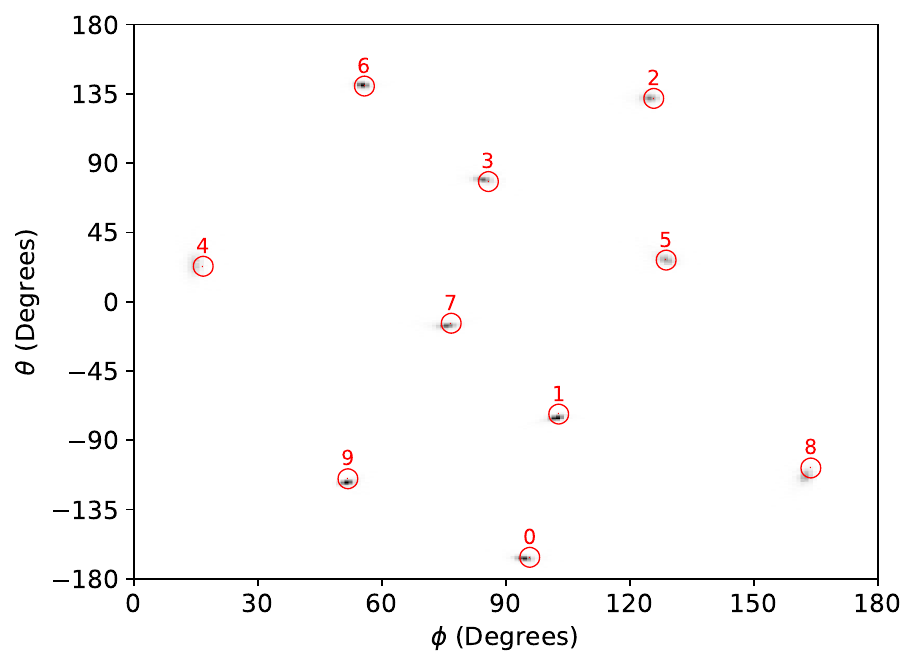}\label{fig_DPNP_histogram}}
\hfil
\subfloat[DPNP (3D view)]{\includegraphics[width=1.75in]{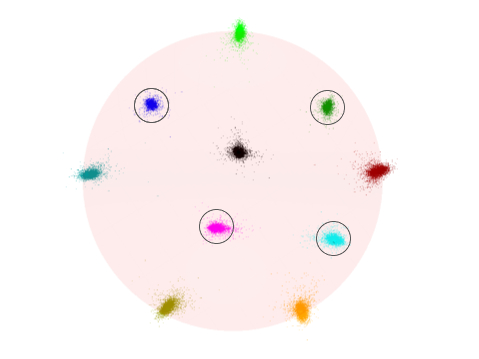}\label{fig_DPNP_3D_histogram}}
\caption{Visualization of the feature space obtained by different models for CIFAR-10 using CE, CL, CCL, SCCL, CPN and the proposed DPP and DPNP models. Sub-figures (a)–(g) present 2D angular histograms of spherical coordinates $\phi$ and $\theta$ of data points after length normalization, mapping the surface of a unit-radius sphere onto each plot. Class centers are marked by red circles and weight vectors are depicted as blue circles. Sub-figure (h) retains the original 3D feature space visualization for the DPNP model, showing the spatial distribution of class members and their clear separation. The encircled classes are located behind while the others face the current view.}
\label{fig_3D_visualizatios}
\end{figure*}

In this section, the feature space obtained by different methods using the reduced ResNet-18 architecture on CIFAR-10 samples is illustrated. This reduced architecture is intentionally selected since it inherently produces a 3D feature space representation for each sample and lets us directly see separability and compactness resulted from each model while avoiding the complexities of traditional high-dimensional projections or dimensionality reduction techniques like t-SNE or PCA. On the other hand, CIFAR-10 does not have too many classes to clutter the view; see Figure \ref{fig_DPNP_3D_histogram}.

However, since 3D plots are more suitable for interactive and live presentations, and on the other hand, many of the studied methods here map the data onto a hypersphere which is a sphere in this particular case, its surface is depicted as 2D histogram plots that effectively represent ($\phi,\theta$), the two angles of the spherical coordinates corresponding to the resultant 3D feature space of each model. The important point to remember when using such a 2D plot is that it corresponds to the surface of a sphere that is opened and mapped onto a rectangle. Hence, up and down edges of each plot should be considered topologically the same, while left and right edges also coincide and form the polar points of the original sphere; see Figure \ref{fig_3D_visualizatios}. 
Class centers or prototypes are marked with large red circles followed by class identifiers to be easily visible, while the weight vectors are similarly indicated with blue circles. The proximity of data points to their respective centers and the alignment of weight vectors with class centers can be evaluated more effectively in these 2D histogram plots.

As already discussed, CE (Figure \ref{fig_CE_histogram}) does not provide particularly compact representations, nor can it align class members with their associated weight vectors. As expected, the simple CE model struggles to provide a well-organized and discriminative feature space. CL shows a slight improvement in intra-class compactness, but weight vectors occasionally diverge from the corresponding class members; see Figure \ref{fig_CL_histogram}. CCL delivers much more compact classes but again separate center/weight parametrization here creates more diversion; refer to Figure \ref{fig_CCL_histogram}. On the other hand, since center points are not directly learned through the loss function gradient, the repulsive effect of CE is not well utilized, and most of the classes are located on one side of the feature space. This leads to ineffective use of the available span and probably weaker generalization.

As depicted in Figure \ref{fig_SCCL_histogram}, SCCL benefits from simpler parametrization and eliminates the divergence between weights and class members. However, the lack of gradient-based learning which is shared with CCL, still piles many classes in a limited region of the feature space. In contrast, CPN can well separate classes through the feature space, a behavior inherited from CE acting on Euclidean distance since prototypes are learned according to the gradient, see Figure \ref{fig_CPN_histogram}. However, dispersion of class members is slightly higher, indicating reduced intra-class compactness compared to CCL and SCCL. This latter effect can be rooted in the difficulties originated from directly working with Euclidean distance in the exponent, as in Distance-based Cross Entropy \cite{Yang2022}.

The DPP model achieves significantly better compactness and separation at the same time, which is evident from its angular histogram depicted in Figure \ref{fig_DPP_histogram}. The DPNP model results, plotted in Figures \ref{fig_DPNP_histogram} and \ref{fig_DPNP_3D_histogram}, further enhance the feature space using explicit constraints for class separation introduced via the concept of negative prototypes. DPNP obtains a well-structured feature space with separated compact classes that can maximize SCR. The alignment between data and the unified weight vectors and class centers in DPP and DPNP is notably stronger, ensuring a more organized and interpretable feature space.

The visual evidence provided here aligns with previously reported quantitative measures, reassuring the superior performance of the proposed models. Additionally, the retained 3D visualization for the DPNP model (Figure \ref{fig_DPNP_3D_histogram}) provides further insight into the spatial distribution of class members, highlighting the consistency between the angular histograms and spatial representations.

\section{Conclusion}
\label{sec:conclusion}
This study introduced a novel approach that integrates positive and negative prototypes for enhanced feature representation learning, called DPNP. This approach seamlessly joins deep learning and discriminative learning around the concept of positive and negative prototypes, while unifying the classifier weight parameters and class centers. These prototypes and the considered attractive and repulsive terms in the loss function significantly improve intra-class compactness and inter-class separation, which are essential for robust classification, especially in tasks with subtle inter-class boundaries or in the lower-dimensional spaces.

In this paper, we studied DPNP in image classification tasks using a variety of network architectures and datasets with different numbers of classes and complexity. Our DPNP model not only achieves higher accuracy across these datasets but also offers a more interpretable and efficient feature space. 
The reduction in the number of parameters, coupled with enhanced class separability, makes this model a promising candidate for a wide range of applications. The implications of this work extend beyond just image classification. The concepts of prototype learning and discriminative feature space can be applied to various other domains where interpretability, accuracy, and adversarial robustness are essential. The framework is designed to be broadly applicable to any deep network architecture.

Future work could explore the application of the DPNP framework to other types of neural network architectures and domains, as well as investigate the potential of this approach in scenarios with more complex or imbalanced datasets. Additionally, further research is necessary on optimizing the training process and exploring different distance metrics to define and utilize prototypes within the feature space.

\bibliographystyle{IEEEtran}
\bibliography{IEEEabrv,DPNP}


\begin{IEEEbiography}
[{\includegraphics[width=1in,height=1.25in,clip,keepaspectratio]{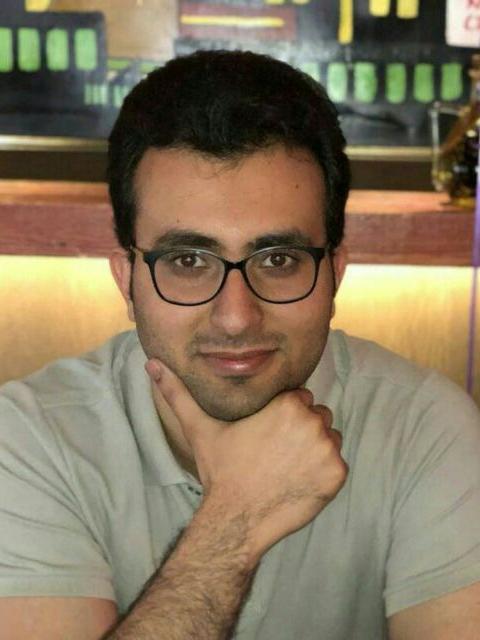}}]
{Ramin Zarei-Sabzevar} received his B.Sc. and M.Sc. degrees in computer engineering and artificial intelligence from Ferdowsi University of Mashhad, Mashhad, Iran in 2014 and 2017, respectively. He is currently a research assistant and a visiting lecturer in Computer Engineering Department, Ferdowsi University of Mashhad.
He is also a lecturer in Computer Engineering department, Sadjad University, Mashhad, Iran.
His areas of research is Neural Networks, Deep Learning, Robot Perception, Probabilistic Models and Information Theoretic Learning.
\end{IEEEbiography}
\begin{IEEEbiography}
[{\includegraphics[width=1in,height=1.25in,clip,keepaspectratio]{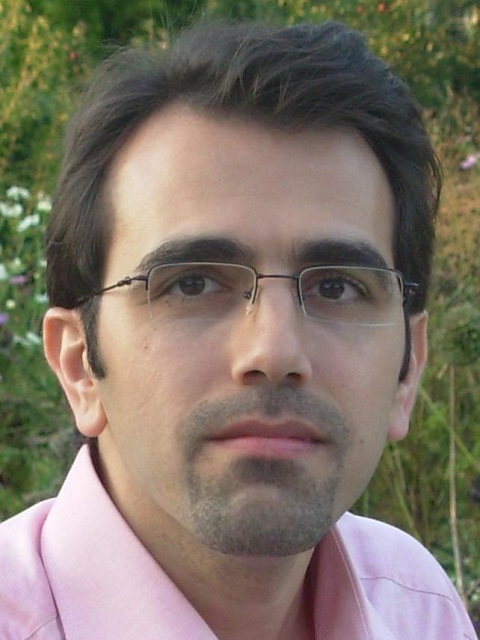}}]
{Ahad Harati} received his BSc. and MSc. degrees in computer engineering and AI \& Robotics from Amirkabir University of Technology and Tehran University, Tehran, Iran in 2000 and 2003 respectively.
He was awarded with Ph.D. in Manufacturing Systems \& Robotics from Swiss Federal Institute of Technology (ETHZ), Zurich, Switzerland, in 2008. He is currently an Associate Professor in Computer Engineering Department, Ferdowsi University of Mashhad. His areas of research is Machine Learning, Probabilistic Models, and Robot Perception.
\end{IEEEbiography}

\vfill

\end{document}